\documentclass[11pt]{article}
\usepackage[a4paper,margin=1in]{geometry}
\usepackage[T1]{fontenc}
\usepackage[utf8]{inputenc}
\usepackage{lmodern}
\usepackage{microtype}
\usepackage{amsmath,amssymb}
\usepackage{graphicx}
\usepackage{booktabs}
\usepackage{multirow}
\usepackage{xcolor}
\usepackage{hyperref}
\usepackage[numbers,sort&compress]{natbib}
\usepackage{caption}
\usepackage{subcaption}
\usepackage{enumitem}
\usepackage{float}
\usepackage{adjustbox}
\usepackage{longtable}
\usepackage{textcomp}
\hypersetup{colorlinks=true,linkcolor=blue!50!black,citecolor=blue!50!black,urlcolor=blue!50!black}
\graphicspath{{figures/}}
\newcommand{\ci}[2]{[#1,\,#2]}

\newcommand{\thisdoi}{10.5281/zenodo.23163693}
\newcommand{\evidencedoi}{10.5281/zenodo.23163299}

\title{Interrupting the Loop:\\
Periodic Subject Changes Raise Judged Surprise and Connection\\
in Base Language Models}
\author{Roberto I. Ono Filho\\
Independent researcher\\
\texttt{ono.roberto@gmail.com} \quad ORCID \href{https://orcid.org/0009-0006-8650-629X}{0009-0006-8650-629X}}
\date{Version 2, 5 October 2026 (version 1: 20 August 2026, arXiv:2608.19893v1)\\
Zenodo \href{https://doi.org/\thisdoi}{doi:\thisdoi}; the changes are listed at the end of the main text}

\begin{document}
\maketitle

\begin{abstract}
Where does the novelty a base language model produces with no task come
from, and what can an LLM judge of a long stream actually see? We dismantle a
cognitively inspired generation loop over 24 conditions on three base models.
Most of its effect lives in one operation: a new subject injected every few
hundred tokens (an \emph{interruption}) into a stream whose literal
repetition is damped (\emph{habituation}). We judge windows of generated text
only, with the premise as the unit ($n=10$) and a judge measured for
repeatability, against a second judge family and against human readers.
Under that protocol the interruption raises judged surprise by 1.2 to 1.4
points and connection by 0.7 to 0.9 over habituation alone. A connective
that asks for continuity hurts; a bare paragraph break adds nothing
detectable; a reset context does at least as well as a kept one; and a
pre-registered replication on new premises confirms the primary contrast.
Three things the window judge could not see changed the first version of
this study, and we think they are of general use. The judge scores the
experimenter's injected sentence as the model's own. A fixed rotation of
injected sentences makes the model replay its own earlier segments (65--80\%
of post-interruption windows at periods 150--300 are mostly verbatim
replay), and the judge scores the replay as surprise and, above all, as
connection; at period 300 the source of 62\% of windows lies beyond the
judge's 600-token horizon. And the local gains do not compose: no arm
produces an integrated document. The
salience monitor, the in-loop judge, memory across interruptions and a
judge-gated Review run with a gate that opens add nothing. On a problem with
a verifier (online bin packing), the interruption multiplies valid, distinct
candidate heuristics three- to fourfold without raising the quality of the
best. We report an evaluation protocol for long generation and a controlled
characterization of a simple intervention, not a mechanism of creativity.

\end{abstract}

\section{Introduction}

Ask a language model for something new and you will usually get something
fluent, plausible and familiar. Three places are commonly blamed, and three
places are commonly optimized. The first is the \emph{sampler}: the tail of
the next-token distribution, tamed by nucleus, typical or min-$p$ truncation,
or courted by higher temperature. The second is the \emph{prompt}: the
input, engineered to steer the model somewhere it would not go on its own.
The third, less often, is the \emph{loop}: what happens when a model's output
becomes its own next input and nobody asks anything.

The loop is where human novelty is usually placed. Insight rarely arrives as
the answer to a strange question. It arises inside a closed circuit of
thought feeding on its own output, during incubation, mind-wandering and
sleep, when a spontaneously generated deviation survives critical review and
is linked back to what came before. The neuroscience of creative cognition
describes three coupled systems: a default-mode network that generates, an
executive network that evaluates, and a salience network that decides what
deserves attention. More creative people show more coupling between the
first two, with salience regions coupling first
\citep{beaty2016dynamics,beaty2018robust}. Creativity itself is often
characterized as connecting semantically distant concepts
\citep{kenett2018semantic}, and incubation, leaving a problem and coming back
to it, has a measured effect on solving it \citep{sio2009incubation}.

This paper asks, with controls, where the novelty a base language model
produces with no task comes from, and which of the operations usually
credited for it survive measurement. The outcome we measure is narrow and we
name it as such: \emph{judged narrative surprise, connection and coherence}
on short windows of forced open-ended continuation, read by an LLM judge
that is itself measured for repeatability, against a second judge family
and against human readers. This is one ingredient of creativity, the
appropriately unexpected turn, and not creativity. The title names the
question the program set out with; the results answer the narrower one.
Along the way the study became, in part, a study of its own instrument. An
LLM judge reading windows of a long stream misses things that change the
answer, and the corrections we had to make are, we think, of use beyond
this paper.

We test the three places in turn. They are three related studies, with
different generators, tasks and sample sizes, joined by one question and
one measurement philosophy. They are not a factorial decomposition of
sampler, prompt and loop under common conditions; the loop study is the
paper's contribution and the other two are its motivation. One measurement
stack is shared by every experiment: base models run locally at 8 bits (a
scope decision, Section~\ref{sec:method}); LLM judges from a different
model family than the generator, sampled $k$ times per window with the
median as the score; the premise, not the window, as the unit of inference
in the loop experiments; and, where a public training corpus exists,
verbatim novelty computed against it.

\paragraph{Terms.} A few words recur and are defined here. A \emph{cell} is
one premise run under one condition. A \emph{battery} is a set of conditions
run on the same ten premises. The \emph{scaffold} is the full architecture
we started from (a salience monitor, an in-loop judge, forgetting,
reseeding, re-encounter), described in Section~\ref{sec:loop-method}. The
\emph{ladder} is the sequence of four arms that the paper keeps returning
to: bare generation, bare plus habituation, habituation plus interruption,
and the scaffold. A \emph{degeneration mode} is the repetitive or
corpus-like state a base model falls into under forced continuation
(literal loops, website footers, exam keys, translation tables).

\paragraph{The sampler and the prompt.} We built the strongest version of
the ``fertile error'' idea we could: an entropy-banded anti-probable decoder
with a coherence floor. It works exactly as far as the surface. Four-gram
novelty against the OLMo-2 training corpus roughly doubles and verbatim
training blocks fall fourfold, while inside a generation loop and in
verified search the decoder shows no detectable difference from plain
sampling at the level of ideas. We also built inputs far from any human
prompt and found no benefit from their improbability under the
operationalizations we tested. Both studies are reported in
Appendix~\ref{sec:sampler}; they are why the program turned to the loop.

\paragraph{The loop.} Left to continue a premise with the end-of-text token
masked, a base model falls within a few hundred tokens into a degeneration
mode and stays there. The scaffold revives it. We then take the scaffold
apart, in four batteries and 24 conditions, and find that most of its effect
lives in two operations, neither of them the elaborate ones:
\emph{habituation}, a windowed repetition penalty that keeps the loop from
repeating its literal past, and \emph{interruption}, a new starting
sentence injected every few hundred tokens. The interruption is the larger
of the two. We then ask what a good interruption is made of. It must lead
away: injecting the premise or the stream's own past is as bad as not
interrupting, a boundary without a new subject adds nothing detectable, and
a boundary that asks for continuity hurts. It must be new each time: a fixed
rotation of four sentences makes the model replay its own earlier segments,
which a window judge does not recognize as replay. It works whether the
earlier text is kept in the model's context or dropped, and dropped scores
higher, in part because the model cannot replay. A clock is as good a metronome as the
salience monitor at the same rate, and no period beats a break every
150--300 tokens on the generated text itself. We also show what the
interruption does not do. Read as whole documents, none of these streams
builds an integrated or developing text, and the interrupted stream reads as
a sequence of restarts; the operator acts on windows, not on wholes. On a
problem with a verifier it multiplies valid, distinct candidates without
improving the best of them. The ladder replicates on three generator models
from two families and under a second judge family, its ordering on surprise
is reproduced by independent human readers on windows of the first protocol,
and a pre-registered replication on ten new premises confirms the primary
contrast. Finally we look inside the
network, descriptively. Residual-stream geometry at 13 sampled layers shows
that bare generation moves least at every sampled layer, that judged
surprise co-varies with surface departure over deep continuity, and that
the interruption that scores best barely moves the deep state.

\paragraph{Contributions.} In the order we now think they matter:
\begin{enumerate}[leftmargin=1.5em,itemsep=0.1em]
\item \textbf{What an LLM judge cannot see in long-form generation.} Three
  artefacts that inflated the first version of this study, and that any
  evaluation of long generation with a windowed LLM judge is exposed to: the
  judge reading text the experimenter injected as the model's own; the model
  replaying its own earlier segments from beyond the judge's horizon, which
  the judge scores as surprise and connection; and local gains that do not
  compose into a whole. With them, the protocol that corrects them:
  generated-only windows, fresh-only estimates, a document-level judgment,
  the premise as the unit, a pipeline test--retest, a second judge family
  and, on first-protocol windows, human readers.
\item \textbf{A controlled characterization of the interruption.} Which part
  of an interruption carries the effect (the new subject; not the boundary,
  not the kept context), its content, timing and period, replications on
  three base models from two families, a post-trained model, an unquantized
  model and a second genre, and a pre-registered confirmatory replication on
  new premises.
\item \textbf{Negative results on the architecture, and a first probe with a
  verifier.} The salience monitor as a trigger, the in-loop judge, the
  forgetting reseed, memory across interruptions, the judge-gated Review run
  with a gate that opens, and accumulation over the document: none adds to
  the minimal operator. On a problem with a verifier the operator
  multiplies valid candidates and not the quality of the best.
\item A calibrated anti-probable decoder whose novelty is real at the surface
  and undetectable at the level of ideas, a null on improbable inputs, and a
  descriptive residual-stream analysis.
\end{enumerate}
Code, the human-rating packs and the dated laboratory notebook are in the
public repository; the run data and LLM judgments are in an evidence deposit
(\href{https://doi.org/\evidencedoi}{doi:\evidencedoi}). Individual human
ratings are not released, to protect rater anonymity; only aggregate
statistics are reported.

\section{Related work}
\label{sec:related}

\paragraph{Decoding and the tail.} Nucleus sampling \citep{holtzman2020curious}
truncates the unreliable tail of the next-token distribution. Locally
typical sampling \citep{meister2023typical} targets human-like surprisal.
Min-$p$ \citep{nguyen2025minp} scales the truncation with the model's
confidence and is exactly the coherence floor we use. Contrastive decoding
\citep{li2023contrastive} and plug-and-play steering \citep{dathathri2020pplm}
shape the distribution toward or away from a reference. All of these
regulate the tail. Our anti-probable sampler courts it, inside an entropy
band and above a floor, and we find that this buys surface novelty only.

\paragraph{Repetition, self-reinforcement and entrainment.} The collapse of
a base model left to itself is a well-studied phenomenon.
\citet{zhu-etal-2023-penalty} name the \emph{self-reinforcement effect}, by
which a repeated token becomes more probable each time it repeats, and
suppress it with a repetition penalty restricted to a recent window. The
graded, windowed penalty we call \emph{habituation} is a decoding-time
operator of the same family, and we claim no novelty for it.
\citet{guan-huang-2023-mitigating} attribute the tendency to repeat to a
learning bias of maximum-likelihood training and correct it with
self-contrastive training. \citet{xu-etal-2023-look} track the distance
between the current next-token distribution and past ones to detect
repetition and topic drift, and steer decoding accordingly. At the
mechanistic level, \citet{niu-etal-2025-llama} show that language models
assign higher probability to tokens present in their context even when those
tokens are irrelevant (\emph{contextual entrainment}), and locate heads that
carry it. Our bare arm is entrainment on the model's own output, and our
self-copy finding is entrainment on the experimenter's injected sentences.
What this paper adds is not a new remedy for repetition. It is a measurement
of what the remedy leaves untouched (a habituated stream is fluent and still
not creative), of what a second, structural operator adds on top of it, and
of how the replay that entrainment produces can escape a windowed judge.

\paragraph{Evaluating creativity, and the judge.} \citet{nakajima-etal-2026-beyond}
show that the widely used Divergent Association Task ignores appropriateness
and propose to score novelty conditional on it. Our three separate
dimensions (surprise, connection, coherence), reported without a composite,
follow the same logic of not letting novelty be bought with incoherence.
\citet{saakyan2026death} show with 8,618 expert annotations that about 91\%
of top-quartile $n$-gram-novel expressions are not judged creative; our
factual-paraphrase escape mode and the sampler's idea-level null are
independent confirmations of the same gap. Infini-gram
\citep{liu2024infinigram} and Rusty-DAWG \citep{merrill2024rustydawg} make
verbatim novelty against a training corpus computable at scale; we use
infini-gram against the OLMo-2 corpus. On the instrument itself,
\citet{fein-etal-2026-litbench} find that the strongest off-the-shelf judge
they tested agrees with human preferences on creative-writing pairs only
73\% of the time, and \citet{haldar-hockenmaier-2025-rating} document low
intra-rater reliability of LLM judges across runs. Both cautions apply to
this study. We answer the second by scoring every window with $k=5$
independent calls and using the median, and by reporting the resolution of
the instrument. We answer the first only partly, with a second judge family
and with human raters on a subset, and we treat the judge as the study's
main limitation (Section~\ref{sec:limitations}). To these known problems
we add three that are specific to judging long streams through windows:
the injected text inside the window, the replay from beyond the judge's
horizon, and the gap between windows and wholes.

\paragraph{Loops and steering in story generation.} Outer loops that steer
a generator are common in story generation. SWAG \citep{pei-etal-2024-swag}
lets a second model choose the next action for the story model. Collective
Critics \citep{bae-kim-2024-collective} refine a plan and its expression with
several critic models. STORYTELLER \citep{li-etal-2025-storyteller} adds a
plot-planning structure to keep long stories coherent and cohesive. Against
this literature our contribution is not the idea of an outer loop but its
reduction. The operator we isolate is one injected sentence, without a
planner, a critic or a reward; the factorial ablation shows which part of
the loop produces the effect; and the temporal characterization (period,
decay, timing) is one a planning framework does not provide.

\paragraph{Verified search.} FunSearch \citep{romeraparedes2024funsearch}
and AlphaEvolve \citep{novikov2025alphaevolve} pair a language model with a
hard evaluator and let selection do the work. Our bin-packing results, in
which neither the anti-probable sampler nor the interruption raises the
quality of the best verified candidate while the interruption multiplies the
number of valid ones, suggest that the generation loop supplies variation
and the evaluator must supply the rest.

\paragraph{Reading the residual stream.} The logit lens is known to be
fragile in intermediate layers, and the tuned lens \citep{belrose2023tunedlens}
was proposed as a less biased alternative. We use the logit lens only for a
coarse commitment layer and otherwise work with mean-centered residual
geometry. The network section is descriptive, and its readings should be
repeated with a tuned lens before they are taken as mechanism.

\paragraph{Creative cognition.} Creative thought as coupling of default-mode
and executive networks \citep{beaty2016dynamics}, with salience regions
coupling first \citep{beaty2018robust}; creativity as connecting semantically
distant concepts \citep{kenett2018semantic}; incubation effects
\citep{sio2009incubation}; predictive processing \citep{clark2013whatever};
dreams as anti-overfitting noise \citep{hoel2021overfitted}; Boden's
novelty/surprise/value triad \citep{boden2004creative}; conceptual blending
\citep{fauconnier2002way}; novelty search \citep{stanley2015greatness}. The
reverie loop of Section~\ref{sec:loop} was an explicit engineering of the
first three into a text loop. Its ablation is, in a sense, a test of which of
these ingredients a language model needs, and the answer, an interruption on
a clock, is humbler than the hypothesis.

\section{Method}
\label{sec:method}

\subsection{Measurement stack}
\label{sec:stack}

Every experiment shares one stack. \textbf{Generators} are three base models
from two families: Qwen3-8B-Base, Qwen3-30B-A3B-Base (a mixture-of-experts
model with 3B active parameters, 53 tokens/s, the main generator) and
OLMo-2-13B, whose training corpus is public. All are quantized to 8 bits
(MLX affine quantization, group size 64) and run on Apple silicon through
MLX, with a numpy sampler that sees the full logits. Restricting to base
models is a scope decision, not a finding. An instruction-tuned model brings
a dialogue format and preference training that change what an open-ended
continuation is; whether the operators studied here transfer to it is left
open (Section~\ref{sec:limitations}). \textbf{Judges} are Claude Opus~5 and,
in the instrument section, Claude Sonnet~5 (Amazon Bedrock) and Kimi K2.6
(OpenRouter). In the loop experiments the judge is always from a different
family than the generator. The idea experiment of Section~\ref{sec:prompt}
used Claude both to develop and to judge, a within-family judgment we flag
there. Each window is judged $k$ times ($k=5$ in the loop experiments, $k=3$
in the earlier ones) with independent 0--10 dimensions, and the median is
the score. For loop windows the dimensions are \emph{surprise} (how
unexpected the window is given the earlier text; 0 is the obvious
continuation, 10 is startling yet not random), \emph{connection} (does it
bring together two distant regions of the earlier text, or an old region
with something new, in a way that makes sense) and \emph{coherence} (does it
hold together as text; 0 is word salad or document boilerplate). For ideas
we use a nearest-equivalent plus novel-delta rubric. The full judge prompts,
model identifiers, parameters and dates are in Appendix~\ref{app:repro}.
\textbf{Instrument calibration} (Section~\ref{sec:instrument}) measures the
intra-window spread of the $k$ judgments and the judge's agreement with a
second judge family and with human raters. \textbf{Objective novelty},
where a public corpus exists, is computed with infini-gram
\citep{liu2024infinigram} against the OLMo-2 corpus.

\paragraph{Unit of analysis and statistics.} A \emph{cell} is one premise
run under one condition for 4,500 generated tokens. Each condition has ten
cells, one per premise, and conditions share the ten premises. The unit of
inference is the cell. The windows of a cell are averaged into one score per
dimension, and every comparison between two conditions is a paired
comparison of ten cell means. We report the mean over cells with a bootstrap
confidence interval over cells. For each pair of conditions we report the
mean paired difference with its bootstrap CI, an exact sign-flip permutation
$p$-value on the ten paired differences (all $2^{10}$ sign patterns) and
Cliff's $\delta$ on the cell means. Where a question involves several
comparisons we add Benjamini--Hochberg $q$-values within that family.
Window-level statistics (bootstrap on windows, Mann--Whitney) appear only in
the appendix, as descriptive, because windows from the same stream are not
independent. This cell-level analysis and the window protocol below were
adopted after an external review of a first version of this manuscript,
which had analyzed windows as the unit. The arms of battery~3 were generated
after that change; the earlier arms were re-judged and re-analyzed under it.

\subsection{The anti-probable sampler}
\label{sec:sampler-method}

At each step the model's next-token distribution is computed as usual.
Tokens below a relative probability floor ($p < 0.05\,p_{\max}$, the min-$p$
rule) are never chosen. If the entropy of the distribution falls inside a
band $[H_{\min}, H_{\max}]$, where the model is undecided but not lost, the
remaining candidates (at most 128) are re-scored by
\[
\text{score}(t) = \log P(t \mid \text{context}) + \lambda \cdot z\!\big(d(t,\ \bar{c})\big),
\]
where $d$ is the cosine distance between the token's input embedding and an
exponential moving average $\bar c$ of the recent context in the model's own
embedding space (half-life 16 tokens in the sampler experiments, 48 in the
loop), and $z$ standardizes the distances across the step's candidates so
that $\lambda$ reads in nats per standard deviation. The token is then
sampled from the softmax of the scores. Below the band the model is
confident and we let it be. Above the band the distribution is a genre fork,
and pushing there produces collapse. The band was $H_{\min}=2.0$,
$H_{\max}=4.5$ nats in the sampler experiments and $[1.8, 4.5]$ in the
loop's drift regime; the values were set on calibration probes of Qwen3-8B
and transferred unchanged to the other models. Two optional terms are used
in the loop. \emph{Habituation} is a repetition penalty: the probability of
a token that occurred in the last 512 positions is divided by $1.15^{m}$,
where $m$ is the number of its occurrences in that window, before the floor
and the re-scoring. The \emph{bridge} bonus (weight 1.5 in the full
scaffold) rewards candidates close to anchor regions visited long ago and
far from the recent context.

\subsection{The reverie loop}
\label{sec:loop-method}

\paragraph{Forced continuation without a task.} A cell starts from a single
sentence, the \emph{premise} (for example \emph{``The town had two clocks,
and nobody remembered which one had been right first.''}), and nothing
else: no instruction, no role, no question. A base model continues it, and
from then on its only input is its own output. Each token is predicted from
the premise plus everything the model has written. In all arms but one the
end-of-text token is masked, so the stream never stops by itself: the model
is forced to keep writing one document for 4,500 tokens. This is a
deliberately hard regime, and part of what the paper measures is how a base
model behaves in it.

\paragraph{Interruption.} At chosen moments the generator is paused, a short
text is \emph{injected} into the stream as if the model had written it
(pre-filled into the model's cache), and generation resumes from there. The
model receives no signal that the text was injected; to it these are the
next tokens of the document. Everything written before stays in the context.
The model keeps all of it and is only made to open a new sentence somewhere
else. Figure~\ref{fig:example} shows a real interruption. The default
injected texts are four neutral subject changes used in rotation
(\emph{``That night she dreamed of something else entirely:''},
\emph{``Meanwhile, in a city with no name,''}, \emph{``There is an older
story about this, and it goes:''}, \emph{``A question nobody had asked
yet:''}). Other contents are studied in Section~\ref{sec:loop}.

\begin{figure}[t]
\centering
\begin{minipage}{0.96\linewidth}\small
\textbf{Premise:} \emph{Every morning the baker counted the loaves twice, and every morning the count was different.}\\[0.3em]
\textbf{\ldots the stream, 2,000 tokens later:} \emph{There was never more than thirty, but there were always twenty-one or twenty-two. That morning he counted them three times, but he got no nearer to the number. He couldn't understand it. He opened the door of the oven and pulled out a loaf of bread. He cut it open. It was all}\\[0.3em]
\textbf{Injected:} \emph{Meanwhile, in a city with no name,}\\[0.3em]
\textbf{The model continues:} \emph{an old woman sat at a window with her cat. She had no idea who she was or where she came from, or why she'd been left alone in this strange city with this strange cat. All she knew was that every day something went wrong with the cat: It would grow ill for no reason and then recover; it would lose its fur for days; even die for a few hours! Every day this happened, but every day it also recovered. And then one day, when the old woman looked out\ldots}
\end{minipage}
\caption{A real interruption (Qwen3-30B-A3B, seed 2, period 300). The
injected sentence is indistinguishable from the model's own text; the baker
remains in the context and can be returned to.}
\label{fig:example}
\end{figure}

\paragraph{The full scaffold (DREAM).} The architecture we started from adds
several mechanisms around the same stream. A \emph{salience monitor} watches
the stream's own telemetry (a fast moving average of token embeddings and
the step entropy) and fires on four events: a semantic \emph{jump} (the
average moved far from where it was 32 tokens earlier), a
\emph{crystallization} (entropy dropped after a stretch of wandering), a
\emph{recurrence} (the average came close to a region visited long ago and
not recently) and \emph{stagnation} (it has not moved for a long stretch). A
surface \emph{genre-collapse} detector watches lexical diversity,
capitalization and layout symbols. A judge (Claude Sonnet~5, $k=3$) is
called on salience events during generation; a passing verdict (median
score $\ge 5$) triggers \emph{escalation}, a narrower regime plus an
injected textual return to the premise. A \emph{kick} (a stretch of harder
push) answers stagnation, and after repeated stagnation comes a subject
change with \emph{selective forgetting}: the working memory is rebuilt in a
fresh cache from the premise, up to three judged-good windows and the last
200 tokens, plus the new seed, so that the accumulated context cannot pull
the seed back into the degeneration mode. Regions visited persist as anchors
for the bridge term.

The full scaffold is the only arm in which a judge is called inside the
loop. In the main generator's scaffold cells the judge was on and reviewed
80 salience events. None reached the threshold, so escalation and the judged
re-encounter never fired, and the judge changed no generated token. In the
replications on the other two generators the scaffold ran without the
in-loop judge. In practice, then, the scaffold's active mechanisms were the
kick and the reseed with forgetting, and in every arm of this paper the
scores come from a judge that sees the stream only after generation. Every
one of these mechanisms was born from a calibration probe in which a base
model without a task fell into a specific degeneration mode: early
end-of-text, erudite rumination, literal four-sentence orbits, website
footers, translation tables. The generator's pretraining diet was the
dominant variable in those probes. OLMo-2 sinks into footers even with
forgetting; Qwen3-30B drifts in prose. With 3,000 tokens of boilerplate in
the cache, every injected seed was pulled back within about twenty tokens;
forgetting is what lets a seed take. The results below show that this
architecture is not what carries the effect. We keep it as the origin of the
study and as one arm.

\paragraph{Conditions.} Table~\ref{tab:conditions} lists the arms. All loop
arms use $\lambda=0$ (no anti-probable push), since the push had no
detectable effect inside the loop (Section~\ref{sec:sampler}). The arms
differ in whether the habituation penalty is on, whether the end-of-text
token is masked, whether the context is preserved or reset at an
interruption, and whether, when, how often and with what the stream is
interrupted. Ten narrative premises (Appendix~\ref{app:repro}) are shared by
all arms. The sampler's random-number seed is the same (0) in every cell, so
two arms that make identical decisions up to a point produce identical
tokens up to that point, and the premise is the only source of variation
between the cells of an arm.

\begin{table}[t]
\centering\small
\begin{tabular}{@{}llp{7.4cm}@{}}
\toprule
arm & habituation & interruption \\
\midrule
bare & off & none: continuous generation, EOS masked \\
bare + habituation & on & none \\
habituation, EOS allowed & on & none; the model may emit end-of-text and start a new document \\
habituation 1.3 & on (stronger) & none \\
interruption, no habituation & off & neutral subject change every 150 or 300 tokens, context preserved \\
habituation + interruption 150 & on & neutral subject change every 150 tokens, context preserved \\
salience only & on & none; salience events mark review windows \\
DREAM scaffold & on & full scaffold: salience, in-loop judge, kick, reseed with forgetting, re-encounter \\
\midrule
content: re-encounter / premise / own past & on & every 150 tokens: a return-to-the-premise stitch / the premise itself / a window of the stream's own past ($\ge$400 tokens back) \\
salience: re-encounter & on & the stitch injected on each salience event, no judge gate \\
period 75 / 300 / 600 / 900 & on & neutral subject change at other periods (900 also with the stitch) \\
\midrule
sham break 300 & on & a paragraph break (``\textbackslash n\textbackslash n'') every 300 tokens, nothing else \\
sham continuity 300 & on & ``And so, as before,'' every 300 tokens: a boundary that asks for continuity \\
reset + subject change 300 & on & neutral subject change every 300 tokens on a \emph{reset} context (premise + injected sentence only) \\
reset + break 300 & on & a paragraph break every 300 tokens on a reset context \\
\midrule
judge-gated 150 & on & the neutral change every 150 tokens unless Opus reads the last 128 tokens as a find (surprise and coherence $\ge5$), which is left to run \\
\bottomrule
\end{tabular}
\caption{Conditions: the habituation $\times$ interruption factorial with its
baselines (top), the content and timing arms, the boundary and context
controls, and the judge-gated arm (bottom). Ten premises each, 4,500 tokens per cell, main
generator Qwen3-30B-A3B-Base; the ladder bare / habituation / habituation +
interruption 150 / scaffold is replicated on Qwen3-8B-Base and OLMo-2-13B.}
\label{tab:conditions}
\end{table}

\paragraph{Judging the stream: generated-only windows.} After generation,
each cell is scored on windows made only of model-generated tokens. In an
interrupted cell, a window of 96 tokens starts 32 tokens after the end of
each injected sentence; we call it the \emph{post-interruption window}.
Further windows start 160, 300, 450, 600 and 750 tokens after the injection
when the segment is long enough, and no window crosses an injection. In a
cell without injections the same window shape is cut on a uniform grid every
150 tokens. The judge sees the 600 tokens preceding the window as context,
which may contain an injected sentence, and grades only the window, on the
three dimensions, five times; the median is the window's score. Up to six
post-interruption windows per cell, evenly spread over the stream, and three
per deeper offset are judged. The primary measure of a cell is the mean of
its post-interruption windows (grid windows in uninterrupted arms); the
deeper offsets give the decay curve of Section~\ref{sec:loop}. Each cell
stores its token stream and the exact position of every event and
injection. The protocol of the first version of this study cut 160-token
windows right before each injection and on a 150-token grid, so that a
window could contain an injected sentence; it is kept in
Appendix~\ref{app:events} for comparison.

\section{The loop}
\label{sec:loop}

All results in this section use Qwen3-30B-A3B-Base unless stated, ten
premises, 4,500 tokens per cell, $\lambda=0$, Claude Opus~5 as judge with
$k=5$, generated-only windows, and the cell as unit. Numbers are means over
cells (one per premise), intervals are 95\% bootstrap CIs over cells, and
paired differences come with an exact sign-flip permutation $p$ on the ten
paired cells and Cliff's $\delta$ on cell means. Full tables, including the
Benjamini--Hochberg $q$-values within each family of comparisons and the
event-protocol tables of the first version, are in Appendix~\ref{app:loop}.
Every figure shows one point per premise.

\subsection{Forced continuation degenerates; the scaffold revives it}

Left to continue a premise with nothing else, the base model falls within a
few hundred tokens into a degeneration mode and stays there: \emph{``Highly
recommended. Highly recommended.\,\ldots''} for two thousand tokens; a table
of three-digit numbers; an English-exam answer key. Judged surprise is 0.45
\ci{0.28}{0.68}, connection 0.30, coherence 2.52 (ten cells). The full
scaffold brings the stream to 2.70 / 1.85 / 6.02, and every premise moves in
the same direction on surprise and connection, nine of ten on coherence (Figure~\ref{fig:q1}). But the scaffold is not the
smallest thing that does this.

\subsection{Taking the loop apart: habituation and interruption}
\label{sec:ablation}

\begin{figure}[t]
\centering
\includegraphics[width=\linewidth]{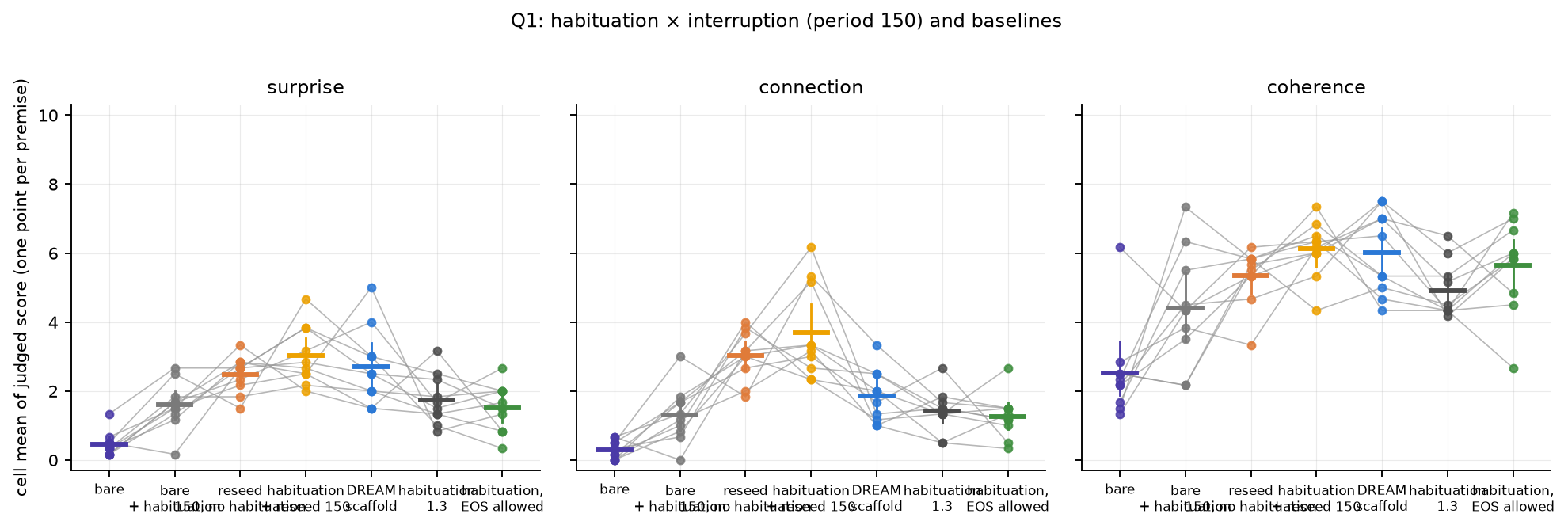}
\caption{Habituation and interruption: bare, bare + habituation, interruption
without habituation, habituation + interruption 150, the full scaffold, and
the two habituation baselines (stronger penalty; EOS allowed). One point per
premise, grey lines join the same premise across arms, bars are cell means
with 95\% bootstrap CIs over cells (Qwen3-30B-A3B, generated-only windows).}
\label{fig:q1}
\end{figure}

\paragraph{The ladder.} Two operations of the scaffold suffice, and they are
not the elaborate ones (Figure~\ref{fig:q1}, Table~\ref{tab:ladder}). Adding
only habituation to bare generation (a windowed repetition penalty, no
interruption) removes the literal loops. Surprise goes from 0.45 to 1.60
($\Delta$ +1.15 \ci{+0.73}{+1.48}, $p=0.004$, $\delta=+0.80$), connection
from 0.30 to 1.32, coherence from 2.52 to 4.40. The stream is fluent and, to
the judge, still not very surprising. Adding a neutral subject change every
150 tokens over the preserved context takes it to 3.02 / 3.68 / 6.12. Over
habituation alone that is +1.42 \ci{+0.92}{+2.03} of surprise ($p=0.002$,
$\delta=+0.85$), +2.37 \ci{+1.57}{+3.28} of connection ($p=0.002$,
$\delta=+0.93$) and +1.72 \ci{+0.35}{+2.90} of coherence ($p=0.04$). This
plain interruption matches the full scaffold on surprise (3.02 vs 2.70),
beats it on connection (3.68 vs 1.85; the scaffold's own gain over
habituation is +0.53 \ci{+0.15}{+0.97}) and equals it on coherence (6.12 vs
6.02). The salience monitor, the in-loop judge (which never fired) and the
forgetting reseed add nothing detectable over the plain interruption, and
the scaffold gives back connection. Battery~3 will show that the loss is not
the forgetting itself, since a complete reset every 300 tokens scores at
least as well as the preserved context. The likelier cause is how rarely the
scaffold interrupts: 0--3 reseeds per cell, after three consecutive
stagnations, against 30 for the clock. On these all-windows numbers,
habituation accounts for part of the surprise and the interruption for the
rest and for most of the additional connection. The self-copy check below
cuts the connection number down and leaves the surprise one.

\paragraph{The factorial and the baselines.} Battery~3 completes the
$2\times2$ (Tables~\ref{tab:gen-q1} and \ref{tab:gen-q3-3}). The neutral
subject change \emph{without} habituation reaches 2.48 / 3.03 / 5.35 at
period 150 and 2.50 / 2.02 / 5.53 at period 300. That is above habituation
alone on surprise (+0.88 \ci{+0.27}{+1.48}, $p=0.03$; +0.90
\ci{+0.45}{+1.37}, $p=0.016$) and, at 150, on connection (+1.72,
$p=0.010$). It is below the same interruption with habituation by 0.4--0.7
on surprise and connection and 0.8--0.9 on coherence, differences that do
not reach significance at $n=10$ (permutation $p=0.07$--$0.34$; on coherence the
bootstrap intervals exclude zero). The two operations are closer to additive
than to interacting, and the interruption is the larger of the two: alone it
gets most of the way, and habituation adds a fluent stream between
injections. Two baselines say what habituation is not. A stronger penalty
(1.3 instead of 1.15) changes nothing (1.75 / 1.42 / 4.90; $\Delta$ +0.15
\ci{-0.25}{+0.60} on surprise). Allowing the end-of-text token, so that the
model may close its document and open another at a boundary of its own
choosing, gives more coherent text (5.65, +1.25 \ci{+0.12}{+2.25}) and no
more surprise or connection (1.52 / 1.27; $-0.08$ and $-0.05$). The model's own
document breaks are boundaries without a subject, and they do what a sham
break does below.

\begin{table}[t]
\centering\small
\begin{tabular}{@{}lccc@{\hspace{1.2em}}ccc@{\hspace{1.2em}}ccc@{}}
\toprule
& \multicolumn{3}{c}{Qwen3-30B-A3B} & \multicolumn{3}{c}{Qwen3-8B} & \multicolumn{3}{c}{OLMo-2-13B} \\
\cmidrule(lr){2-4}\cmidrule(lr){5-7}\cmidrule(lr){8-10}
arm & S & C & H & S & C & H & S & C & H \\
\midrule
bare & 0.45 & 0.30 & 2.52 & 0.43 & 0.40 & 3.88 & 1.28 & 0.93 & 3.03 \\
bare + habituation & 1.60 & 1.32 & 4.40 & 1.37 & 1.00 & 4.35 & 1.96 & 1.26 & 3.99 \\
habituation + interruption 150 & \textbf{3.02} & \textbf{3.68} & \textbf{6.12} & \textbf{2.76} & \textbf{3.17} & 5.15 & 2.77 & \textbf{2.87} & 4.05 \\
DREAM scaffold & 2.70 & 1.85 & 6.02 & 2.70 & 2.35 & \textbf{5.38} & \textbf{3.60} & 2.23 & \textbf{6.23} \\
\bottomrule
\end{tabular}
\caption{The ladder on three generator models from two families: judged
surprise (S), connection (C) and coherence (H), means over cells
(generated-only windows, Opus~5, $k=5$; ten premises per arm). Habituation
buys part of the surprise; the interruption buys the rest and most of the
additional connection; the scaffold adds nothing detectable over the plain
interruption on the Qwen models and, on OLMo, adds surprise and coherence
(its selective forgetting lifts the stream out of a web-boilerplate mode)
while giving back connection.}
\label{tab:ladder}
\end{table}

\paragraph{Self-copy: what the window judge cannot see.} Before going
further, a check that the document-level judgment
(Section~\ref{sec:document}) forced on us. A base model given the same four
subject-change sentences in rotation learns the rotation. After \emph{``That
night she dreamed of something else entirely:''} it tends to write what it
wrote the last time that sentence appeared, 600 or 1,200 tokens earlier. We
flag a judged window as \emph{copied} when at least half of its 12-token
shingles occur earlier in the same stream, and we note whether the source
lies within the 600 tokens the judge sees (Table~\ref{tab:gen-self}). Bare
generation is 67\% copied; these are its literal loops, all visible to the
judge, who scores them 0.03. Habituation alone is 27\% copied. The clock
arms with the fixed rotation are 65--80\% copied at periods 150 and 300, 38\%
at 600 and 12\% at 900, and at 300 the source of 62\% of all windows lies
outside the judge's context. The reset arms are 0\% copied and the scaffold
6\%. A copied window scores high on connection for the obvious reason, it
\emph{is} earlier text, and, when its source is out of view, on surprise
too. So we recompute the ladder on \emph{fresh} windows only. The
interruption still lifts the stream over habituation. At period 150, surprise
+1.19 \ci{+0.42}{+2.09} ($p=0.010$), connection +0.71 \ci{+0.36}{+1.09}
($p=0.008$), coherence +1.78. At 300, +1.25 \ci{+0.71}{+1.78} ($p=0.004$),
+0.71, +1.51. At 900, +1.35, +0.92, +1.75. The reset arm, which cannot copy,
adds +1.99 \ci{+1.63}{+2.34}, +1.98 \ci{+1.68}{+2.30} and +1.86 (all
$p\le0.004$; every premise on surprise and connection, nine of ten on coherence). We conclude that the surprise effect of the
interruption is real on generated, fresh text and only slightly smaller than
on all windows (+1.19 against +1.42 at period 150; +1.25 against +1.30 at
300). The connection effect is largely a self-copy artefact (+2.37 on all
windows, +0.71 on fresh ones), and the reset arm's advantage over the
preserved context on connection (+1.27 \ci{+0.80}{+1.72} on fresh windows)
is in good part that it does not replay itself. From here on we report both
numbers where they differ and treat the fresh-only ones as the primary
estimate. The practical lesson is plain: a fixed rotation of
four sentences is a cycle the model will close. Give it a new sentence each
time, or reset.

\begin{figure}[t]
\centering
\includegraphics[width=\linewidth]{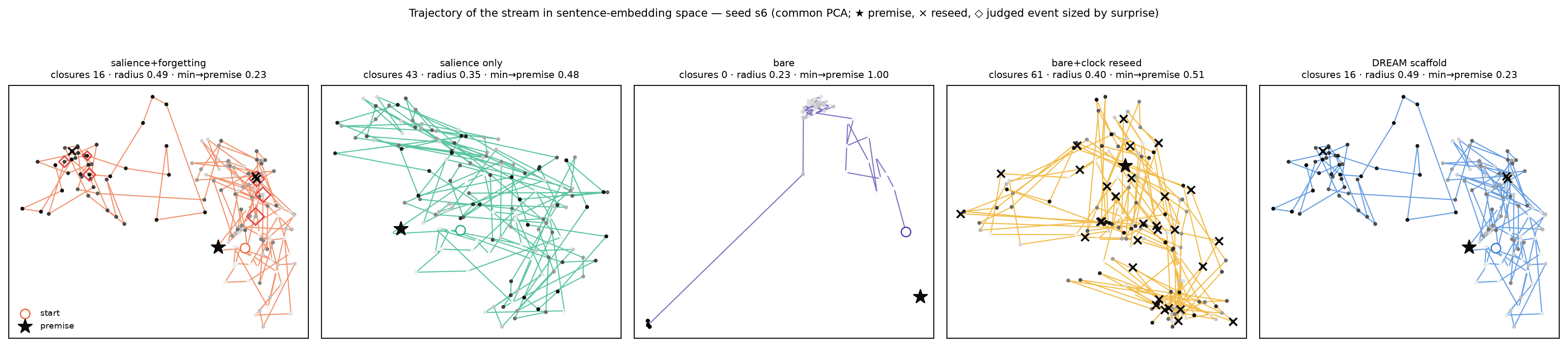}
\caption{Where the stream goes: trajectories in sentence-embedding space
(64-token windows, one PCA per seed), seed 6. $\star$ premise, $\circ$ start,
$\times$ reseed, $\diamond$ judged event sized by surprise. Bare makes one
jump and freezes; the scaffold orbits and returns; the clock reseed jumps
across the space keeping its context.}
\label{fig:traj}
\end{figure}

\paragraph{Where the stream goes.} In sentence-embedding space
(Figure~\ref{fig:traj}) bare generation has explored radius 0.18 and mean
step 0.14 and ends 0.78 from the premise. The scaffold has radius 0.51 and
returns to within 0.57 of the premise. The clock reseed spreads over the
space in jumps while keeping the whole context; the reset arm of battery~3
(Section~\ref{sec:battery3}) will show that keeping it is not what the judge
rewards.

\subsection{The interruption must lead away}
\label{sec:content}

\begin{figure}[t]
\centering
\includegraphics[width=\linewidth]{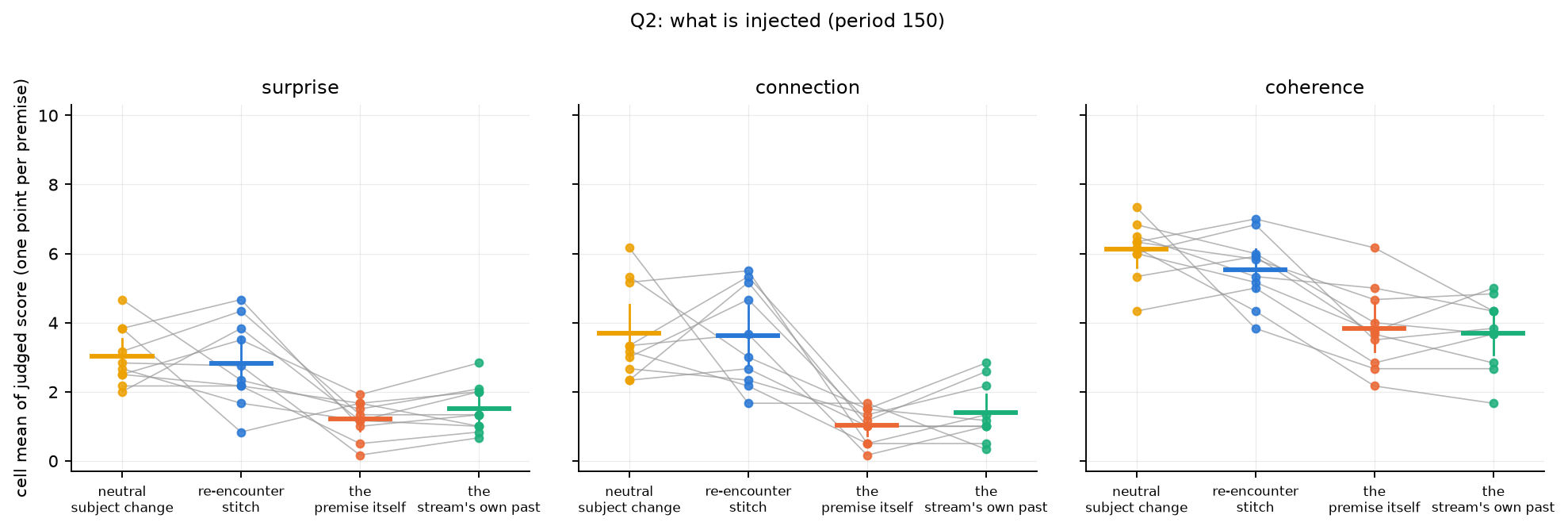}
\caption{Content of the interruption at period 150: neutral change,
re-encounter stitch, the premise itself, the stream's own past. Injecting a
return is as bad as not interrupting.}
\label{fig:q2}
\end{figure}

With the same period (150), a neutral subject change and a re-encounter
stitch that asks the stream to return to its beginning are indistinguishable
(3.02 / 3.68 / 6.12 vs 2.83 / 3.62 / 5.53; every paired CI covers zero).
Injecting the \emph{premise itself} (1.21 / 1.03 / 3.84) or a 64-token
window of the stream's \emph{own past} (1.51 / 1.39 / 3.68) is as bad as not
interrupting at all. Against the neutral change the premise loses $-1.81$
\ci{-2.33}{-1.28} of surprise and the own past $-1.51$ \ci{-2.10}{-0.87};
connection $-2.65$ and $-2.29$; coherence $-2.27$ and $-2.43$; all
$p\le0.004$, $\delta$ from $-0.85$ to $-1.00$ (Figure~\ref{fig:q2}).
Qualitatively, the own-past injection reinforces whatever mode the stream is
in: an exam-key stream was fed its own questions back. A return works only
as a short stitch that still opens a new sentence; a literal return closes
the loop.

\subsection{Timing: salience against the clock}
\label{sec:timing}

\begin{figure}[t]
\centering
\includegraphics[width=\linewidth]{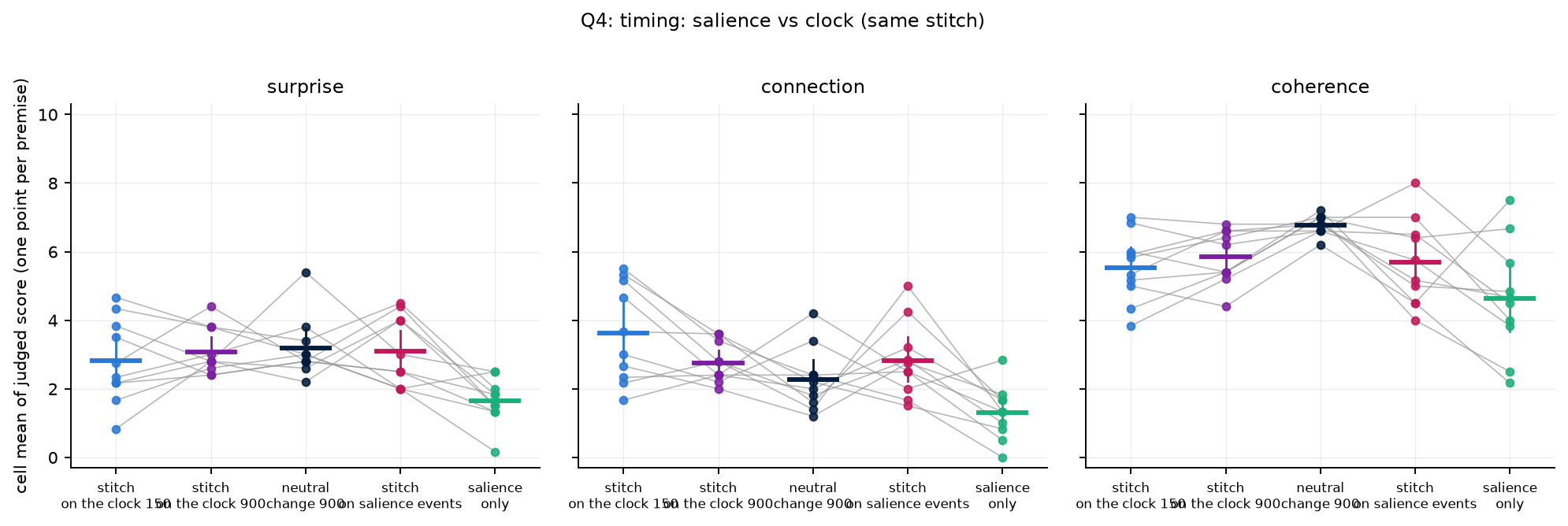}
\caption{When to interrupt: the re-encounter stitch on the clock (150 and
900), the neutral change every 900, the stitch on salience events, and
salience only (nothing injected). Post-interruption windows; one point per
premise.}
\label{fig:q4}
\end{figure}

Salience events (jump, crystallization, recurrence; 1--9 per cell, median
about 5) are the moments the scaffold's monitor picks out of the stream's own
telemetry. As a policy for \emph{which} windows to judge they beat a clock in
the first version of this study. The question here is whether they are also
the right moments to \emph{interrupt}. With the same stitch, timing by
salience events and timing by a clock of matched frequency (every 900
tokens) are indistinguishable on generated text. The post-interruption
windows score 3.09 / 2.82 / 5.68 against 3.08 / 2.76 / 5.84 (paired
differences $+0.01$ \ci{-0.77}{+0.77} on surprise, $+0.06$ on connection).
The windows at least 300 tokens after the last injection, where the stream
is left alone, score 1.85 / 1.80 / 4.18 against 2.10 / 1.73 / 4.78 ($-0.25$
\ci{-0.80}{+0.25}) (Figure~\ref{fig:q4}, Table~\ref{tab:gen-q4b}). Salience
monitoring by itself, with nothing injected, leaves the stream where
habituation alone leaves it (1.65 / 1.30 / 4.63 vs 1.60 / 1.32 / 4.40). The
first version of this study reported a large stream-level advantage of the
clock over salience timing (4.7 vs 2.2 on uniform windows). That advantage
was carried by the injected sentences inside the clock arm's windows, of
which it has more. What remains is a null at the present resolution. Salience
is a good reader of the stream and, as a metronome, neither better nor worse
than a clock ticking at the same rate. Since a clock is free and regular,
the clock is what we recommend.

\subsection{Period and decay}
\label{sec:period}

\begin{figure}[t]
\centering
\includegraphics[width=\linewidth]{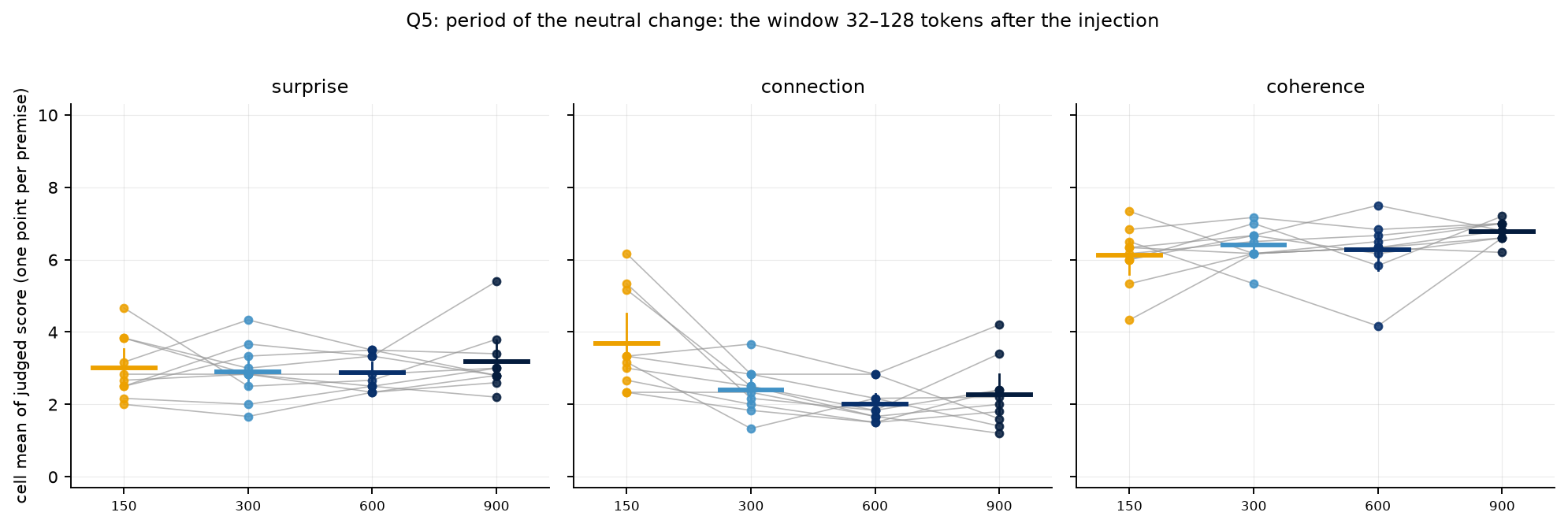}
\caption{Period of the neutral change: the post-interruption window (32--128
generated tokens after the injection) at periods 150, 300, 600 and 900; one
point per premise.}
\label{fig:q5}
\end{figure}

\begin{figure}[t]
\centering
\includegraphics[width=0.7\linewidth]{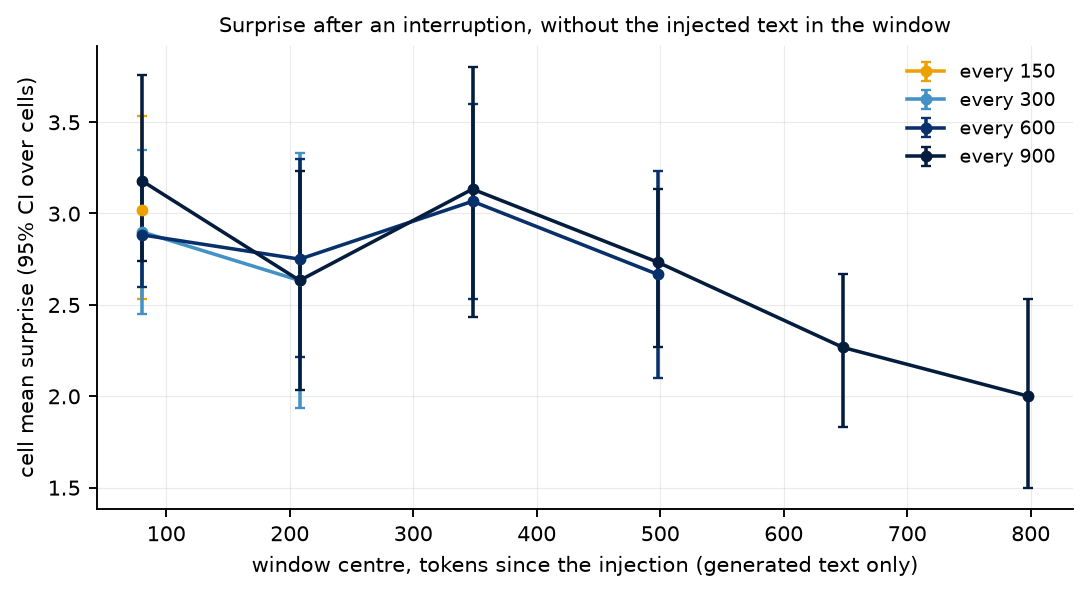}
\caption{Decay: cell-mean surprise of generated-only windows by tokens since
the injection, per period.}
\label{fig:q5b}
\end{figure}

The first version of this study reported that the yield of an interruption
grew with the length of the thread it broke (5.3--5.9 after 300--900-token
threads against 3.1 after 150), that it decayed over the next few hundred
tokens, and that 300 tokens was the best rhythm. With the injected sentence
excluded from the judged window, that finding largely disappears
(Figure~\ref{fig:q5}). The post-interruption window (32--128 generated
tokens after the injection) scores 3.02 on surprise at period 150, 2.90 at
300, 2.88 at 600 and 3.18 at 900, and every paired difference against 150
covers zero. Connection is highest at the shortest period on all windows
(3.68 vs 2.40, 2.00 and 2.26; $\Delta$ $-1.28$ to $-1.68$, $p\le0.008$), but
that difference is self-copy: on fresh windows connection is 2.0 to 2.3 at
every period (Table~\ref{tab:gen-self}). Coherence rises a little with the
period (6.12, then 6.40, 6.27, 6.78). Deeper into a segment the surprise of
generated text holds for a few hundred tokens and then sinks toward the
level of habituation alone (1.60): at period 900 it reads 3.18, 2.63, 3.13,
2.73, 2.27 and 2.00 at 32, 160, 300, 450, 600 and 750 tokens after the
injection (Figure~\ref{fig:q5b}). The stream estimate (offset means weighted
by the stretch of the segment they represent) is therefore 3.02 / 3.68 at
150, 2.78 / 2.38 at 300, 2.84 / 2.12 at 600 and 2.66 / 2.09 at 900
(Table~\ref{tab:gen-q5b}). Among the tested periods, then, none is better
than another on fresh generated text; what the longer periods buy is a
little coherence and fewer replays. A period of 75 cannot be measured under
this protocol, because no 96-token window of generated text fits between two
injections. Under the event protocol its stream was dead (0.88), which is
the one part of the earlier claim that survives: a thread needs some tens
of tokens to exist before it can be broken. The larger numbers of the first
version were, in good part, the judge reading the injected sentence. What
survives is the direction of every effect and an operating range of a break
every 150--300 tokens.

\subsection{What carries the effect: boundary, content, context}
\label{sec:battery3}

\begin{figure}[t]
\centering
\includegraphics[width=\linewidth]{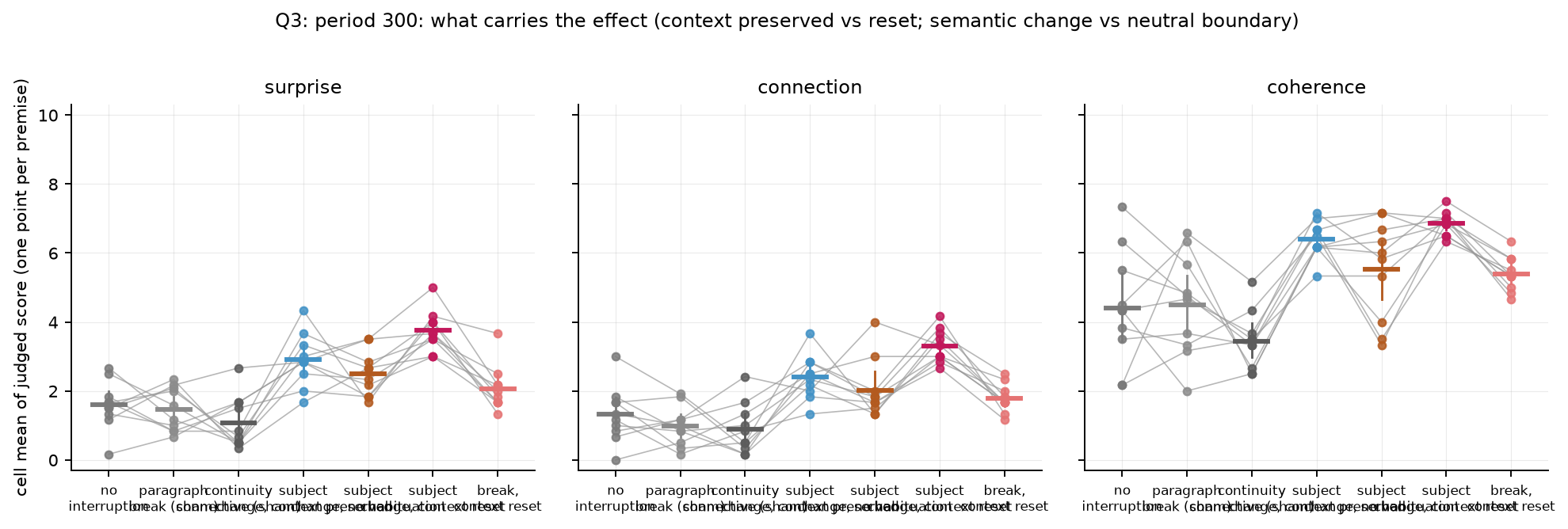}
\caption{What carries the effect at period 300: no interruption, a paragraph
break (sham), a continuity connective (sham), the subject change on the
preserved context, the subject change without habituation, the subject
change on a reset context, and a break on a reset context; one point per
premise.}
\label{fig:q3}
\end{figure}

An interruption is three things at once: a boundary in the text, a new
subject, and a context that is kept or dropped. Battery~3 separates them at
period 300 (Figure~\ref{fig:q3}, Tables~\ref{tab:gen-q3} and
\ref{tab:gen-q3-3}). A bare paragraph break every 300 tokens, a sham
interruption, leaves the stream where habituation alone leaves it (1.47 /
0.97 / 4.50 vs 1.60 / 1.32 / 4.40). A connective that asks for continuity
(\emph{``And so, as before,''}) is, if anything, worse than nothing (1.08 /
0.89 / 3.45; $-0.52$ \ci{-1.23}{+0.25} on surprise, $-0.95$ on coherence). A
boundary that points back is what the premise and own-past injections were.
The neutral subject change over the preserved context gives 2.90 / 2.40 /
6.40: +1.30 \ci{+0.68}{+1.92} on surprise ($p=0.004$), +1.08 on connection
($p=0.014$), +2.00 on coherence ($p=0.02$), and against the sham break +1.43
/ +1.42 / +1.90 (all $p\le0.006$). The same subject change on a \emph{reset}
context, where the model keeps only the premise and the injected sentence,
gives 3.75 / 3.30 / 6.87. That is higher than the preserved context on
every dimension (+0.85 \ci{+0.42}{+1.28}, $p=0.014$; +0.90 \ci{+0.53}{+1.25},
$p=0.004$; +0.47 \ci{+0.03}{+0.90}), and eight of ten premises move the same way on
surprise. Resetting without a new subject (a break on a reset context) gives
2.07 / 1.78 / 5.40, little more than the sham. The subject change against the
break is +1.68 / +1.52 / +1.47 under reset and +1.43 / +1.42 / +1.90 under
the preserved context ($p\le0.006$). We read this as follows. The new subject
is what produces the effect; the boundary by itself does nothing; and
dropping the earlier text from the model's context does not hurt what the
judge reads, it helps.

Two things must be said about that last result. Under reset the model cannot
integrate the earlier text. What it can still do is return to the premise,
sometimes verbatim, in a fresh mini-story, and the judge, seeing 600 tokens
of earlier text about the same premise, scores that as connected and
coherent. Under the preserved context the model can integrate, returns less,
and is entrained by 3,000 tokens of its own past. Connection as this rubric
measures it does not separate a long-range integration from a return to the
shared beginning. The preserved context is therefore a choice in the
operator we studied, not a demonstrated ingredient of the judged effect, and
Section~\ref{sec:document} asks the question at the level of the whole
document, where the difference between integration and restart should show
if it exists.

The same three arms replicate on Qwen3-8B-Base against its own habituation
cells. The sham break does nothing ($-0.05$ \ci{-0.77}{+0.80} on surprise).
The subject change on the preserved context adds +1.38 \ci{+0.74}{+1.98} of
surprise and +1.28 of connection, and on the reset context +2.40
\ci{+2.08}{+2.70} and +2.08, every premise (Table~\ref{tab:gen-reset}). On
OLMo-2-13B the sham again does nothing ($-0.04$). The preserved context adds
+1.64 \ci{+0.74}{+2.51} and +2.28 \ci{+1.58}{+2.95}, the reset context +2.14
\ci{+1.32}{+2.92} and +1.64 \ci{+1.10}{+2.25} (Table~\ref{tab:gen-reset-3}).
There the reset wins on surprise and coherence and the preserved context on
connection (3.54 vs 2.90). OLMo is the one generator on which keeping the
earlier text buys connection the judge can see, and the one whose
interruption arm copies itself least. The pattern that holds everywhere is
the one that matters: a boundary alone does nothing, a new subject does, and
a reset context is at least as good on surprise.

\subsection{A confirmatory replication on new premises}
\label{sec:confirm}

Everything above was designed and analyzed in sequence, and the questions of
batteries~2 and~3 were formulated after seeing battery~1. Before any result
of battery~3 or of the generated-only judgments was seen, we wrote ten new
premises (Appendix~\ref{app:repro}), fixed a different random seed and
chose five arms at period 300: habituation alone, the sham break, the
subject change on the preserved context, the subject change without
habituation, and the subject change on a reset context. We pre-registered
four contrasts in the laboratory notebook. H1 (primary): the subject change
on the preserved context beats habituation alone on surprise. H2: it beats
the sham break. H3: it beats the reset context on connection. H4
(two-sided): habituation matters given the interruption. Fifty new cells
were then generated and judged under the same protocol
(Table~\ref{tab:gen-confirmatory}, one-sided exact permutation tests). H1 is
confirmed: +1.48 \ci{+0.70}{+2.07}, $p=0.003$; on fresh windows only, +0.82
\ci{+0.03}{+1.44}, $p=0.035$. H2 is confirmed on all windows (+0.85
\ci{+0.20}{+1.45}, $p=0.02$) and not on fresh ones (+0.52, $p=0.20$). H3 is
refuted in the direction battery~3 had found: the reset context scores
\emph{higher} on connection ($-0.53$ \ci{-1.02}{-0.02} for preserved minus
reset; $-0.92$ on fresh windows) and on surprise (4.08 vs 2.83). H4 is not
supported (+0.43 \ci{-0.08}{+0.92}, $p=0.16$ two-sided). The document-level
judgment repeats on the new premises as well: the reset arm is above the
preserved context on all four document dimensions ($p\le0.03$), and the
preserved context at or below habituation alone
(Tables~\ref{tab:gen-confirmatory-5} and~\ref{tab:gen-confirmatory-6}). The
replication thus confirms the
effect the paper rests on, on premises and a random seed the program had
never seen. It also settles the two questions the exploratory batteries had
left open, in the less flattering direction: the context need not be kept,
and habituation is not what makes the interruption work.

\subsection{The whole document}
\label{sec:document}

\begin{figure}[t]
\centering
\includegraphics[width=\linewidth]{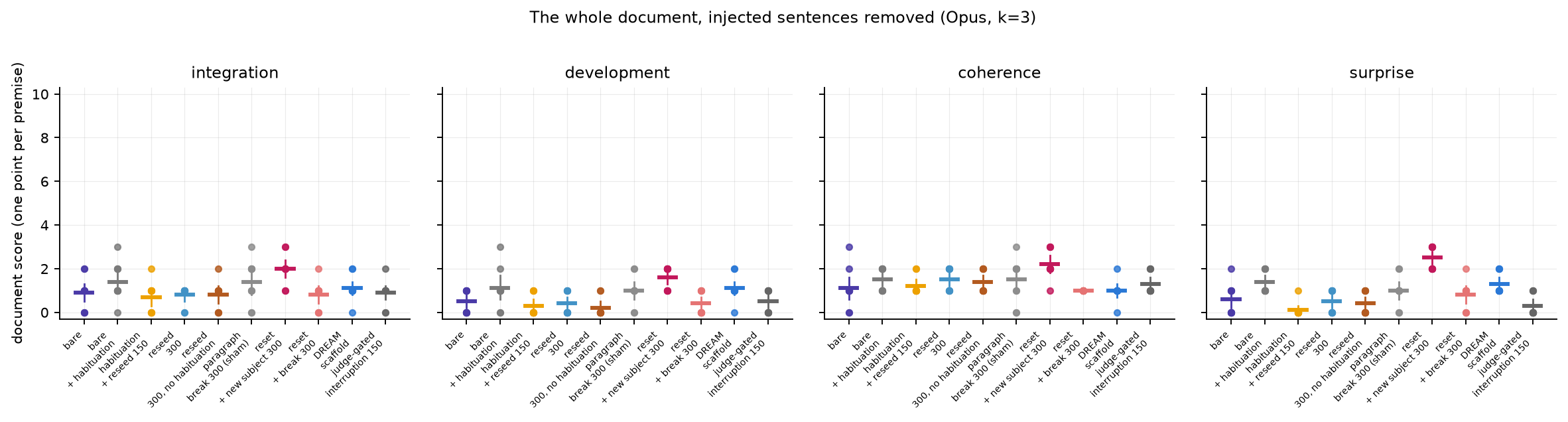}
\caption{The whole document, injected sentences removed: integration,
development, coherence and surprise of the 4,500-token stream (Opus~5,
$k=3$; one point per premise). No arm builds a whole; the interrupted stream
over the preserved context reads as a sequence of restarts; the reset arm's
clean parts, each returning to the premise, read best.}
\label{fig:document}
\end{figure}

Every score above is local: a window of 96 tokens read against 600. Does the
interrupted loop build anything over 4,500 tokens? We had the judge read the
\emph{whole} stream of each cell, with the injected sentences removed, and
score four things once (Opus~5, $k=3$, median; one score per cell):
\emph{integration} (are characters, motifs or ideas from earlier parts taken
up later and joined), \emph{development} (does something build rather than
restart or repeat), \emph{coherence} (does it read as one text) and
\emph{surprise} (does the whole go somewhere unpredictable yet sensible).
The answer is no, for every arm (Figure~\ref{fig:document},
Table~\ref{tab:gen-document}): no condition exceeds 2.5 on any document
dimension. Worse for the operator: judged as a whole, the interrupted stream
over the preserved context is a sequence of restarts and reads \emph{below}
the uninterrupted habituated stream. At period 150 its document surprise is
0.10 against 1.40 ($-1.30$ \ci{-1.60}{-1.00}, $p=0.002$) and its development
0.30 against 1.10; at period 300, surprise $-0.90$ ($p=0.03$) and
integration $-0.60$; the scaffold is at the level of habituation. The
judge's notes say why: the interrupted stream ``repeats the same four-part
block verbatim'' and ``cycles four short fragments'', the self-copy of
Section~\ref{sec:ablation} seen whole. The longer periods and the stitch
arms, with fewer replays and longer stretches, sit between (600 and 900:
0.9--1.7 on every dimension). The reset arm reads best of all (integration
2.00, development 1.60, coherence 2.20, surprise 2.50; above the preserved
context on all four, $p\le0.03$, and above habituation on surprise, +1.10,
$p=0.008$), because each of its parts is a clean piece that returns to the
premise, which the document judge, like the window judge, credits as a kind
of unity. So the local gains of Sections~\ref{sec:ablation}
to~\ref{sec:battery3} do not compose. An interruption is a local operation
that makes the next hundred tokens more surprising, connected and coherent
than they would have been, and leaves the document a collection. Keeping
the earlier text in the model's context, which is what a writer would want,
produces no integration that either judge can see, and the accumulation the
architecture was built for does not happen. We take this as the clearest
statement of what the paper has and has not found: an operator on windows,
not on wholes.

\subsection{The Review, with a gate that opens}
\label{sec:gate}

The scaffold's one instrument for building on a find, the judge-gated
escalation, never fired (Section~\ref{sec:loop-method}). To test the idea
rather than the accident, we ran the Review as the architecture intended it,
with a gate that does open. Every 150 tokens, before the scheduled subject
change, Opus~5 reads the last 128 generated tokens against the 600 before
them (one call). If the window is a find (surprise $\ge5$ and coherence
$\ge5$), the interruption is withheld and the thread is left to run;
otherwise the neutral subject change is injected as in the clock arm. Two
hypotheses were written before the run. G1: the judge-gated arm beats the
clock at 150 on the post-interruption window. G2: it beats it at the
document level on development and integration, since finds are allowed to
develop. The gate opened on 40 of 300 reads (13\%), so the arm interrupted a
little less often than the clock and let 40 good threads run. Neither
hypothesis is supported (Table~\ref{tab:gen-judge-gated}). On the
post-interruption windows the gated arm is at 2.67 / 3.03 / 5.93 against the
clock's 3.02 / 3.68 / 6.12 ($-0.35$ \ci{-0.95}{+0.12} on surprise, $-0.65$
\ci{-1.47}{+0.05} on connection). At the document level it is where the
clock is (0.90 / 0.50 / 1.30 / 0.30 against 0.70 / 0.30 / 1.20 / 0.10; every
paired difference within $\pm0.2$), below habituation alone on surprise
($-1.10$ \ci{-1.40}{-0.80}). A real judge deciding when not to interrupt,
on this generator and at this rate of finds, adds nothing to a clock,
locally or in the whole. The Review stage of the architecture is not refuted
as an idea for a different regime, but in this one it had its chance and
did not take it.

\subsection{Coda: the same operator on a problem with a verifier}
\label{sec:problem}

The document-level result says that the interruption is a variation
operator, not a builder. The natural next question is whether variation is
worth anything where value can be measured. As a first probe, run after
everything above and pre-registered in the notebook, we gave the main
generator a Python working notebook for online bin packing, the FunSearch
benchmark: a priority function scores the feasible bins, and fewer bins is
better. The notebook's docstring states the item-size distribution, first
fit and best fit are given as worked examples, and the model writes ideas
and \texttt{def priority} functions for 4,500 tokens. Every complete
function is extracted and scored by the verifier on held-out instances of
the same distribution. We ran ten distributions (the cells) $\times$ two
random seeds $\times$ four arms: plain (habituation, no interruption), a
distinct ``new angle'' comment injected every 300 tokens (fifteen different
ones, so that the self-copy cycle cannot form), the same angles on a reset
context, and an empty comment line every 300 tokens (sham). Pre-registered:
B1 (primary), the angle arm beats plain on the gain of its best candidate
over best fit; B2, it writes more distinct valid functions; B3, angle vs
reset on gain; B4, sham equals plain. Results are in
Appendix~\ref{app:problem}. B1 is \emph{not} supported ($-0.006$
\ci{-0.016}{+0.001} of gain, one-sided $p=0.94$). No arm beats best fit on
held-out instances in more than one of twenty streams, and the mean best
candidate sits just above it in every arm. B2 is supported: the interrupted
notebook writes 9.2 valid and 5.1 distinct valid functions per stream
against 2.0 and 1.8 for the plain one (+3.3 distinct, \ci{+0.8}{+6.7},
$p=0.012$). The sham writes fewer than plain and its best is worse
($-0.016$, $p=0.016$), so B4 is refuted in the direction of harm, as in the
narrative batteries. The reset arm has the most distinct valid functions
(6.6) and the best candidates: it reaches best-fit quality in 15 of 20
streams, +0.007 of gain over plain ($p=0.03$), and it is above the angle arm
on gain (B3, $p=0.04$). The picture is the one the document judge had drawn.
Interruption multiplies valid, distinct attempts three- to fourfold and does
not raise the ceiling of the best one; in a single stream, variation without
selection is organized noise. Whether the same operator inside a selection
loop, a FunSearch-style evolution over the candidates it multiplies, turns
variation into value is the next experiment, not this one.

\subsection{Three generator models from two families}
\label{sec:families}

\begin{figure}[t]
\centering
\includegraphics[width=\linewidth]{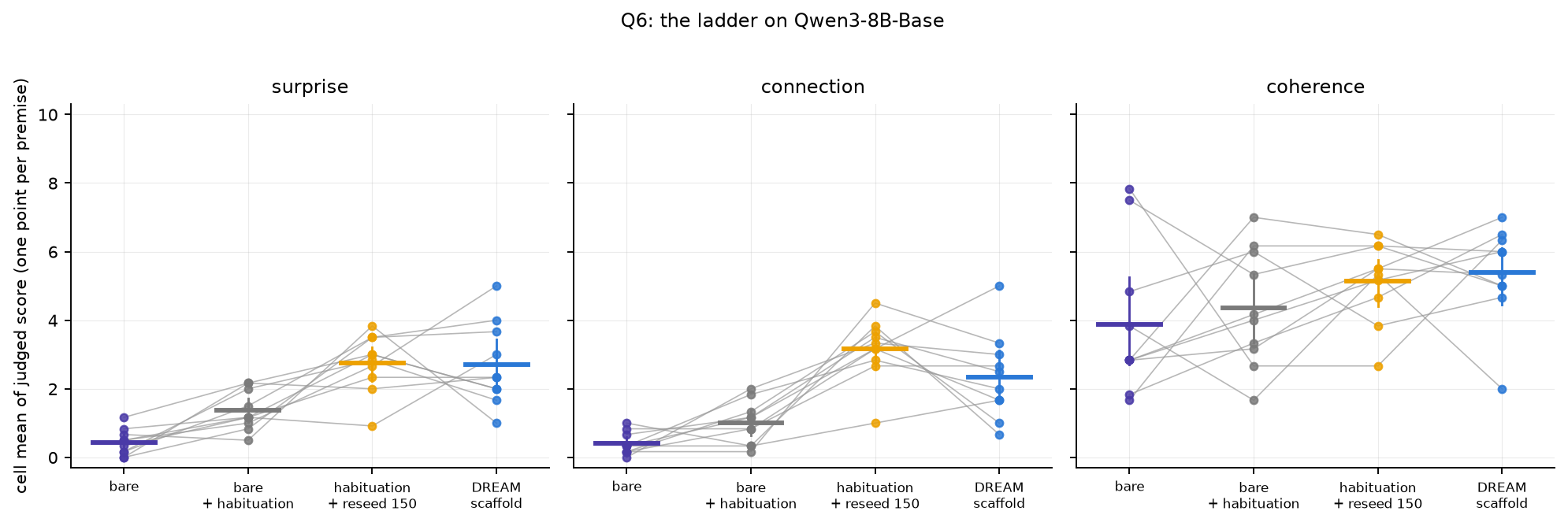}
\includegraphics[width=\linewidth]{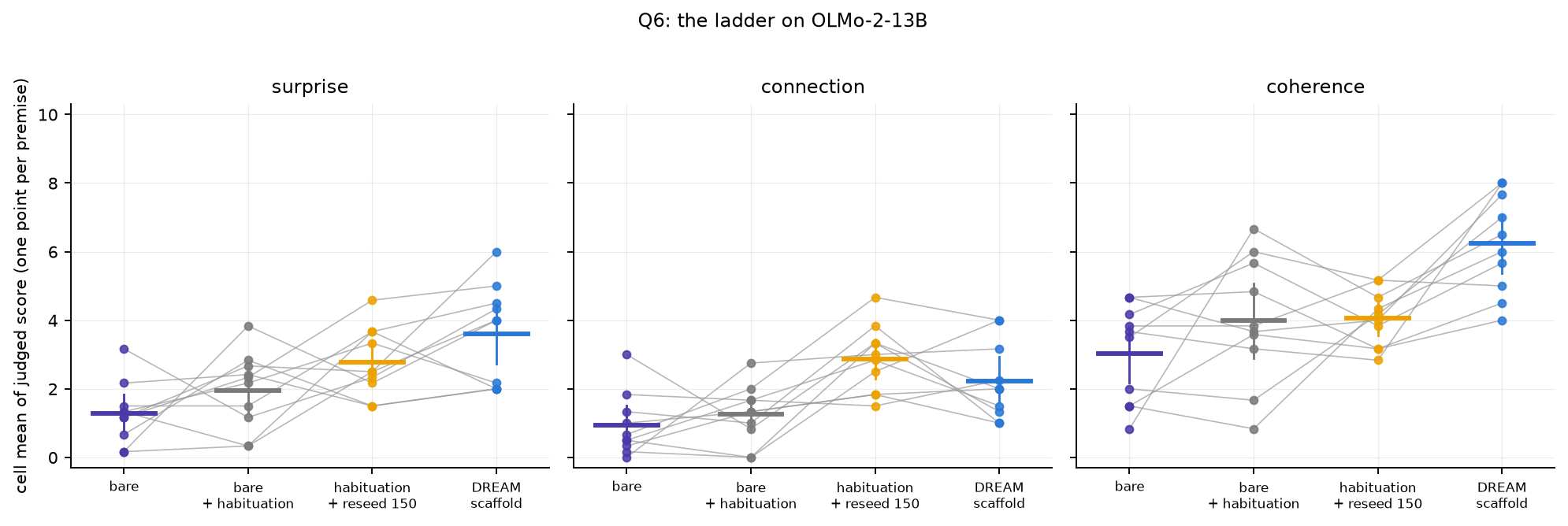}
\caption{The ladder on Qwen3-8B-Base (top) and OLMo-2-13B (bottom); one
point per premise.}
\label{fig:families}
\end{figure}

The ladder replicates on Qwen3-8B-Base and OLMo-2-13B (Table~\ref{tab:ladder},
Figure~\ref{fig:families}). On the 8B it reads 0.43, 1.37, 2.76 on surprise
(interruption over habituation +1.39 \ci{+0.72}{+2.07}, $p=0.008$;
connection +2.17 \ci{+1.53}{+2.83}, $p=0.002$), with the scaffold at 2.70 /
2.35. OLMo's bare stream wanders across web genres rather than locking into
literal loops, and is therefore less dead (1.28). There the interruption over
habituation gains connection (+1.61 \ci{+0.83}{+2.35}, $p=0.004$) and only
some surprise (+0.82 \ci{-0.21}{+1.81}, $p=0.17$), and costs nothing
detectable in coherence. Self-copy is present there too: 78\% of the
interruption arm's windows on the 8B and 45\% on OLMo
(Table~\ref{tab:gen-self-2}); on fresh windows the interruption adds +0.95
(8B) and +1.0 (OLMo) of surprise over habituation. On fresh windows the
connection gain is much smaller on the 8B (+0.17, against +2.17 on all
windows; OLMo +1.24; cell means, Table~\ref{tab:gen-self-2}). On OLMo the
full scaffold is the best arm on surprise (3.60, +1.64 over habituation, $p=0.04$) and
coherence (6.23, +2.24, $p=0.02$), with less connection than the plain
interruption (2.23 vs 2.87). What the scaffold does for OLMo is what a reset
does: it pulls the stream out of a web-boilerplate mode, the mechanism the
calibration probes had suggested. On a generator whose degeneration mode is
boilerplate the reset earns its keep, and battery~3 says it costs nothing on
the one that drifts in prose. Quantization carries none of this. The same
ladder on the unquantized (bf16) Qwen3-8B-Base gives 0.62, 1.35, 2.66 on
surprise and 0.48, 1.03, 3.23 on connection (interruption over habituation
+1.31 \ci{+0.20}{+2.34} and +2.19 \ci{+1.28}{+3.07}), against 0.43, 1.37,
2.76 and 0.40, 1.00, 3.17 at 8 bits.

\paragraph{A second genre.} All premises above are narrative. Ten expository
openings written for the purpose (\emph{``The history of the umbrella is
mostly a history of people refusing to carry one.''}, \emph{``Salt was once
the reason cities existed where they do.''}; the full list is in the
repository), run on the main generator at period 300, give the same
picture. Habituation alone reads 1.35 / 0.95 / 4.48; the subject change on
the preserved context 2.95 / 2.42 / 6.52 (+1.60 \ci{+1.18}{+2.02} on
surprise, +1.47 on connection, $p=0.002$, every premise); on the reset
context 2.95 / 2.45 / 6.63 (+1.60, +1.50). Here the two contexts are
indistinguishable (Table~\ref{tab:gen-second}).

\paragraph{A post-trained model.} As one step outside the scope decision, we
ran the same ladder on the post-trained Qwen3-8B (the released chat model,
8-bit), continued raw, without a chat template, from the same premises. Its
degeneration modes are its own: multiple-choice exam items with answer keys,
and a chain-of-thought voice that treats the premise as a riddle
(\emph{``Wait, let me think of a classic riddle\ldots''}). The ladder holds
at a lower level, 0.58, 0.75, 1.87 on surprise and 0.42, 0.75, 2.28 on
connection. Habituation does less for it than for the base models (+0.17 on
surprise), and the interruption still adds +1.12 \ci{+0.57}{+1.77}
($p=0.002$) and +1.53 \ci{+0.90}{+2.30} ($p=0.002$). Whether the operators
help such a model when it is used as intended, in a dialogue, is not tested
here.

\section{The instrument, measured}
\label{sec:instrument}

Every idea-level claim in this paper rests on an LLM judge, so we measured
the judge before using it, in three ways: its repeatability, its agreement
with a second judge family, and its agreement with human readers.
Repeatability is not validity. Five calls to the same model measure the
stability of one reader, not whether that reader is right; only the last two
checks bear on validity.

\begin{figure}[t]
\centering
\includegraphics[width=0.7\linewidth]{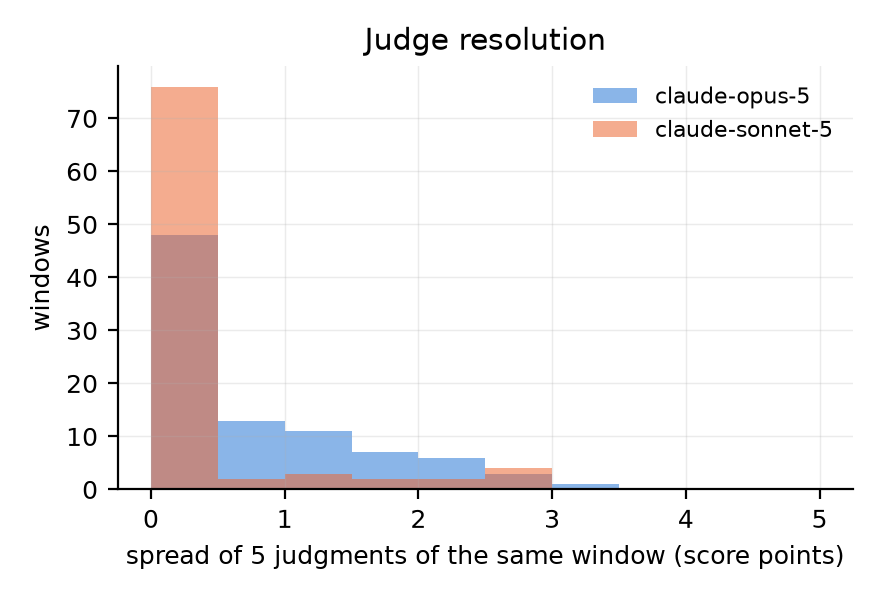}
\caption{Spread of five judgments of the same window (delta rubric, 89
windows). Sonnet~5 concentrates at zero spread by scoring zero; Opus~5 gives
continuous scores at $\pm0.7$ noise.}
\label{fig:judge}
\end{figure}

\paragraph{Repeatability.} On the same 89 windows judged $k=5$ times by two
models, Claude Opus~5 gives continuous scores with a median intra-window
spread of 0.71. Claude Sonnet~5 has spread 0.27, but by scoring zero on the
connection dimension almost everywhere; a ruler that only reads ``0'' is
consistent (Figure~\ref{fig:judge}). The three-dimension
surprise/connection/coherence rubric has spreads of 0.45--0.70 under Opus. A
test--retest of the whole pipeline came for free. Two arms of the ablation
battery, the full scaffold and its no-re-encounter ablation, turned out to
be byte-identical in all ten cells, because the in-loop judge never passed
and the re-encounter never fired, and they were judged independently. On
their 79 generated-only windows the medians of five agree exactly in 91\% of
windows on surprise (mean absolute difference 0.09), 85\% on connection
(0.15) and 71\% on coherence (0.29); the cell means differ by $-0.01$
\ci{-0.09}{+0.07} on surprise. Two consequences shape the paper. A binary
threshold on a noisy judge is a coin (the same window scored 5.24 one night
and 4.48 the next), so every result uses continuous medians of $k$ samples.
And a per-judgment noise of about $\pm0.7$ is one of several sources of
variance, not a threshold of detectability. The median of five reduces it,
windows within a stream vary more than that, cells vary more still, and the
power of a comparison is set by the ten paired cells, not by the judge's
spread. That is why the sampler's null inside the loop
(Appendix~\ref{sec:sampler}), obtained on windows and with a half-point
difference, is reported as ``no detectable effect at the present
resolution'', not ``absent''.

\paragraph{A second judge family.} To check that the loop results are not
the taste of one model family, a stratified sample of 140 already-judged
event-protocol windows (20 per condition across bare, habituation, clock
reseed, scaffold, re-encounter stitch, period 300 and salience-timed
re-encounter) was re-judged by Kimi K2.6 (Moonshot, via OpenRouter) with the
same rubric and $k=5$. Agreement with Opus~5 on the median of five is
Spearman $\rho=+0.85$ on surprise, $+0.77$ on connection and $+0.71$ on
coherence ($n=140$ windows from 70 cells; the correlations are computed on
windows and their $p$-values are not corrected for the clustering by cell).
Kimi is more generous (means about one point higher) and noisier
(intra-window spread 2.0--2.6), but every condition ordering replicates:
bare 0.55, habituation 3.15, clock reseed 4.30, scaffold 4.10, period 300
5.65 on surprise, with the neutral reseed and the re-encounter stitch
highest on connection under both judges. We repeated the check on the
generated-only protocol with a random, not stratified, sample of 160
post-interruption windows from eight conditions including the battery-3
arms. The agreement is $\rho=+0.75$ on surprise, $+0.80$ on connection and
$+0.65$ on coherence, and the ordering is again reproduced (under Kimi,
surprise runs bare 1.0, habituation 2.5, sham break 2.1, interruption 150
3.3, scaffold 4.9, interruption 300 5.4, reset 5.7). Two families agreeing
does not make the rubric valid; it rules out one family's taste as the whole
explanation.

\paragraph{Human rating, round 1.} Three independent raters, recruited on a
freelance platform without knowledge of the hypotheses or conditions
(Appendix~\ref{app:human}), rated 63 event-protocol windows blind: nine per
condition, stratified by judged surprise, shuffled, no labels, the earlier
text shown. They were a native English reader with professional editing
experience, a native or bilingual English speaker, and an advanced
non-native reader with graduate training in language. They rated surprise
and coherence (not connection) with the judge's rubric. All three returned
complete ratings using the whole of both scales, with distinct personal
calibrations (mean surprise from 3.4 to 6.1). Each rater alone agrees with
Opus~5 on surprise (Spearman $\rho$ from $+0.40$ to $+0.50$; all $p\le
10^{-3}$) and on coherence ($+0.23$ to $+0.60$). The mean of the three
agrees more strongly than any one (surprise $\rho=+0.58$, $p=7\times10^{-7}$;
coherence $+0.52$, $p=10^{-5}$), about as strongly as the raters agree with
each other (pairwise Spearman on surprise from 0.25 to 0.71, Krippendorff's
interval $\alpha=0.30$; on coherence 0.69, 0.60, 0.42, $\alpha=0.38$; the low
$\alpha$ reflects the different scale calibrations that rank correlation
ignores). The human consensus reproduces the condition ordering that carries
the paper. On surprise, means over nine windows: period-300 interruption 6.5,
neutral clock reseed 6.4, re-encounter stitch 5.3, salience-timed 5.0,
scaffold 3.6, habituation only 3.6, bare 1.5; interruption vs bare
\ci{+3.1}{+6.5}, habituation vs bare \ci{+0.1}{+3.9}, and the plain
interruption above the full scaffold by a wider margin than under Opus. On
coherence the human readers rank habituation without interruption highest
(7.7) and the interrupted arms 5.7--6.3: to a human reader, the subject
change inside a window costs some fluency that the LLM judge does not
penalize. Three limits of this round are plain. The sample was stratified by
the judge's own surprise score, which widens the range and can inflate a
rank correlation. Connection, the dimension on which the interruption's
all-windows advantage was largest, was not rated. And inter-rater agreement
is low to moderate, so the human data support the ordering of conditions and
a moderate agreement with the judge, not the judge's exact numbers. A second
round, on 56 generated-only, fresh windows from eight conditions with
connection as a third dimension, was prepared but not run; the human data
bear on the first protocol only.

\section{Inside the network: a descriptive look}
\label{sec:network}

Judges and sentence embeddings see the text; the model's own residual stream
sees the computation. This section is exploratory and descriptive. It
reports correlational geometry of the residual stream with judged windows as
observations (clustered by cell; for the pooled correlations we give
cluster-robust intervals from a bootstrap over cells, and the
within-condition values are descriptive), it uses a logit lens for one
coarse quantity, and it makes no causal intervention. It characterizes what
the judged dimensions co-vary with in the model's state. It does not
establish a mechanism. We re-run every finished stream through its generator
(one forward pass with a KV cache, exact token positions) and capture the
residual at 13 layers: every 4th of 48 for Qwen3-30B-A3B, every 3rd for the
36-layer Qwen3-8B and the 40-layer OLMo-2. The residual is mean-pooled over
64-token windows (stride 32) and mean-centered per layer before cosine
geometry; a logit lens (the final norm and unembedding applied to the
intermediate state) is read at each captured layer. Three questions.

\begin{figure}[t]
\centering
\includegraphics[width=\linewidth]{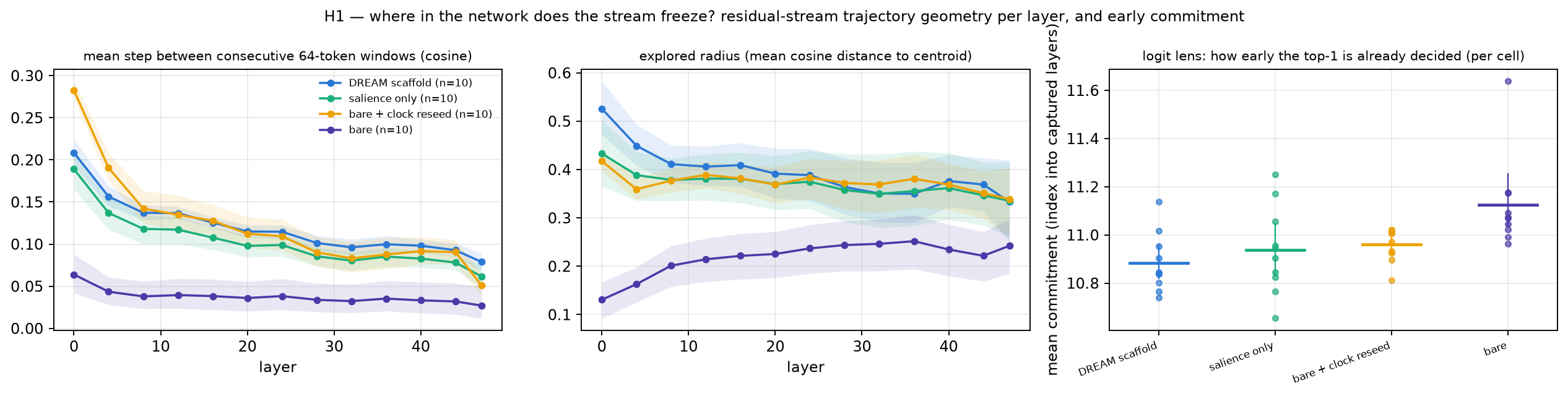}
\caption{N1: residual-stream trajectory geometry per layer and condition
(Qwen3-30B-A3B, ablation battery): mean step, explored radius, and logit-lens
commitment. Bare generation moves least at every sampled layer, most so at the surface.}
\label{fig:n1}
\end{figure}

\paragraph{N1: where does the stream freeze?} We compute per-layer
trajectory geometry of the window vectors (mean step between consecutive
windows, explored radius) by condition, and the logit-lens \emph{commitment
layer}, the captured layer from which the top-1 token no longer changes.
Bare generation's mean step is a third to a quarter of every interrupted
condition's at every sampled layer. Its explored radius is 0.13 at the input
layers against 0.42--0.53, and the gap narrows with depth but never closes
(0.24 vs 0.33--0.34 at layer 47, CIs disjoint; Figure~\ref{fig:n1}). The
three interrupted conditions are geometrically alike inside the network;
what separates them is what the judge reads. On the 30B, bare generation's
top-1 stabilizes \emph{later} (commitment index 11.12 \ci{11.04}{11.25} of 12
vs 10.88--10.96) while its final distribution is far more confident (final
entropy 0.34 vs 0.42--1.08), a pattern consistent with copying, a late-layer
computation that ends certain. The ordering does not replicate on the 8B, so
we report it as an observation. The 8B and OLMo replicate the geometry (bare
radius 0.14 and 0.36 at layer 0 against 0.46--0.61).

\begin{figure}[t]
\centering
\includegraphics[width=\linewidth]{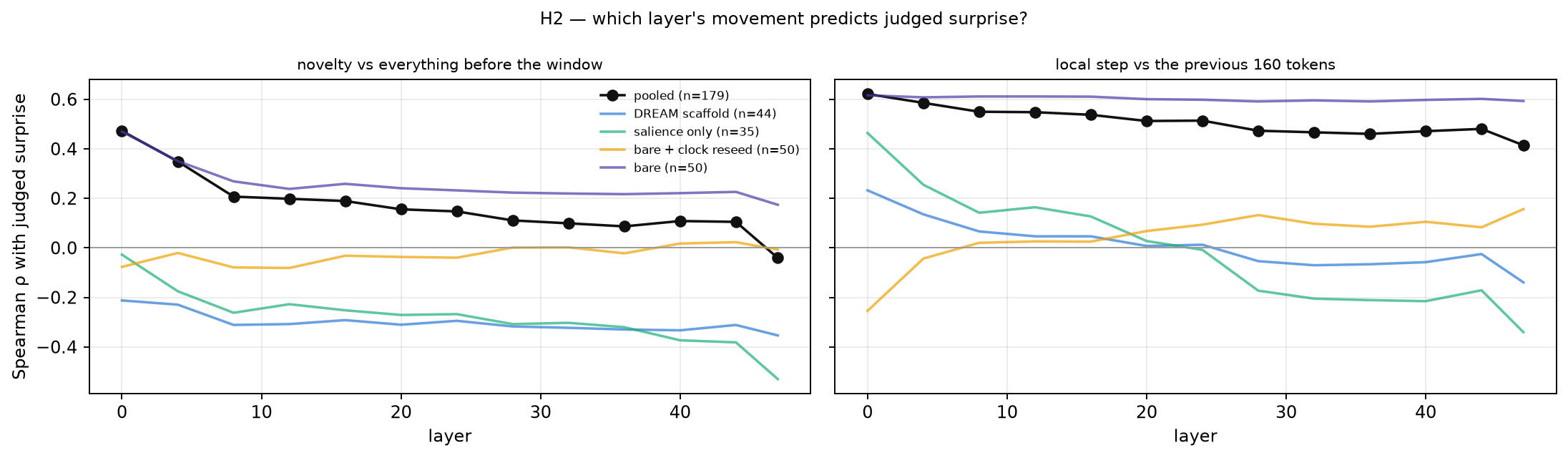}
\caption{N2: Spearman correlation between the judged window's movement
(novelty vs everything before it, left; local step, right) at each layer and
judged surprise, pooled and within condition.}
\label{fig:n2}
\end{figure}

\paragraph{N2: which layer's movement predicts judged surprise?} For every
judged window we compute its novelty at layer $l$ (cosine distance between
the window's mean state and the mean state of everything before it) and its
local step (against the previous 160 tokens), and correlate both with judged
surprise (Figure~\ref{fig:n2}). Pooled over conditions, novelty against the
past correlates with surprise at the input layers ($\rho=+0.47$ at layer 0;
cell-bootstrap 95\% CI \ci{+0.22}{+0.67}) and not at the top ($-0.04$,
\ci{-0.26}{+0.17}), but the pooled number is carried by the bare/interrupted
contrast. Within interrupted conditions the sign flips with depth. Local
step at layer 0 correlates positively with surprise (salience-only $+0.46$,
scaffold $+0.23$) and at layer 47 negatively ($-0.34$, $-0.14$); novelty
against the whole past is negative at depth (salience-only $-0.53$ at layer
47). Across the 668 judged windows of battery~2, layer-0 movement predicts
surprise within every condition ($\rho$ +0.3 to +0.6), and higher final
entropy predicts it everywhere ($\rho$ +0.3 to +0.6): confident is
unsurprising. In these data, judged surprise co-varies with departure at the
surface layers and with continuity at depth, new at the surface and
continuous underneath.

\begin{figure}[t]
\centering
\includegraphics[width=\linewidth]{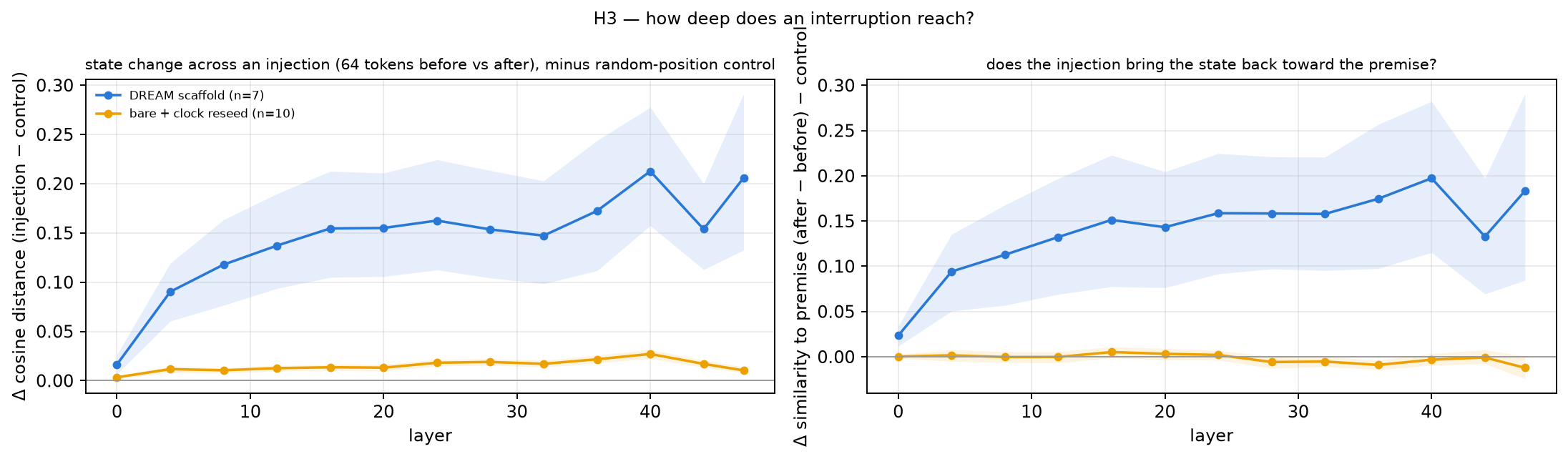}
\caption{N3: how deep an interruption reaches (Qwen3-30B-A3B). Left: state
change across an injection per layer, minus random-position control. Right:
change in similarity to the premise's own state. The forgetting reseed moves
the deep state and returns it toward the premise; the clock reseed over
preserved context barely touches it, and scores as high on the judged
dimensions.}
\label{fig:n3}
\end{figure}

\paragraph{N3: how deep does an interruption reach?} We measure the cosine
distance between the 64 tokens before an injection and the 64 after it (the
injected text skipped), per layer, minus the same at random positions, and
the change in similarity to the premise's own state (Figure~\ref{fig:n3}).
Across a scaffold reseed \emph{with forgetting}, the state after the
injection differs from the state before by +0.09 (layer 4) growing to +0.21
(layer 40) beyond control, and its similarity to the premise's state rises by
+0.09 to +0.20 with depth. Forgetting moves the deep state toward the
premise's own state, a re-encounter with the beginning as the network
represents it. Across a clock reseed \emph{over preserved context} the same
measures are +0.01--0.03 and about 0 at every sampled layer, and this arm
scores as high as any on the judged dimensions. In this reading, judged
surprise and connection do not require deep representational shifts; they
accompany surface departures over a deep state that is left nearly intact.
We take this as a description of what the judged dimensions track, not as an
account of why an arm scores as it does. The reset arm of battery~3, which by
construction moves the deep state at every interruption and scores as high
as the preserved context, was not captured and is the obvious next probe.
The dissociation replicates on the 8B (+0.03 to +0.10 vs +0.01--0.03) and,
larger, on OLMo (+0.20 to +0.38 with a return of +0.13 to +0.18, vs
+0.02--0.06 with no return). In battery 2, the depth of the state change
scales with the length of the interrupted segment (period 75: about 0; 150:
+0.01--0.02; 300: +0.02--0.04; 600: +0.04--0.075) and not with what is
injected, a dependence on segment length that the judged scores of the
generated text do not show (Section~\ref{sec:period}). The salience-timed
stitch is the one 150-scale injection that reliably moves the deep state and
pulls it toward the premise, yet on generated text it scores no higher than
the same stitch on a clock of matched frequency (Section~\ref{sec:timing});
deep movement does not buy judged quality here either. Along
the stream, similarity to the premise state is U-shaped in depth (about 0.9
at layer 0, about 0.5 mid-network, rising at the top) and ordered at the top
layer: scaffold 0.85, salience-only 0.79, clock reseed 0.67, bare 0.55. The
sentence-embedding ``return to the premise'' of the scaffold is visible at
the top layer and not in the middle of the network.

\section{Discussion}
\label{sec:discussion}

\paragraph{The scaffold and the beam.} We started from an architecture
modeled on the mind (three networks, incubation, re-encounter), and
measurement returned something simpler than the architecture. The scaffold
works, but when it is taken apart most of its effect lives in two
operations that were not the noble ones: not repeating one's literal past,
and being made, every few hundred tokens, to start a new sentence somewhere
else. We call this minimal operator the \emph{interrupted loop}. Whether the
earlier text is kept in the model's context while it does so turned out not
to matter to the judge (Section~\ref{sec:battery3}); we kept it, and it is a
choice, not an ingredient. The elaborate parts fall into four groups. Some
add nothing detectable: the salience monitor as a trigger, and the in-loop
judge, which never fired. Some interrupt too rarely: the forgetting reseed
fires 0--3 times per cell, and the scaffold's lower connection tracks that
rarity rather than the forgetting, since a complete reset every 300 tokens
scores at least as well as the preserved context. Some work opposite to
their design: a literal return to the premise closes the loop. And one does
no better than the trivial alternative: salience is a good reader of the
stream and, as a metronome, no better than a clock at the same rate. Only
because the whole scaffold was built and then dismantled do we know which
beam carries the weight, and which pieces are conditional. The reset earns
its keep on a generator whose degeneration mode is web boilerplate, and is
indifferent on one that drifts in prose.

\paragraph{What an interruption is.} Battery~3 takes the interruption apart
as the earlier batteries took the scaffold apart. A boundary is not an
interruption. A paragraph break every 300 tokens leaves the stream where
habituation leaves it, and the model's own document breaks (EOS allowed) buy
coherence and nothing else. A boundary that points back is worse than none.
What works is a new subject. Habituation adds to it, a fluent stream between
injections, but the new subject alone gets most of the way. A new subject
also means new each time. The four-sentence rotation we used is a cycle the
model closes by replaying its own earlier segments, and a window judge with
a 600-token horizon reads the replay as surprise and, worse, as connection.
The fresh-only numbers are the ones to believe, and a reset context, or a
fresh sentence at each break, is the cure. The new subject works whether the
earlier text stays in the model's context or is thrown away, at least for
what the judge can read, and thrown away it scores higher. On a reset
context the model writes a fresh mini-story that returns to the premise, and
the judge scores it as connected and coherent; on the preserved context the
model can integrate, returns less, and is entrained by thousands of tokens
of its own past. The operator is therefore smaller than we first wrote it:
damp repetition, and every few hundred tokens give the model a new sentence
that leads away. Keeping the memory is what a writer would want and what a
600-token judge cannot reward. The difference between integration and
return is exactly the kind of long-range quality this instrument does not
see, and it is where a human study with the whole document in view should
look next.

\paragraph{Windows and wholes.} The document-level judgment
(Section~\ref{sec:document}) is the result we would least like to have and
are most obliged to report. None of these streams builds anything, and the
interrupted stream over the preserved context, whose windows are the best,
reads as a whole below the uninterrupted habituated one. The interruption is
an operator on the next hundred tokens. The architecture we started from was
built for accumulation (finds kept, developed, returned to), and its one
instrument for that, the judge-gated escalation, never fired in the
scaffold. Run on purpose, with a real judge and a gate that opened on 13\% of
its reads (Section~\ref{sec:gate}), it did no better than the clock at
either level. What would make the local gains compose is not a better
trigger; it is something this loop does not have. The coda on bin packing
(Section~\ref{sec:problem}) says what the operator is worth where value can
be measured: three to four times more valid, distinct attempts and no better
best. That is the profile of a variation operator, and variation is worth
something only under selection: a task with a verifier and a loop that keeps
a find because it is worth something, not because a judge liked the window.

\paragraph{Surface novelty and judged surprise are different quantities.}
The anti-probable sampler doubles $n$-gram novelty against the training
corpus and shows no detectable effect on judged surprise or connection. The
interruption lifts judged surprise over habituation alone by 1.2 to 1.4
points on fresh text and 1.3 to 1.4 on all generated-only windows (about 2.5
over bare generation) without touching the sampler at all. Inside
the network, read descriptively, the two live at different depths: what the
judge calls surprise co-varies with lexical departure over an intact deep
context, and the interruption that scores best barely moves the deep state.
This is also why the neat expectation ``novelty equals distance from what
came before'' fails within interrupted streams, where deep departure from
the stream's own past is negatively related to judged surprise. The reader
rewards the window that is new on the surface and continuous underneath,
which is what a good turn in a story is.

\paragraph{Rhythm, revised.} The first version of this study made much of a
rhythm: a break every 75 tokens develops nothing, the same break scores
higher after a longer thread, and 300 tokens looked like the best period. On
generated text only, most of that is gone. The post-interruption surprise
does not depend on the length of the thread it breaks; on fresh windows,
neither does connection; and a stream left alone for many hundreds of tokens
sinks back toward the level of habituation alone. What survives is humbler
and, we think, still usable. The interruption is a resource that has to be
spent regularly, every 150--300 tokens on these generators; a thread needs
some tens of tokens to exist before breaking it is worth anything; and
nothing is gained by waiting.

\paragraph{What this is not.} It is not a claim that these streams are good
literature. Most windows are judged well below the midpoint, and the texts
are what a base model produces under forced continuation. It is not a claim
about creativity in the full sense, since no value, interest or originality
against the world is measured, nor about scale, since the generators are
8--30B parameters, mostly at 8 bits, and only one post-trained model was
run, in raw continuation. It is a characterization, with controls, of a
simple intervention on forced open-ended generation, made on three base
models from two families and under two judge families, and it points at the
loop's structure rather than at the two places where effort is usually
spent.

\section{Limitations}
\label{sec:limitations}

\paragraph{What was measured.} The outcome is \emph{judged narrative
surprise, connection and coherence} on short windows of forced open-ended
continuation, read by an LLM judge and, on a subset, by human readers. That
is a defensible operationalization of one ingredient of creativity, the
appropriately unexpected turn. It is not creativity: no value, interest or
originality against the world is measured, and the study makes no claim
about ideas or problems. Its title names the question the program set out
with; the results answer a narrower one.

\paragraph{Design.} The premises are ten English narrative openings, plus
ten expository openings for one replication and ten new narrative premises
for the confirmatory battery. Three generator models from two families run
the core ladder; one generator (Qwen3-30B-A3B) runs the content, timing,
period and control batteries. Every arm ran once per premise with the same
random seed, so the ten cells are the whole sample and a paired comparison
of ten is the strongest inference available. Batteries 1 to 3 are
exploratory. They were designed in sequence, the questions of batteries~2
and~3 were formulated after seeing battery~1 and after an external review,
and the analysis reported here (generated-only windows, cell as unit) was
fixed before battery~3 was generated but after the earlier arms had been
seen under the first protocol. The pre-registered exception is the
replication of Section~\ref{sec:confirm}, run on new premises and a new
random seed with its contrasts written down in advance; the judge-gated arm
and the verifier probe were pre-registered in the same way. Everything else
should be read as a well-controlled exploratory study.

\paragraph{Instrument.} The judge is one model family. It is calibrated for
repeatability and agrees with a second family ($\rho=0.71$--$0.85$ on a
judge-stratified sample of first-protocol windows, $0.65$--$0.80$ on a random
sample of generated-only windows) and with the consensus of three human
readers ($\rho=0.58$ on surprise, $0.52$ on coherence, on a judge-stratified
sample), whose agreement with each other is low to moderate
($\alpha=0.30$--$0.38$). Human readers rated only surprise and coherence, on
63 first-protocol windows, and they penalize the fluency cost of a subject
change more than the judge does. The judge sees 600 tokens of
context, so ``connection'' means connection within that horizon. Text that
the stream copies from further back is invisible to it as a copy, and a
fixed rotation of injected sentences produces exactly such copies
(Section~\ref{sec:ablation}). We report fresh-only estimates for that
reason, and a judge with the whole document in view for the same reason.

\paragraph{Baselines and generality.} The bare arm is a deliberately hard
regime (forced continuation with end-of-text masked, no repetition control).
We report it as the floor of the ladder, not as a fair decoding baseline; the
fair comparisons are habituation alone, habituation with a stronger penalty,
EOS allowed, and the sham and reset controls. Stronger decoding baselines
(look-back or contrastive decoding, learned repetition control) were not
tested. Base models were used by a scope decision. The one post-trained
model we ran, in raw continuation, shows the same ladder at a lower level
and says nothing about the dialogue use such models are made for.

\paragraph{The network section.} The residual-stream analysis is
correlational and descriptive: mean-pooled 64-token windows at 13 sampled
layers, a logit lens for a coarse commitment layer, no attention analysis,
no tuned lens, no causal intervention. It characterizes what the judged
dimensions correlate with in the model's state. It does not establish a
mechanism.

\section{Conclusion}

Three findings carry this paper. First, the novelty a base language model
produces with no task comes, in the regime we studied, from the structure of
the loop and from one simple part of it: damp the model's literal repetition
and interrupt it every few hundred tokens with a sentence that leads
somewhere new, and its generated text is judged more surprising and
somewhat more connected on windows that contain none of the injected words
and, in the fresh-only estimates, none of the stream's own earlier text; a
second judge family reproduces the ordering of conditions, and human readers
reproduced it on surprise for windows of the first protocol. A
pre-registered replication on new premises
confirms the main contrast. The new subject is what does the work, with or
without the earlier text in the model's memory; a boundary alone does
nothing. Neither the sampler, at the resolution we could reach, nor the
prompt, under the operationalizations we tested, does any of this.

Second, the gains do not compose. Read as documents, these streams remain
collections; a Review that decides when not to interrupt does not change
that; and on a problem with a verifier the same interruption multiplies the
valid, distinct candidates a base model writes without improving the best
of them. The interruption is a variation operator. Whether that variation
becomes value inside a selection loop, where a find is kept because it is
worth something, is the question this study leaves open and the next one
asks.

Third, a windowed LLM judge of a long stream misses things that change the
answer. It scores the experimenter's injected sentence as the model's own,
it scores the model's replay of its own earlier segments as surprise and
connection when the source lies beyond its horizon, and it cannot tell a
local gain from a whole. Generated-only windows, fresh-only estimates, a
document-level reading, the premise as the unit and a measured judge are
what it took to see this, and we suspect they are what it takes in general.
What we have characterized is a controllable intervention on forced
open-ended generation and the instrument needed to characterize it honestly.

\paragraph{Acknowledgements and tooling.} The experiments, analyses and text of this
paper were produced by the author working with Claude (Anthropic) as a
programming and writing assistant inside the Claude Code environment, and the
corrections of version~2 were prepared the same way; every design decision,
result and claim was reviewed by the author. The code and the dated
laboratory notebook are public in the repository, and the run data and LLM
judgments are in the evidence deposit
(\href{https://doi.org/\evidencedoi}{doi:\evidencedoi}).
Judges were run on Amazon Bedrock and OpenRouter; generators ran locally on
Apple silicon via MLX.

\paragraph{Changes in version 2.} This version corrects version~1 (20 August
2026); no conclusion changes. Version~1 is the preprint; what the text calls
the first version of this study is the earlier, pre-review draft that used
the event-window protocol (\texttt{docs/PAPER\_DREAM.md} in the repository).
(i)~Status: the second round of human ratings, described in version~1 as in
progress, was prepared but not run. (ii)~Human validation: the human readers rated only surprise and coherence, on windows
of the first protocol, which could contain the injected sentence; the
introduction and the conclusion now state that scope, and the agreement with
the second judge family is given for both of its samples. (iii)~Self-copy:
the abstract ran together two measures, the share of post-interruption
windows that are replay (65--80\% at periods 150--300) and the share whose
source lies beyond the judge's horizon (62\% at period 300); they are now
separate. (iv)~Numbers: the main-text numbers of Section~\ref{sec:loop} and
of the appendix text now match the appendix tables. In version~1 some had been computed before the last re-judged
windows were added; most differences are below 0.1 points and none exceeds
0.3. With the final values the fresh-only effect on connection is +0.7 to
+0.9 (version~1: +0.8), the fresh-only effect on surprise is only slightly
smaller than the all-windows one (version~1: about one point smaller), and
the fresh-only connection gain on the other generators is now reported, three
counts of premises moving the same way are given as the final values show them
(for example eight of ten, not every premise, for reset against preserved context on
surprise), and differences on coherence once said to have intervals that include
zero are described as not significant at $n=10$. For rater privacy, the individual
raters' means and agreements are now given as ranges and the payment without
amounts. Three statements were corrected: the salience-timed stitch was called the
arm the judge rates lowest, which held only under the first protocol;
``two to three points'' of surprise on all windows holds against bare
generation, not against habituation; and the abstract said that a paragraph
break adds nothing detectable ``on fresh text'', for which no contrast is
reported (the all-windows contrast is).
(v)~Availability: version~1 said that the per-run data, judgments and rater
files were released with the repository; they were not. The run data and the
LLM judgments are now in the evidence deposit, and individual human ratings
are not released, to protect rater anonymity. A duplicated table caption and
a table reference that pointed to the wrong table were fixed.

\bibliographystyle{plainnat}
\bibliography{references}

\appendix
\section{Two motivating nulls: the sampler and the prompt}
\label{sec:sampler}

The program began where effort is usually spent, the sampling operator and
the input, and found nothing at the level of ideas in either place. Both
studies are small and were run before the loop experiments. They are kept
here as the motivation for them, with their limits stated; the full tables
are in the repository.

\begin{figure}[t]
\centering
\includegraphics[width=0.72\linewidth]{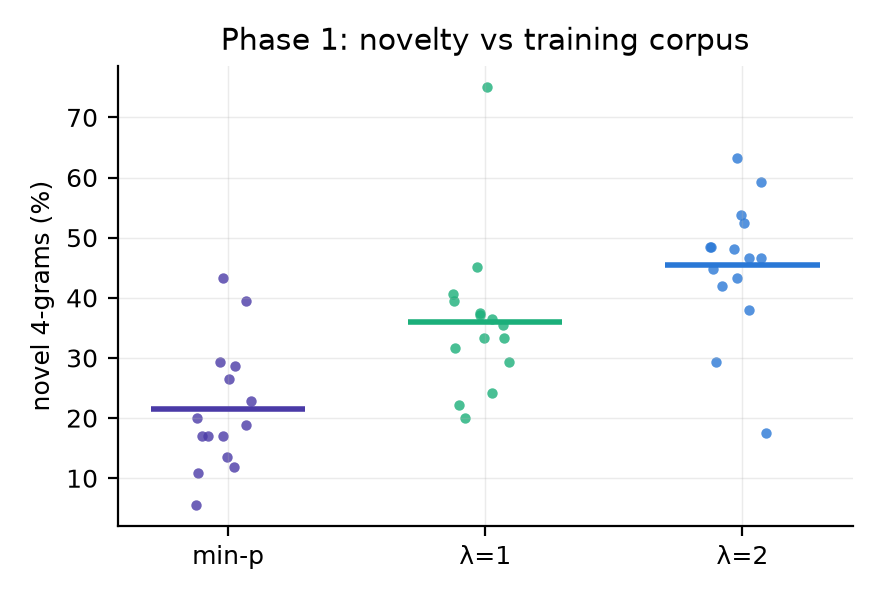}
\caption{Objective novelty against the OLMo-2 training corpus: novel 4-grams
per cell, 3 prompts $\times$ 5 seeds. The anti-probable push roughly doubles
surface novelty over min-$p$ sampling.}
\label{fig:phase1}
\end{figure}

\paragraph{The sampler doubles surface novelty.} On OLMo-2-13B, whose
training corpus is public, we ran a fully paired grid of 3 prompts $\times$ 5
seeds $\times$ \{min-$p$ baseline, $\lambda=1$, $\lambda=2$\} and counted,
with infini-gram, which $n$-grams of each generation occur anywhere in the
corpus (Figure~\ref{fig:phase1}). Novel 4-grams rise from 21.5\% (baseline)
to 36.0\% ($\lambda=1$) and 45.5\% ($\lambda=2$). The paired bootstrap
$\Delta$ is +14.6\,pp \ci{7.9}{21.9} and +24.0\,pp \ci{18.2}{29.5},
monotone in $\lambda$ and present in each prompt separately. The share of generations
containing a verbatim training block of eight or more words falls from 0.80
to 0.20 in both machine arms. The coherence cost is about +1.2 perplexity
under a cross-family model, uncorrelated with novelty within arms (Spearman
$+0.12$). Three escape modes were mapped: recitation, collage and factual paraphrase.
The entropy ceiling of the band, above which the model is at a genre fork
and pushing produces collapse, transferred across model families.

\paragraph{And has no detectable effect at the level of ideas.} Two tests.
The first is verified search: online bin packing, the FunSearch benchmark,
with a deterministic verifier in place of a judge, in evolutionary loops of
5 runs $\times$ 8 generations $\times$ 20--40 candidates on Qwen3-8B and
Qwen3-30B-A3B, 3,200 verified heuristics on the larger model. There the
anti-probable operator does not separate from plain sampling; both arms
rediscover best fit at will and neither beats it on held-out instances. The
second is the reverie loop of Section~\ref{sec:loop}, where the same push
against plain sampling in the same scaffold shows no detectable difference
in judged surprise, connection or novel delta across 1,335 window-level
judgments by two judges under three rubrics (on the surprise rubric, 2.97 vs
3.48, CI of the difference \ci{-1.50}{+0.49}). With ten cells per arm and a
judge whose per-window noise is about $\pm0.7$, an effect of half a point
would not have been detected. The conclusion is a null at the present
resolution, not an absence.

\section{The prompt}
\label{sec:prompt}

\begin{figure}[t]
\centering
\includegraphics[width=0.72\linewidth]{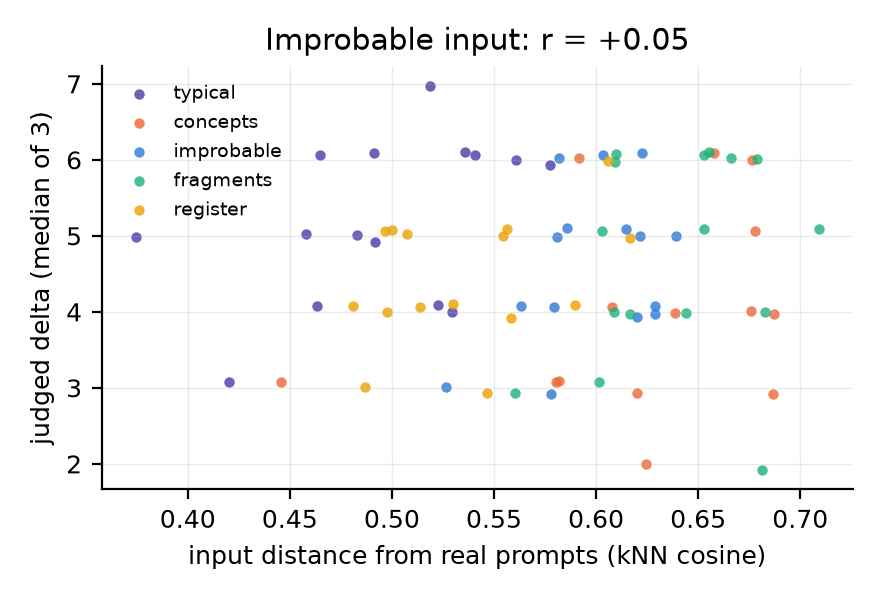}
\caption{Judged novel delta of the developed idea against the input's
semantic distance from 10,000 real prompts, per arm. No relationship.}
\label{fig:prompt}
\end{figure}

We composed three kinds of input: a typical request, a distant concept pair
drawn from a band of embedding distances, and a composed improbable
context, plus two ablations (fragments only, register only). Claude Opus~5
developed each into an idea and Claude Sonnet~5 judged the result against
its nearest known equivalent ($k=3$; $n\approx15$ per arm; developer and
judge from the same family, the one within-family judgment in this paper).
Input improbability was measured two ways: semantic $k$NN distance to 10,000
real instructions (Alpaca, embedded with all-mpnet-base-v2) and perplexity.
The typical prompt matched or beat every improbable arm, and two of the
improbable arms scored lower by a margin whose interval excluded zero.
Improbability did not correlate with judged novelty ($|r|<0.2$,
Figure~\ref{fig:prompt}), and a pilot effect ($n=8$) had not replicated. We
found no benefit from prompt improbability under the tested
operationalizations; whether other operationalizations, developers or judges
would find one is open.

\section{Reproducibility}
\label{app:repro}

\paragraph{Code and data.} The code is in the public repository at
\url{https://github.com/RobertoOno/interrupting-the-loop}: the sampler core,
the MLX adapter, the salience monitor, the reverie engine, the judges, the
novelty client, resumable battery runners with thermal logging, and the
analysis and figure scripts (117 tests). The human-rating packs (the windows
and the guide shown to raters) and the dated laboratory notebook are in the
same repository. The run data (every cell's token stream and the positions
of its events and injections) and the LLM judgments are in the evidence
deposit at \href{https://doi.org/\evidencedoi}{doi:\evidencedoi}. Individual
human ratings are not released, to protect rater anonymity; only the
aggregate statistics reported here are. The judge cost of the entire
program, including the re-judging under the
generated-only protocol, the document-level judgments and the replications,
was about US\$500.

\paragraph{Generators.} \texttt{Qwen3-30B-A3B-Base} (mixture of experts, 3B
active parameters), \texttt{Qwen3-8B-Base} and \texttt{OLMo-2-1124-13B},
each converted with \texttt{mlx\_lm} to 8-bit affine quantization with group
size 64 and run through MLX on an Apple M5 Pro (48\,GB unified memory) at
about 53, 31 and 17 tokens/s. Sampling is done in numpy on the full logits
returned by the model. All loop arms use temperature 1.0 and the min-$p$ floor
0.05; the anti-probable push is off ($\lambda=0$).

\paragraph{Loop hyperparameters.} Cell length 4,500 generated tokens; end-of-text
masked except in the EOS-allowed arm. Habituation: repetition window 512
tokens, penalty factor 1.15 per occurrence (1.3 in the strong-habituation
arm), applied to the log-probabilities before the floor. Interruption on a
clock: a stagnation check every \emph{period} tokens (75, 150, 300, 600 or
900) that always fires, injecting the next text of the rotation; the four
neutral subject changes are listed in Section~\ref{sec:loop-method}, the four
return-to-the-premise stitches are \emph{``And this, it turned out, was the
same thing as the beginning, because''}, \emph{``It came back to where it had
started, of course; it always did, but this time''}, \emph{``Which is exactly
what the first line had meant, seen from here:''}, \emph{``So the opening
sentence had been true after all, only not in the way it seemed:''}; the own-past
arm injects a 64-token window of the stream from at least 400 tokens back;
the sham arms inject a paragraph break or \emph{``And so, as before,''}. Reset
arms rebuild the cache from the premise and the injected text only. The full
scaffold: salience monitor with jump lag 32 and jump threshold 0.55, stagnation
window 600 and threshold 0.08, entropy-drop trigger 0.45 after a window at or
above 2.0 nats, recurrence threshold 0.15 with minimum age 96, refractory 24
steps; kick of 80 tokens ($\lambda=4$, band $[1.2, 5.0]$, half-life 12) after
a stagnation, subject change with forgetting after 3 consecutive stagnations
(keeping the premise, up to 3 judged-good windows and the last 200 tokens);
escalation of 120 tokens on a passing review ($\lambda=0.5$, band $[2.5,
4.5]$); anchors every 200 tokens, at most 12; genre check every 64 tokens over
160 tokens with threshold 0.5; drift regime $\lambda=2.5$, band $[1.8, 4.5]$,
half-life 48, bridge 1.5. In the arms of this paper $\lambda$ and the bridge
are set to 0 in every regime. The random-number seed of the sampler is 0 in
every cell.

\paragraph{Premises.} The ten premises, shared by all arms and generators:
(0) \emph{She kept a notebook of things that had almost happened.} (1)
\emph{The map was accurate in every detail except one, and nobody could say
which.} (2) \emph{Every morning the baker counted the loaves twice, and every
morning the count was different.} (3) \emph{He had inherited his grandfather's
watch and, with it, the habit of arriving early to places that no longer
existed.} (4) \emph{The river changed its name each time it crossed the
border, and the villagers kept all the names.} (5) \emph{The town had two
clocks, and nobody remembered which one had been right first.} (6) \emph{Her
brother collected sounds the way other children collected stones.} (7)
\emph{The letter arrived forty years late and was still, somehow, on time.}
(8) \emph{Every door in the house opened onto the same room, seen from a
different year.} (9) \emph{The translator kept a list of words she refused to
translate.} They were written by the author before any loop experiment and
not changed afterwards.

\paragraph{Judge.} Model \texttt{anthropic.claude-opus-5} on Amazon Bedrock
(effort \texttt{high}, adaptive thinking, \texttt{max\_tokens} 4000, default
temperature), called through the Anthropic SDK; the event-protocol judgments
were made on 14--17 August 2026 and the generated-only judgments on 18--19
August 2026. The second judge family was \texttt{moonshotai/kimi-k2.6}
through OpenRouter with reasoning disabled (17 August 2026). The in-loop judge
of the full scaffold was \texttt{anthropic.claude-sonnet-5} with the reverie
rubric (a nearest-equivalent and novel-delta rubric whose geometric-mean score
gated escalation at 5). The system prompt of the surprise judge, verbatim:

\begin{quote}\small\ttfamily
You judge a stretch of a language model's unsupervised reverie. You see the
EARLIER text and the RECENT window. Rate three INDEPENDENT things on 0-10 and
return ONLY a JSON object with keys: "surprise" (how unexpected is the recent
window given the earlier text, a reader could not have predicted where it
went; 0 = obvious continuation, 10 = genuinely startling yet not random),
"connection" (does the window bring together two distant regions of the
earlier text, or an old region with something new, in a way that makes sense;
0 if it merely continues one thread), "coherence" (does the window hold
together as text; 0 = word salad or document boilerplate), "note" (one blunt
sentence). Rate each dimension on its own merits; a window can be surprising
and incoherent, or coherent and dull.
\end{quote}

\noindent The user message is \texttt{EARLIER TEXT:\textbackslash n\{600 preceding
tokens\}\textbackslash n\textbackslash nRECENT WINDOW:\textbackslash n\{window\}}.
The judge is not told which condition, model or protocol produced the window.

\paragraph{The idea experiment.} Inputs were developed by
\texttt{anthropic.claude-opus-5} and judged by \texttt{anthropic.claude-sonnet-5}
($k=3$) with the nearest-equivalent rubric; input improbability was measured
as the mean cosine distance to the 10 nearest of 10,000 instructions from the
Alpaca dataset (\texttt{tatsu-lab/alpaca}) embedded with
\texttt{all-mpnet-base-v2}, and as perplexity under Qwen3-8B-Base.

\section{The human study}
\label{app:human}

\paragraph{Recruitment, task and compensation.} Raters were recruited on a
freelance marketplace (Workana) with a public posting for a paid text-rating
task in English; applicants were selected by the author among the platform's
respondents and were not told the hypotheses, the conditions or which model
produced the texts (the guide says only that the passages were written by a
language model left to write on its own). Each rater received a self-contained HTML
pack (rating guide embedded, English interface) with 63 windows drawn from
seven conditions of the main generator, presented in a fixed random order,
without any condition label, each with its preceding context; they rated
surprise and coherence on the same 0--10 scales given to the model judge, at
their own pace, and returned a JSON file. The guide asked them not to use AI
tools or web searches for any part of the task, and raters confirmed in
writing that they had not. Each rater was paid per pack at the platform's
market rate for an estimated 60--90 minutes of work. A second round (56
generated-only, fresh windows from eight conditions, seven per condition,
one per randomly chosen cell, not stratified by the judge; three dimensions
including connection; five raters) was prepared but not run.

\paragraph{Ethics.} The study involved adult professional raters reading
machine-generated fiction; there was no deception, no sensitive content, no
vulnerable population and no collection of personal data beyond the platform
account used for payment. Participation was voluntary and paid; raters were
informed in the posting that the texts were generated by language models and
that their ratings would be used, in aggregate, in a research publication.
The author is an independent researcher without access to an institutional
review board; the protocol was designed to be minimal-risk and the raters
remain anonymous in this paper. Their individual ratings are not released;
only aggregate statistics are reported.

\section{The event-window protocol of the first version}
\label{app:events}

The first version of this study cut 160-token windows at two kinds of review
point, right before each injection and on a 150-token grid, so that a window
after an interruption could contain the injected sentence, and treated
windows as the unit. The tables of that protocol are kept below for
comparison with the generated-only protocol of the main text. Across the 14
conditions of the first two batteries, the condition means under the two
protocols correlate at Spearman $\rho=+0.81$ on surprise, $+0.93$ on
connection and $+0.83$ on coherence (Table~\ref{tab:gen-protocol}); the
directions of the ladder, content and timing effects agree, the sizes differ
where an injected sentence sat inside the event window, and the period and
timing findings of the first version (a yield growing with the period, a
stream-level advantage of the clock over salience timing) do not hold on
generated text (Sections~\ref{sec:timing} and~\ref{sec:period}).

\section{Battery B: the problem with a verifier}
\label{app:problem}

Tables of Section~\ref{sec:problem} (generated by
\texttt{scripts/problem\_analysis.py} from \texttt{scripts/problem\_verify.py}):
the premise, the fifteen angle comments and the verifier are in
\texttt{src/creative\_machine/problem\_premises.py} and
\texttt{src/creative\_machine/domains/binpack.py}. Held-out instances: five
of 100 items per variant, seeds 200--209; training instances (reported in
the data) seeds 100--109.

\begin{table}[H]
\centering
\scriptsize
\begin{adjustbox}{max width=\textwidth}
\begin{tabular}{@{}lcccccc@{}}
\toprule
arm & n & candidates & valid & distinct valid & gain over best fit [CI] & cells beating best fit \\
\midrule
plain & 10 & 2.6 & 2.0 & 1.8 & -0.0092 [-0.0142, -0.0049] & 5\% \\
sham300 & 10 & 6.2 & 1.6 & 0.8 & -0.0253 [-0.0341, -0.0165] & 0\% \\
angle300 & 10 & 14.8 & 9.2 & 5.1 & -0.0155 [-0.0266, -0.0060] & 5\% \\
reset300 & 10 & 12.3 & 7.5 & 6.6 & -0.0022 [-0.0052, +0.0002] & 5\% \\
\bottomrule
\end{tabular}
\end{adjustbox}
\caption{Battery B: candidates, valid and distinct valid functions per stream, and gain of the best candidate over best fit on held-out instances.}
\label{tab:problem-1}
\end{table}

\begin{table}[H]
\centering
\scriptsize
\begin{adjustbox}{max width=\textwidth}
\begin{tabular}{@{}lccccc@{}}
\toprule
hypothesis & contrast & measure & $\Delta$ [CI] & p & n \\
\midrule
B1 (primary) & angle300 vs plain & gain & -0.0063 [-0.0159, +0.0009] & 0.9375 & 10 \\
B2 & angle300 vs plain & distinct & +3.3000 [+0.7500, +6.6500] & 0.0117 & 10 \\
B3 (two-sided) & angle300 vs reset300 & gain & -0.0133 [-0.0250, -0.0034] & 0.0410 & 10 \\
B4 (two-sided) & sham300 vs plain & gain & -0.0160 [-0.0265, -0.0065] & 0.0156 & 10 \\
exploratory & reset300 vs plain & gain & +0.0070 [+0.0020, +0.0121] & 0.0293 & 10 \\
exploratory & reset300 vs plain & distinct & +4.8000 [+2.7000, +6.8500] & 0.0039 & 10 \\
exploratory & angle300 vs plain & valid & +7.2000 [+2.6000, +12.8012] & 0.0078 & 10 \\
exploratory & angle300 vs sham300 & distinct & +4.3000 [+1.2000, +8.1500] & 0.0078 & 10 \\
\bottomrule
\end{tabular}
\end{adjustbox}
\caption{Battery B. Pre-registered contrasts (exact sign-flip permutation; one-sided where pre-specified).}
\label{tab:problem-2}
\end{table}

\begin{table}[H]
\centering
\scriptsize
\begin{adjustbox}{max width=\textwidth}
\begin{tabular}{@{}lcccc@{}}
\toprule
arm & mean best excess & mean best-fit excess & cells with best == best fit (within 1e-9) & cells with best < best fit \\
\midrule
plain & 0.1095 & 0.1002 & 7/20 & 1/20 \\
sham300 & 0.1255 & 0.1002 & 2/20 & 0/20 \\
angle300 & 0.1158 & 0.1002 & 7/20 & 1/20 \\
reset300 & 0.1024 & 0.1002 & 15/20 & 1/20 \\
\bottomrule
\end{tabular}
\end{adjustbox}
\caption{Battery B. Best candidate against the baselines (held-out).}
\label{tab:problem-3}
\end{table}

\section{Full tables}
\label{app:loop}

The tables below are generated from the run data by the analysis scripts
(\texttt{scripts/analysis\_gen.py} for the generated-only protocol,
\texttt{analysis.py} and \texttt{analysis\_b2.py} for the event protocol,
\texttt{hidden\_analysis.py} for the residual stream) and typeset by
\texttt{scripts/appendix\_tex.py}; the complete set, including every per-layer
table of the residual-stream analysis, ships with the code as
\texttt{docs/APPENDIX\_*.md}. In the event-protocol tables the unit is the
judged window (median of $k=5$ Opus~5 judgments), windows before 100
generated tokens are excluded, and intervals are 95\% bootstrap CIs over
windows; they are descriptive.

\subsection*{Generated-only windows, unit = cell (primary analysis)}
\begin{table}[H]
\centering
\scriptsize
\begin{adjustbox}{max width=\textwidth}
\begin{tabular}{@{}lcccc@{}}
\toprule
condition & n cells & surprise & connection & coherence \\
\midrule
bare & 10 & 0.45 [0.28, 0.68] & 0.30 [0.15, 0.45] & 2.52 [1.88, 3.43] \\
bare + habituation & 10 & 1.60 [1.18, 2.00] & 1.32 [0.87, 1.82] & 4.40 [3.45, 5.42] \\
reseed 150, no habituation & 10 & 2.48 [2.17, 2.78] & 3.03 [2.60, 3.45] & 5.35 [4.82, 5.77] \\
habituation + reseed 150 & 10 & 3.02 [2.53, 3.53] & 3.68 [2.92, 4.52] & 6.12 [5.60, 6.57] \\
DREAM scaffold & 10 & 2.70 [2.10, 3.40] & 1.85 [1.42, 2.33] & 6.02 [5.32, 6.73] \\
habituation 1.3 & 10 & 1.75 [1.35, 2.20] & 1.42 [1.07, 1.78] & 4.90 [4.47, 5.40] \\
habituation, EOS allowed & 10 & 1.52 [1.10, 1.92] & 1.27 [0.90, 1.67] & 5.65 [4.78, 6.38] \\
\bottomrule
\end{tabular}
\end{adjustbox}
\caption{Generated-only protocol. Q1: habituation $\times$ interruption (period 150) and baselines: cell means (mean over cells [95\% CI over cells]).}
\label{tab:gen-q1}
\end{table}

\begin{table}[H]
\centering
\scriptsize
\begin{adjustbox}{max width=\textwidth}
\begin{tabular}{@{}p{4.6cm}cccccc@{}}
\toprule
condition & dim & $\Delta$ [CI] & p (perm) & q (BH) & Cliff $\delta$ & n seeds \\
\midrule
bare & surprise & -1.15 [-1.48, -0.73] & 0.004 & 0.023 & -0.80 & 10 \\
bare & connection & -1.02 [-1.52, -0.50] & 0.008 & 0.035 & -0.81 & 10 \\
bare & coherence & -1.88 [-3.30, -0.52] & 0.035 & 0.070 & -0.64 & 10 \\
reseed 150, no habituation & surprise & +0.88 [+0.27, +1.48] & 0.031 & 0.070 & +0.72 & 10 \\
reseed 150, no habituation & connection & +1.72 [+0.88, +2.45] & 0.010 & 0.035 & +0.91 & 10 \\
reseed 150, no habituation & coherence & +0.95 [-0.02, +1.93] & 0.121 & 0.168 & +0.41 & 10 \\
habituation + reseed 150 & surprise & +1.42 [+0.92, +2.03] & 0.002 & 0.018 & +0.85 & 10 \\
habituation + reseed 150 & connection & +2.37 [+1.57, +3.28] & 0.002 & 0.018 & +0.93 & 10 \\
habituation + reseed 150 & coherence & +1.72 [+0.35, +2.90] & 0.043 & 0.070 & +0.61 & 10 \\
DREAM scaffold & surprise & +1.10 [+0.40, +1.88] & 0.023 & 0.070 & +0.62 & 10 \\
DREAM scaffold & connection & +0.53 [+0.15, +0.97] & 0.031 & 0.070 & +0.40 & 10 \\
DREAM scaffold & coherence & +1.62 [+0.20, +2.97] & 0.043 & 0.070 & +0.60 & 10 \\
habituation 1.3 & surprise & +0.15 [-0.25, +0.60] & 0.609 & 0.731 & +0.02 & 10 \\
habituation 1.3 & connection & +0.10 [-0.42, +0.55] & 0.754 & 0.817 & +0.10 & 10 \\
habituation 1.3 & coherence & +0.50 [-0.65, +1.52] & 0.430 & 0.552 & +0.23 & 10 \\
habituation, EOS allowed & surprise & -0.08 [-0.50, +0.35] & 0.771 & 0.817 & -0.02 & 10 \\
habituation, EOS allowed & connection & -0.05 [-0.55, +0.40] & 0.895 & 0.895 & -0.06 & 10 \\
habituation, EOS allowed & coherence & +1.25 [+0.12, +2.25] & 0.064 & 0.097 & +0.49 & 10 \\
\bottomrule
\end{tabular}
\end{adjustbox}
\caption{Generated-only protocol. vs bare + habituation: paired by seed.}
\label{tab:gen-q1-2}
\end{table}

\begin{table}[H]
\centering
\scriptsize
\begin{adjustbox}{max width=\textwidth}
\begin{tabular}{@{}lcccc@{}}
\toprule
condition & n cells & surprise & connection & coherence \\
\midrule
neutral subject change & 10 & 3.02 [2.53, 3.53] & 3.68 [2.92, 4.52] & 6.12 [5.60, 6.57] \\
re-encounter stitch & 10 & 2.83 [2.12, 3.58] & 3.62 [2.78, 4.50] & 5.53 [4.93, 6.12] \\
the premise itself & 10 & 1.21 [0.87, 1.51] & 1.03 [0.73, 1.32] & 3.84 [3.17, 4.59] \\
the stream's own past & 10 & 1.51 [1.12, 1.94] & 1.39 [0.92, 1.92] & 3.68 [3.07, 4.27] \\
\bottomrule
\end{tabular}
\end{adjustbox}
\caption{Generated-only protocol. Q2: what is injected (period 150): cell means (mean over cells [95\% CI over cells]).}
\label{tab:gen-q2}
\end{table}

\begin{table}[H]
\centering
\scriptsize
\begin{adjustbox}{max width=\textwidth}
\begin{tabular}{@{}p{4.6cm}cccccc@{}}
\toprule
condition & dim & $\Delta$ [CI] & p (perm) & q (BH) & Cliff $\delta$ & n seeds \\
\midrule
re-encounter stitch & surprise & -0.19 [-1.14, +0.68] & 0.707 & 0.795 & -0.13 & 10 \\
re-encounter stitch & connection & -0.07 [-1.42, +1.13] & 0.949 & 0.949 & -0.08 & 10 \\
re-encounter stitch & coherence & -0.59 [-1.45, +0.16] & 0.223 & 0.286 & -0.43 & 10 \\
the premise itself & surprise & -1.81 [-2.33, -1.28] & 0.002 & 0.004 & -1.00 & 10 \\
the premise itself & connection & -2.65 [-3.45, -1.90] & 0.002 & 0.004 & -1.00 & 10 \\
the premise itself & coherence & -2.27 [-3.07, -1.52] & 0.002 & 0.004 & -0.87 & 10 \\
the stream's own past & surprise & -1.51 [-2.10, -0.87] & 0.004 & 0.006 & -0.85 & 10 \\
the stream's own past & connection & -2.29 [-3.29, -1.42] & 0.002 & 0.004 & -0.90 & 10 \\
the stream's own past & coherence & -2.43 [-3.23, -1.67] & 0.002 & 0.004 & -0.94 & 10 \\
\bottomrule
\end{tabular}
\end{adjustbox}
\caption{Generated-only protocol. vs neutral subject change: paired by seed.}
\label{tab:gen-q2-2}
\end{table}

\begin{table}[H]
\centering
\scriptsize
\begin{adjustbox}{max width=\textwidth}
\begin{tabular}{@{}lcccc@{}}
\toprule
condition & n cells & surprise & connection & coherence \\
\midrule
no interruption & 10 & 1.60 [1.18, 2.00] & 1.32 [0.87, 1.82] & 4.40 [3.45, 5.42] \\
paragraph break (sham) & 10 & 1.47 [1.09, 1.84] & 0.97 [0.65, 1.33] & 4.50 [3.64, 5.34] \\
continuity connective (sham) & 10 & 1.08 [0.67, 1.55] & 0.89 [0.50, 1.34] & 3.45 [2.98, 3.97] \\
subject change, context preserved & 10 & 2.90 [2.45, 3.35] & 2.40 [2.03, 2.78] & 6.40 [6.08, 6.70] \\
subject change, no habituation & 10 & 2.50 [2.13, 2.90] & 2.02 [1.58, 2.57] & 5.53 [4.67, 6.35] \\
subject change, context reset & 10 & 3.75 [3.42, 4.12] & 3.30 [3.02, 3.58] & 6.87 [6.67, 7.07] \\
break, context reset & 10 & 2.07 [1.73, 2.48] & 1.78 [1.55, 2.03] & 5.40 [5.12, 5.70] \\
\bottomrule
\end{tabular}
\end{adjustbox}
\caption{Generated-only protocol. Q3: period 300: what carries the effect (context preserved vs reset; semantic change vs neutral boundary): cell means (mean over cells [95\% CI over cells]).}
\label{tab:gen-q3}
\end{table}

\begin{table}[H]
\centering
\scriptsize
\begin{adjustbox}{max width=\textwidth}
\begin{tabular}{@{}p{4.6cm}cccccc@{}}
\toprule
condition & dim & $\Delta$ [CI] & p (perm) & q (BH) & Cliff $\delta$ & n seeds \\
\midrule
paragraph break (sham) & surprise & -0.13 [-0.66, +0.39] & 0.664 & 0.703 & -0.15 & 10 \\
paragraph break (sham) & connection & -0.34 [-0.76, +0.07] & 0.178 & 0.246 & -0.24 & 10 \\
paragraph break (sham) & coherence & +0.10 [-0.98, +1.32] & 0.898 & 0.898 & +0.07 & 10 \\
continuity connective (sham) & surprise & -0.52 [-1.23, +0.25] & 0.244 & 0.293 & -0.37 & 10 \\
continuity connective (sham) & connection & -0.43 [-1.18, +0.38] & 0.340 & 0.382 & -0.35 & 10 \\
continuity connective (sham) & coherence & -0.95 [-2.27, +0.38] & 0.211 & 0.271 & -0.40 & 10 \\
subject change, context preserved & surprise & +1.30 [+0.68, +1.92] & 0.004 & 0.018 & +0.85 & 10 \\
subject change, context preserved & connection & +1.08 [+0.47, +1.63] & 0.014 & 0.040 & +0.74 & 10 \\
subject change, context preserved & coherence & +2.00 [+0.78, +3.22] & 0.016 & 0.040 & +0.68 & 10 \\
subject change, no habituation & surprise & +0.90 [+0.45, +1.37] & 0.016 & 0.040 & +0.72 & 10 \\
subject change, no habituation & connection & +0.70 [-0.07, +1.50] & 0.148 & 0.246 & +0.53 & 10 \\
subject change, no habituation & coherence & +1.13 [+0.00, +2.42] & 0.133 & 0.246 & +0.34 & 10 \\
subject change, context reset & surprise & +2.15 [+1.72, +2.63] & 0.002 & 0.018 & +1.00 & 10 \\
subject change, context reset & connection & +1.98 [+1.53, +2.38] & 0.002 & 0.018 & +0.93 & 10 \\
subject change, context reset & coherence & +2.47 [+1.55, +3.35] & 0.004 & 0.018 & +0.81 & 10 \\
break, context reset & surprise & +0.47 [-0.08, +1.03] & 0.174 & 0.246 & +0.41 & 10 \\
break, context reset & connection & +0.47 [-0.10, +1.00] & 0.156 & 0.246 & +0.47 & 10 \\
break, context reset & coherence & +1.00 [+0.10, +1.87] & 0.074 & 0.167 & +0.48 & 10 \\
\bottomrule
\end{tabular}
\end{adjustbox}
\caption{Generated-only protocol. vs no interruption: paired by seed.}
\label{tab:gen-q3-2}
\end{table}

\begin{table}[H]
\centering
\scriptsize
\begin{adjustbox}{max width=\textwidth}
\begin{tabular}{@{}p{4.6cm}ccccc@{}}
\toprule
contrast & dim & $\Delta$ [CI] & p (perm) & Cliff $\delta$ & n \\
\midrule
reset vs preserved (300) & surprise & +0.85 [+0.42, +1.28] & 0.014 & +0.66 & 10 \\
reset vs preserved (300) & connection & +0.90 [+0.53, +1.25] & 0.004 & +0.79 & 10 \\
reset vs preserved (300) & coherence & +0.47 [+0.03, +0.90] & 0.090 & +0.54 & 10 \\
with vs without habituation (300) & surprise & +0.40 [-0.17, +1.05] & 0.289 & +0.34 & 10 \\
with vs without habituation (300) & connection & +0.38 [-0.30, +1.02] & 0.342 & +0.43 & 10 \\
with vs without habituation (300) & coherence & +0.87 [+0.05, +1.77] & 0.125 & +0.33 & 10 \\
with vs without habituation (150) & surprise & +0.53 [-0.13, +1.28] & 0.219 & +0.30 & 10 \\
with vs without habituation (150) & connection & +0.65 [-0.30, +1.65] & 0.256 & +0.21 & 10 \\
with vs without habituation (150) & coherence & +0.77 [+0.07, +1.45] & 0.074 & +0.63 & 10 \\
subject change vs break, both reset (300) & surprise & +1.68 [+1.23, +2.13] & 0.002 & +0.90 & 10 \\
subject change vs break, both reset (300) & connection & +1.52 [+1.17, +1.90] & 0.002 & +1.00 & 10 \\
subject change vs break, both reset (300) & coherence & +1.47 [+1.18, +1.78] & 0.002 & +0.99 & 10 \\
subject change vs break, both preserved (300) & surprise & +1.43 [+0.78, +2.07] & 0.006 & +0.85 & 10 \\
subject change vs break, both preserved (300) & connection & +1.42 [+0.95, +1.96] & 0.002 & +0.93 & 10 \\
subject change vs break, both preserved (300) & coherence & +1.90 [+0.98, +2.84] & 0.004 & +0.76 & 10 \\
\bottomrule
\end{tabular}
\end{adjustbox}
\caption{Generated-only protocol. Q3b: direct paired contrasts among the interruption arms.}
\label{tab:gen-q3-3}
\end{table}

\begin{table}[H]
\centering
\scriptsize
\begin{adjustbox}{max width=\textwidth}
\begin{tabular}{@{}lcccc@{}}
\toprule
condition & n cells & surprise & connection & coherence \\
\midrule
stitch on the clock 150 & 10 & 2.83 [2.12, 3.58] & 3.62 [2.78, 4.50] & 5.53 [4.93, 6.12] \\
stitch on the clock 900 & 10 & 3.08 [2.72, 3.50] & 2.76 [2.42, 3.12] & 5.84 [5.36, 6.30] \\
neutral change 900 & 10 & 3.18 [2.74, 3.76] & 2.26 [1.76, 2.84] & 6.78 [6.60, 6.94] \\
stitch on salience events & 10 & 3.09 [2.50, 3.69] & 2.82 [2.23, 3.51] & 5.68 [4.97, 6.47] \\
salience only & 10 & 1.65 [1.23, 2.02] & 1.30 [0.85, 1.77] & 4.63 [3.67, 5.65] \\
\bottomrule
\end{tabular}
\end{adjustbox}
\caption{Generated-only protocol. Q4: timing: salience vs clock (same stitch): cell means (mean over cells [95\% CI over cells]).}
\label{tab:gen-q4}
\end{table}

\begin{table}[H]
\centering
\scriptsize
\begin{adjustbox}{max width=\textwidth}
\begin{tabular}{@{}p{4.6cm}cccccc@{}}
\toprule
condition & dim & $\Delta$ [CI] & p (perm) & q (BH) & Cliff $\delta$ & n seeds \\
\midrule
stitch on the clock 150 & surprise & -0.25 [-0.90, +0.39] & 0.482 & 0.724 & -0.24 & 10 \\
stitch on the clock 150 & connection & +0.86 [+0.17, +1.58] & 0.057 & 0.170 & +0.30 & 10 \\
stitch on the clock 150 & coherence & -0.32 [-0.78, +0.13] & 0.230 & 0.395 & -0.22 & 10 \\
neutral change 900 & surprise & +0.10 [-0.52, +0.80] & 0.832 & 0.998 & +0.08 & 10 \\
neutral change 900 & connection & -0.50 [-1.16, +0.26] & 0.227 & 0.395 & -0.47 & 10 \\
neutral change 900 & coherence & +0.94 [+0.48, +1.38] & 0.008 & 0.031 & +0.77 & 10 \\
stitch on salience events & surprise & +0.01 [-0.77, +0.77] & 1.000 & 1.000 & -0.04 & 10 \\
stitch on salience events & connection & +0.06 [-0.69, +0.77] & 0.918 & 1.000 & -0.01 & 10 \\
stitch on salience events & coherence & -0.16 [-0.88, +0.59] & 0.695 & 0.927 & -0.15 & 10 \\
salience only & surprise & -1.43 [-2.13, -0.83] & 0.002 & 0.023 & -0.92 & 10 \\
salience only & connection & -1.46 [-2.16, -0.76] & 0.006 & 0.031 & -0.86 & 10 \\
salience only & coherence & -1.21 [-2.17, -0.01] & 0.076 & 0.183 & -0.46 & 10 \\
\bottomrule
\end{tabular}
\end{adjustbox}
\caption{Generated-only protocol. vs stitch on the clock 900: paired by seed.}
\label{tab:gen-q4-2}
\end{table}

\begin{table}[H]
\centering
\scriptsize
\begin{adjustbox}{max width=\textwidth}
\begin{tabular}{@{}lcccc@{}}
\toprule
condition & n cells & surprise & connection & coherence \\
\midrule
stitch on the clock 900 & 10 & 2.10 [1.62, 2.52] & 1.73 [1.40, 2.01] & 4.78 [4.10, 5.43] \\
neutral change 900 & 10 & 2.53 [2.16, 2.87] & 2.03 [1.70, 2.37] & 5.33 [4.74, 5.88] \\
neutral change 600 & 10 & 2.87 [2.38, 3.30] & 2.27 [1.88, 2.67] & 5.48 [4.95, 5.98] \\
stitch on salience events & 10 & 1.85 [1.36, 2.30] & 1.80 [1.17, 2.43] & 4.18 [3.73, 4.60] \\
DREAM scaffold & 7 & 1.64 [1.17, 2.14] & 1.46 [1.05, 1.79] & 5.61 [4.65, 6.44] \\
\bottomrule
\end{tabular}
\end{adjustbox}
\caption{Generated-only protocol. Q4b: deep windows (at least 300 tokens after the last injection): the stream when it is left alone: cell means (mean over cells [95\% CI over cells]).}
\label{tab:gen-q4b}
\end{table}

\begin{table}[H]
\centering
\scriptsize
\begin{adjustbox}{max width=\textwidth}
\begin{tabular}{@{}p{4.6cm}cccccc@{}}
\toprule
condition & dim & $\Delta$ [CI] & p (perm) & q (BH) & Cliff $\delta$ & n seeds \\
\midrule
neutral change 900 & surprise & +0.43 [-0.12, +0.98] & 0.195 & 0.299 & +0.35 & 10 \\
neutral change 900 & connection & +0.31 [-0.13, +0.72] & 0.225 & 0.299 & +0.30 & 10 \\
neutral change 900 & coherence & +0.55 [-0.13, +1.31] & 0.205 & 0.299 & +0.24 & 10 \\
neutral change 600 & surprise & +0.77 [+0.03, +1.46] & 0.078 & 0.299 & +0.60 & 10 \\
neutral change 600 & connection & +0.54 [+0.02, +1.09] & 0.125 & 0.299 & +0.43 & 10 \\
neutral change 600 & coherence & +0.70 [-0.15, +1.58] & 0.182 & 0.299 & +0.42 & 10 \\
stitch on salience events & surprise & -0.25 [-0.80, +0.25] & 0.439 & 0.479 & -0.19 & 10 \\
stitch on salience events & connection & +0.07 [-0.52, +0.67] & 0.820 & 0.820 & -0.08 & 10 \\
stitch on salience events & coherence & -0.60 [-1.41, +0.18] & 0.186 & 0.299 & -0.35 & 10 \\
DREAM scaffold & surprise & -0.42 [-1.10, +0.22] & 0.344 & 0.412 & -0.35 & 7 \\
DREAM scaffold & connection & -0.26 [-0.55, +0.06] & 0.188 & 0.299 & -0.35 & 7 \\
DREAM scaffold & coherence & +0.93 [+0.27, +1.62] & 0.062 & 0.299 & +0.43 & 7 \\
\bottomrule
\end{tabular}
\end{adjustbox}
\caption{Generated-only protocol. vs stitch on the clock 900: paired by seed.}
\label{tab:gen-q4b-2}
\end{table}

\begin{table}[H]
\centering
\scriptsize
\begin{adjustbox}{max width=\textwidth}
\begin{tabular}{@{}lcccc@{}}
\toprule
condition & n cells & surprise & connection & coherence \\
\midrule
150 & 10 & 3.02 [2.53, 3.53] & 3.68 [2.92, 4.52] & 6.12 [5.60, 6.57] \\
300 & 10 & 2.90 [2.45, 3.35] & 2.40 [2.03, 2.78] & 6.40 [6.08, 6.70] \\
600 & 10 & 2.88 [2.60, 3.17] & 2.00 [1.72, 2.30] & 6.27 [5.70, 6.72] \\
900 & 10 & 3.18 [2.74, 3.76] & 2.26 [1.76, 2.84] & 6.78 [6.60, 6.94] \\
\bottomrule
\end{tabular}
\end{adjustbox}
\caption{Generated-only protocol. Q5: period of the neutral change: the window 32--128 tokens after the injection: cell means (mean over cells [95\% CI over cells]).}
\label{tab:gen-q5}
\end{table}

\begin{table}[H]
\centering
\scriptsize
\begin{adjustbox}{max width=\textwidth}
\begin{tabular}{@{}p{4.6cm}cccccc@{}}
\toprule
condition & dim & $\Delta$ [CI] & p (perm) & q (BH) & Cliff $\delta$ & n seeds \\
\midrule
300 & surprise & -0.12 [-0.75, +0.47] & 0.781 & 0.781 & -0.02 & 10 \\
300 & connection & -1.28 [-2.10, -0.53] & 0.008 & 0.023 & -0.64 & 10 \\
300 & coherence & +0.28 [-0.28, +0.82] & 0.389 & 0.700 & +0.22 & 10 \\
600 & surprise & -0.13 [-0.70, +0.37] & 0.713 & 0.781 & -0.04 & 10 \\
600 & connection & -1.68 [-2.43, -1.02] & 0.002 & 0.009 & -0.88 & 10 \\
600 & coherence & +0.15 [-0.60, +0.82] & 0.723 & 0.781 & +0.17 & 10 \\
900 & surprise & +0.16 [-0.24, +0.58] & 0.490 & 0.735 & +0.14 & 10 \\
900 & connection & -1.42 [-1.87, -0.99] & 0.002 & 0.009 & -0.64 & 10 \\
900 & coherence & +0.66 [+0.22, +1.10] & 0.023 & 0.053 & +0.62 & 10 \\
\bottomrule
\end{tabular}
\end{adjustbox}
\caption{Generated-only protocol. vs 150: paired by seed.}
\label{tab:gen-q5-2}
\end{table}

\begin{table}[H]
\centering
\scriptsize
\begin{adjustbox}{max width=\textwidth}
\begin{tabular}{@{}p{4.6cm}ccccccc@{}}
\toprule
period & offset 32 & offset 160 & offset 300 & offset 450 & offset 600 & offset 750 & stream estimate (S / C / H) \\
\midrule
150 & 3.02 (n=10) & -- & -- & -- & -- & -- & 3.02 / 3.68 / 6.12 \\
300 & 2.90 (n=10) & 2.63 (n=10) & -- & -- & -- & -- & 2.78 / 2.38 / 6.24 \\
600 & 2.88 (n=10) & 2.75 (n=10) & 3.07 (n=10) & 2.67 (n=10) & -- & -- & 2.84 / 2.12 / 5.80 \\
900 & 3.18 (n=10) & 2.63 (n=10) & 3.13 (n=10) & 2.73 (n=10) & 2.27 (n=10) & 2.00 (n=10) & 2.66 / 2.09 / 5.81 \\
\bottomrule
\end{tabular}
\end{adjustbox}
\caption{Generated-only protocol. Q5b: decay: cell mean surprise by tokens since the injection (window start), generated text only; stream estimate = offset means weighted by the stretch of the segment each window represents.}
\label{tab:gen-q5b}
\end{table}

\begin{table}[H]
\centering
\scriptsize
\begin{adjustbox}{max width=\textwidth}
\begin{tabular}{@{}lcccc@{}}
\toprule
condition & n cells & surprise & connection & coherence \\
\midrule
bare & 10 & 0.43 [0.22, 0.67] & 0.40 [0.22, 0.60] & 3.88 [2.72, 5.25] \\
bare + habituation & 10 & 1.37 [1.03, 1.72] & 1.00 [0.65, 1.37] & 4.35 [3.35, 5.35] \\
habituation + reseed 150 & 10 & 2.76 [2.23, 3.22] & 3.17 [2.57, 3.65] & 5.15 [4.42, 5.77] \\
DREAM scaffold & 10 & 2.70 [2.00, 3.43] & 2.35 [1.65, 3.12] & 5.38 [4.45, 6.10] \\
\bottomrule
\end{tabular}
\end{adjustbox}
\caption{Generated-only protocol. Q6: the ladder on Qwen3-8B-Base: cell means (mean over cells [95\% CI over cells]).}
\label{tab:gen-q6}
\end{table}

\begin{table}[H]
\centering
\scriptsize
\begin{adjustbox}{max width=\textwidth}
\begin{tabular}{@{}p{4.6cm}cccccc@{}}
\toprule
condition & dim & $\Delta$ [CI] & p (perm) & q (BH) & Cliff $\delta$ & n seeds \\
\midrule
bare & surprise & -0.93 [-1.35, -0.53] & 0.004 & 0.018 & -0.84 & 10 \\
bare & connection & -0.60 [-1.08, -0.13] & 0.062 & 0.094 & -0.61 & 10 \\
bare & coherence & -0.47 [-2.13, +1.35] & 0.641 & 0.641 & -0.21 & 10 \\
habituation + reseed 150 & surprise & +1.39 [+0.72, +2.07] & 0.008 & 0.018 & +0.79 & 10 \\
habituation + reseed 150 & connection & +2.17 [+1.53, +2.83] & 0.002 & 0.018 & +0.90 & 10 \\
habituation + reseed 150 & coherence & +0.80 [-0.15, +1.73] & 0.172 & 0.193 & +0.28 & 10 \\
DREAM scaffold & surprise & +1.33 [+0.53, +2.20] & 0.016 & 0.028 & +0.69 & 10 \\
DREAM scaffold & connection & +1.35 [+0.62, +2.15] & 0.008 & 0.018 & +0.67 & 10 \\
DREAM scaffold & coherence & +1.03 [-0.12, +2.17] & 0.143 & 0.183 & +0.36 & 10 \\
\bottomrule
\end{tabular}
\end{adjustbox}
\caption{Generated-only protocol. vs bare + habituation: paired by seed.}
\label{tab:gen-q6-2}
\end{table}

\begin{table}[H]
\centering
\scriptsize
\begin{adjustbox}{max width=\textwidth}
\begin{tabular}{@{}lcccc@{}}
\toprule
condition & n cells & surprise & connection & coherence \\
\midrule
bare & 10 & 1.28 [0.78, 1.83] & 0.93 [0.47, 1.52] & 3.03 [2.18, 3.83] \\
bare + habituation & 10 & 1.96 [1.30, 2.61] & 1.26 [0.77, 1.76] & 3.99 [2.89, 5.07] \\
habituation + reseed 150 & 10 & 2.77 [2.19, 3.39] & 2.87 [2.30, 3.45] & 4.05 [3.57, 4.52] \\
DREAM scaffold & 10 & 3.60 [2.73, 4.47] & 2.23 [1.60, 2.94] & 6.23 [5.37, 7.05] \\
\bottomrule
\end{tabular}
\end{adjustbox}
\caption{Generated-only protocol. Q6: the ladder on OLMo-2-13B: cell means (mean over cells [95\% CI over cells]).}
\label{tab:gen-q6-3}
\end{table}

\begin{table}[H]
\centering
\scriptsize
\begin{adjustbox}{max width=\textwidth}
\begin{tabular}{@{}p{4.6cm}cccccc@{}}
\toprule
condition & dim & $\Delta$ [CI] & p (perm) & q (BH) & Cliff $\delta$ & n seeds \\
\midrule
bare & surprise & -0.68 [-1.63, +0.26] & 0.219 & 0.281 & -0.40 & 10 \\
bare & connection & -0.33 [-1.11, +0.45] & 0.484 & 0.545 & -0.25 & 10 \\
bare & coherence & -0.96 [-2.33, +0.12] & 0.184 & 0.275 & -0.29 & 10 \\
habituation + reseed 150 & surprise & +0.82 [-0.21, +1.81] & 0.166 & 0.275 & +0.36 & 10 \\
habituation + reseed 150 & connection & +1.61 [+0.83, +2.35] & 0.004 & 0.035 & +0.82 & 10 \\
habituation + reseed 150 & coherence & +0.06 [-0.99, +1.20] & 0.938 & 0.938 & +0.01 & 10 \\
DREAM scaffold & surprise & +1.64 [+0.42, +2.97] & 0.043 & 0.097 & +0.53 & 10 \\
DREAM scaffold & connection & +0.97 [+0.26, +1.84] & 0.023 & 0.070 & +0.48 & 10 \\
DREAM scaffold & coherence & +2.24 [+0.87, +3.63] & 0.020 & 0.070 & +0.68 & 10 \\
\bottomrule
\end{tabular}
\end{adjustbox}
\caption{Generated-only protocol. vs bare + habituation: paired by seed.}
\label{tab:gen-q6-4}
\end{table}

\begin{table}[H]
\centering
\scriptsize
\begin{adjustbox}{max width=\textwidth}
\begin{tabular}{@{}lcccc@{}}
\toprule
condition & surprise event / gen & connection event / gen & coherence event / gen & n cells (event / gen) \\
\midrule
abl\_forget & 3.44 / 2.72 & 1.73 / 1.90 & 3.88 / 6.13 & 10 / 10 \\
salience only & 2.86 / 1.65 & 1.53 / 1.30 & 3.85 / 4.63 & 10 / 10 \\
bare & 0.28 / 0.45 & 0.20 / 0.30 & 2.06 / 2.52 & 10 / 10 \\
bare + habituation & 1.70 / 1.60 & 1.08 / 1.32 & 4.08 / 4.40 & 10 / 10 \\
habituation + reseed 150 & 3.08 / 3.02 & 3.15 / 3.68 & 4.78 / 6.12 & 10 / 10 \\
reseed 300 & 4.01 / 2.90 & 3.12 / 2.40 & 5.98 / 6.40 & 10 / 10 \\
reseed 600 & 3.64 / 2.88 & 2.64 / 2.00 & 5.50 / 6.27 & 10 / 10 \\
reseed 900 & 4.28 / 3.18 & 2.53 / 2.26 & 5.55 / 6.78 & 10 / 10 \\
stitch 900 & 3.91 / 3.08 & 3.01 / 2.76 & 5.17 / 5.84 & 10 / 10 \\
premise 150 & 1.02 / 1.21 & 0.98 / 1.03 & 3.03 / 3.84 & 10 / 10 \\
re-encounter stitch 150 & 2.55 / 2.83 & 3.53 / 3.62 & 4.67 / 5.53 & 10 / 10 \\
own past 150 & 1.37 / 1.51 & 1.50 / 1.39 & 3.03 / 3.68 & 10 / 10 \\
salience-timed stitch & 3.00 / 3.09 & 2.26 / 2.82 & 3.85 / 5.68 & 10 / 10 \\
DREAM scaffold & 3.50 / 2.70 & 1.70 / 1.85 & 3.82 / 6.02 & 10 / 10 \\
\bottomrule
\end{tabular}
\end{adjustbox}
\caption{Generated-only protocol. Protocol agreement: condition means, event windows (v1) vs generated-only windows (unit = cell).}
\label{tab:gen-protocol}
\end{table}

\begin{table}[H]
\centering
\scriptsize
\begin{adjustbox}{max width=\textwidth}
\begin{tabular}{@{}p{4.6cm}ccccc@{}}
\toprule
condition & n windows & copied & copied, source outside the judge's 600 tokens & fresh windows: S / C / H (cell means, n cells) & copied windows: S / C / H \\
\midrule
bare & 60 & 67\% & 0\% & 1.77 / 1.38 / 5.55 (n=10) & 0.03 / 0.00 / 1.45 \\
bare + habituation & 60 & 27\% & 8\% & 1.76 / 1.32 / 5.01 (n=10) & 0.94 / 1.00 / 2.69 \\
habituation 1.3 & 60 & 18\% & 3\% & 1.96 / 1.54 / 5.39 (n=10) & 0.64 / 0.73 / 2.91 \\
habituation, EOS allowed & 60 & 15\% & 3\% & 1.53 / 1.30 / 5.83 (n=10) & 1.22 / 1.11 / 3.67 \\
reseed 150, no habituation & 60 & 80\% & 7\% & 3.03 / 1.83 / 6.17 (n=10) & 2.38 / 3.31 / 5.25 \\
habituation + reseed 150 & 60 & 72\% & 12\% & 2.95 / 2.03 / 6.78 (n=10) & 3.09 / 4.28 / 6.19 \\
reseed 300, no habituation & 60 & 68\% & 63\% & 2.45 / 1.65 / 5.65 (n=10) & 2.54 / 2.20 / 5.54 \\
reseed 300 & 60 & 65\% & 62\% & 3.02 / 2.03 / 6.52 (n=10) & 2.82 / 2.59 / 6.36 \\
reseed 600 & 60 & 38\% & 35\% & 2.84 / 2.04 / 6.33 (n=10) & 2.91 / 2.04 / 6.30 \\
reseed 900 & 50 & 12\% & 12\% & 3.12 / 2.25 / 6.75 (n=10) & 4.00 / 2.67 / 7.00 \\
re-encounter stitch 150 & 60 & 68\% & 45\% & 3.33 / 2.83 / 6.31 (n=10) & 2.71 / 3.85 / 5.29 \\
stitch 900 & 50 & 10\% & 6\% & 3.22 / 2.83 / 5.97 (n=10) & 2.40 / 2.60 / 4.60 \\
premise 150 & 60 & 60\% & 7\% & 2.23 / 1.61 / 5.00 (n=10) & 0.67 / 0.78 / 3.04 \\
own past 150 & 60 & 72\% & 27\% & 2.50 / 1.42 / 4.88 (n=10) & 1.26 / 1.43 / 3.26 \\
salience-timed stitch & 30 & 30\% & 27\% & 3.10 / 3.12 / 5.66 (n=9) & 3.89 / 3.00 / 5.33 \\
paragraph break 300 (sham) & 60 & 32\% & 5\% & 2.16 / 1.41 / 5.03 (n=10) & 0.26 / 0.53 / 3.05 \\
continuity connective 300 (sham) & 60 & 73\% & 3\% & 3.32 / 1.99 / 5.92 (n=10) & 0.36 / 0.45 / 2.66 \\
reset + new subject 300 & 60 & 0\% & 0\% & 3.75 / 3.30 / 6.87 (n=10) & -- \\
reset + break 300 & 60 & 0\% & 0\% & 2.07 / 1.78 / 5.40 (n=10) & -- \\
DREAM scaffold & 31 & 6\% & 0\% & 2.69 / 1.92 / 6.24 (n=10) & 2.50 / 1.50 / 2.00 \\
\bottomrule
\end{tabular}
\end{adjustbox}
\caption{Generated-only protocol. Self-copy: verbatim reproduction of the stream's own earlier text inside the judged windows.}
\label{tab:gen-self}
\end{table}

\begin{table}[H]
\centering
\scriptsize
\begin{adjustbox}{max width=\textwidth}
\begin{tabular}{@{}lcccc@{}}
\toprule
generator & condition & n windows & copied & fresh windows: S / C / H (n cells) \\
\midrule
Qwen3-8B-Base & bare & 60 & 63\% & 1.15 / 1.11 / 6.70 (n=9) \\
Qwen3-8B-Base & bare + habituation & 60 & 45\% & 2.53 / 1.90 / 6.03 (n=10) \\
Qwen3-8B-Base & habituation + reseed 150 & 60 & 78\% & 3.48 / 2.07 / 5.77 (n=10) \\
Qwen3-8B-Base & DREAM scaffold & 26 & 8\% & 2.78 / 2.41 / 5.47 (n=10) \\
OLMo-2-13B & bare & 60 & 45\% & 2.11 / 1.51 / 4.89 (n=10) \\
OLMo-2-13B & bare + habituation & 60 & 20\% & 2.20 / 1.33 / 4.31 (n=10) \\
OLMo-2-13B & habituation + reseed 150 & 60 & 45\% & 3.19 / 2.57 / 3.89 (n=10) \\
OLMo-2-13B & DREAM scaffold & 28 & 4\% & 3.60 / 2.23 / 6.48 (n=10) \\
\bottomrule
\end{tabular}
\end{adjustbox}
\caption{Generated-only protocol. Self-copy on the other generators (ladder arms; copied = >= 50\% shingles seen earlier in the stream).}
\label{tab:gen-self-2}
\end{table}

\begin{table}[H]
\centering
\scriptsize
\begin{adjustbox}{max width=\textwidth}
\begin{tabular}{@{}p{4.6cm}ccccc@{}}
\toprule
contrast & dim & $\Delta$ [CI] & p (perm) & Cliff $\delta$ & n seeds \\
\midrule
interruption 150 vs habituation & surprise & +1.19 [+0.42, +2.09] & 0.010 & +0.74 & 10 \\
interruption 150 vs habituation & connection & +0.71 [+0.36, +1.09] & 0.008 & +0.49 & 10 \\
interruption 150 vs habituation & coherence & +1.78 [+0.41, +2.90] & 0.035 & +0.64 & 10 \\
interruption 300 vs habituation & surprise & +1.25 [+0.71, +1.78] & 0.004 & +0.87 & 10 \\
interruption 300 vs habituation & connection & +0.71 [+0.33, +1.20] & 0.008 & +0.63 & 10 \\
interruption 300 vs habituation & coherence & +1.51 [+0.49, +2.47] & 0.023 & +0.66 & 10 \\
reset + subject change 300 vs habituation & surprise & +1.99 [+1.63, +2.34] & 0.002 & +1.00 & 10 \\
reset + subject change 300 vs habituation & connection & +1.98 [+1.68, +2.30] & 0.002 & +1.00 & 10 \\
reset + subject change 300 vs habituation & coherence & +1.86 [+1.13, +2.62] & 0.004 & +0.81 & 10 \\
reset vs preserved (300) & surprise & +0.73 [+0.28, +1.20] & 0.023 & +0.58 & 10 \\
reset vs preserved (300) & connection & +1.27 [+0.80, +1.72] & 0.004 & +0.93 & 10 \\
reset vs preserved (300) & coherence & +0.35 [-0.10, +0.78] & 0.199 & +0.29 & 10 \\
scaffold vs habituation & surprise & +0.93 [+0.21, +1.71] & 0.062 & +0.53 & 10 \\
scaffold vs habituation & connection & +0.59 [+0.15, +1.08] & 0.055 & +0.35 & 10 \\
scaffold vs habituation & coherence & +1.23 [+0.03, +2.43] & 0.078 & +0.52 & 10 \\
interruption 900 vs habituation & surprise & +1.35 [+0.84, +1.79] & 0.004 & +0.95 & 10 \\
interruption 900 vs habituation & connection & +0.92 [+0.30, +1.54] & 0.023 & +0.66 & 10 \\
interruption 900 vs habituation & coherence & +1.75 [+0.82, +2.65] & 0.008 & +0.78 & 10 \\
\bottomrule
\end{tabular}
\end{adjustbox}
\caption{Generated-only protocol. Fresh windows only: paired contrasts (cells with at least one fresh window).}
\label{tab:gen-self-3}
\end{table}

\begin{table}[H]
\centering
\scriptsize
\begin{adjustbox}{max width=\textwidth}
\begin{tabular}{@{}p{4.6cm}ccccc@{}}
\toprule
condition & n & integration & development & coherence & surprise \\
\midrule
bare & 10 & 0.90 [0.50, 1.30] & 0.50 [0.20, 0.80] & 1.10 [0.60, 1.60] & 0.60 [0.20, 1.00] \\
bare + habituation & 10 & 1.40 [0.90, 1.90] & 1.10 [0.60, 1.70] & 1.50 [1.20, 1.80] & 1.40 [1.10, 1.70] \\
habituation + reseed 150 & 10 & 0.70 [0.30, 1.10] & 0.30 [0.00, 0.60] & 1.20 [1.00, 1.50] & 0.10 [0.00, 0.30] \\
reseed 300 & 10 & 0.80 [0.50, 1.00] & 0.40 [0.10, 0.70] & 1.50 [1.20, 1.80] & 0.50 [0.20, 0.80] \\
reseed 300, no habituation & 10 & 0.80 [0.40, 1.20] & 0.20 [0.00, 0.50] & 1.40 [1.10, 1.70] & 0.40 [0.10, 0.70] \\
paragraph break 300 (sham) & 10 & 1.40 [0.80, 2.00] & 1.00 [0.60, 1.40] & 1.50 [0.90, 2.10] & 1.00 [0.60, 1.40] \\
reset + new subject 300 & 10 & 2.00 [1.60, 2.40] & 1.60 [1.30, 1.90] & 2.20 [1.80, 2.60] & 2.50 [2.20, 2.80] \\
\bottomrule
\end{tabular}
\end{adjustbox}
\caption{Generated-only protocol. Document level: the whole 4,500-token stream, injected sentences removed, Opus k=3 (230 documents).}
\label{tab:gen-document}
\end{table}

\begin{table}[H]
\centering
\scriptsize
\begin{adjustbox}{max width=\textwidth}
\begin{tabular}{@{}p{4.6cm}ccccc@{}}
\toprule
contrast & dim & $\Delta$ [CI] & p (perm) & Cliff $\delta$ & n \\
\midrule
interruption 150 vs habituation & integration & -0.70 [-1.50, +0.10] & 0.203 & -0.46 & 10 \\
interruption 150 vs habituation & development & -0.80 [-1.50, -0.30] & 0.062 & -0.56 & 10 \\
interruption 150 vs habituation & coherence & -0.30 [-0.60, +0.00] & 0.250 & -0.30 & 10 \\
interruption 150 vs habituation & surprise & -1.30 [-1.60, -1.00] & 0.002 & -0.94 & 10 \\
interruption 300 (preserved) vs habituation & integration & -0.60 [-1.10, -0.10] & 0.109 & -0.42 & 10 \\
interruption 300 (preserved) vs habituation & development & -0.70 [-1.40, -0.10] & 0.125 & -0.48 & 10 \\
interruption 300 (preserved) vs habituation & coherence & +0.00 [-0.30, +0.30] & 1.000 & +0.00 & 10 \\
\bottomrule
\end{tabular}
\end{adjustbox}
\caption{Generated-only protocol. Document level: paired contrasts.}
\label{tab:gen-document-2}
\end{table}

\begin{table}[H]
\centering
\scriptsize
\begin{adjustbox}{max width=\textwidth}
\begin{tabular}{@{}lcccc@{}}
\toprule
condition & n cells & surprise & connection & coherence \\
\midrule
bare & 10 & 0.62 [0.40, 0.85] & 0.48 [0.25, 0.73] & 3.32 [2.33, 4.48] \\
bare + habituation & 10 & 1.35 [0.90, 1.87] & 1.03 [0.73, 1.33] & 4.83 [4.10, 5.65] \\
habituation + reseed 150 & 10 & 2.66 [2.00, 3.29] & 3.23 [2.52, 3.93] & 5.03 [4.38, 5.70] \\
\bottomrule
\end{tabular}
\end{adjustbox}
\caption{Generated-only protocol. P2: the ladder on Qwen3-8B-Base without quantization (bf16): cell means (mean over cells [95\% CI over cells]).}
\label{tab:gen-p2}
\end{table}

\begin{table}[H]
\centering
\scriptsize
\begin{adjustbox}{max width=\textwidth}
\begin{tabular}{@{}p{4.6cm}cccccc@{}}
\toprule
condition & dim & $\Delta$ [CI] & p (perm) & q (BH) & Cliff $\delta$ & n seeds \\
\midrule
bare & surprise & -0.73 [-1.18, -0.28] & 0.020 & 0.029 & -0.60 & 10 \\
bare & connection & -0.55 [-0.83, -0.28] & 0.008 & 0.016 & -0.62 & 10 \\
bare & coherence & -1.52 [-2.03, -0.97] & 0.004 & 0.016 & -0.64 & 10 \\
habituation + reseed 150 & surprise & +1.31 [+0.20, +2.34] & 0.055 & 0.066 & +0.69 & 10 \\
habituation + reseed 150 & connection & +2.19 [+1.28, +3.07] & 0.006 & 0.016 & +0.94 & 10 \\
habituation + reseed 150 & coherence & +0.20 [-0.90, +1.28] & 0.768 & 0.768 & +0.18 & 10 \\
\bottomrule
\end{tabular}
\end{adjustbox}
\caption{Generated-only protocol. vs bare + habituation: paired by seed.}
\label{tab:gen-p2-2}
\end{table}

\begin{table}[H]
\centering
\scriptsize
\begin{adjustbox}{max width=\textwidth}
\begin{tabular}{@{}lcccc@{}}
\toprule
condition & n cells & surprise & connection & coherence \\
\midrule
bare & 10 & 0.58 [0.40, 0.80] & 0.42 [0.27, 0.57] & 3.62 [2.87, 4.28] \\
bare + habituation & 10 & 0.75 [0.47, 1.03] & 0.75 [0.50, 1.00] & 4.22 [3.45, 4.83] \\
habituation + reseed 150 & 10 & 1.87 [1.42, 2.35] & 2.28 [1.67, 2.93] & 4.87 [3.93, 5.75] \\
\bottomrule
\end{tabular}
\end{adjustbox}
\caption{Generated-only protocol. P1: the ladder on the post-trained Qwen3-8B (8-bit), no chat template: cell means (mean over cells [95\% CI over cells]).}
\label{tab:gen-p1}
\end{table}

\begin{table}[H]
\centering
\scriptsize
\begin{adjustbox}{max width=\textwidth}
\begin{tabular}{@{}p{4.6cm}cccccc@{}}
\toprule
condition & dim & $\Delta$ [CI] & p (perm) & q (BH) & Cliff $\delta$ & n seeds \\
\midrule
bare & surprise & -0.17 [-0.43, +0.08] & 0.320 & 0.384 & -0.24 & 10 \\
bare & connection & -0.33 [-0.62, -0.08] & 0.086 & 0.129 & -0.48 & 10 \\
bare & coherence & -0.60 [-1.03, -0.10] & 0.045 & 0.090 & -0.33 & 10 \\
habituation + reseed 150 & surprise & +1.12 [+0.57, +1.77] & 0.002 & 0.006 & +0.79 & 10 \\
habituation + reseed 150 & connection & +1.53 [+0.90, +2.30] & 0.002 & 0.006 & +0.87 & 10 \\
habituation + reseed 150 & coherence & +0.65 [-0.53, +1.95] & 0.395 & 0.395 & +0.30 & 10 \\
\bottomrule
\end{tabular}
\end{adjustbox}
\caption{Generated-only protocol. vs bare + habituation: paired by seed.}
\label{tab:gen-p1-2}
\end{table}

\begin{table}[H]
\centering
\scriptsize
\begin{adjustbox}{max width=\textwidth}
\begin{tabular}{@{}lcccc@{}}
\toprule
condition & n cells & surprise & connection & coherence \\
\midrule
no interruption & 10 & 1.37 [1.03, 1.72] & 1.00 [0.65, 1.37] & 4.35 [3.35, 5.35] \\
paragraph break (sham) & 10 & 1.32 [0.78, 2.03] & 1.04 [0.50, 1.73] & 4.28 [3.48, 5.04] \\
subject change, context preserved & 10 & 2.74 [2.14, 3.39] & 2.28 [1.70, 2.93] & 5.98 [5.47, 6.45] \\
subject change, context reset & 10 & 3.77 [3.65, 3.88] & 3.08 [2.80, 3.40] & 6.32 [5.85, 6.72] \\
\bottomrule
\end{tabular}
\end{adjustbox}
\caption{Generated-only protocol. Reset vs preserved vs sham at period 300 on Qwen3-8B-Base: cell means (mean over cells [95\% CI over cells]).}
\label{tab:gen-reset}
\end{table}

\begin{table}[H]
\centering
\scriptsize
\begin{adjustbox}{max width=\textwidth}
\begin{tabular}{@{}p{4.6cm}cccccc@{}}
\toprule
condition & dim & $\Delta$ [CI] & p (perm) & q (BH) & Cliff $\delta$ & n seeds \\
\midrule
paragraph break (sham) & surprise & -0.05 [-0.77, +0.80] & 0.953 & 0.953 & -0.31 & 10 \\
paragraph break (sham) & connection & +0.04 [-0.60, +0.76] & 0.932 & 0.953 & -0.12 & 10 \\
paragraph break (sham) & coherence & -0.08 [-1.27, +1.07] & 0.914 & 0.953 & +0.01 & 10 \\
subject change, context preserved & surprise & +1.38 [+0.74, +1.98] & 0.008 & 0.018 & +0.77 & 10 \\
subject change, context preserved & connection & +1.28 [+0.73, +1.83] & 0.006 & 0.018 & +0.74 & 10 \\
subject change, context preserved & coherence & +1.63 [+0.70, +2.57] & 0.012 & 0.021 & +0.58 & 10 \\
subject change, context reset & surprise & +2.40 [+2.08, +2.70] & 0.002 & 0.009 & +1.00 & 10 \\
subject change, context reset & connection & +2.08 [+1.62, +2.52] & 0.002 & 0.009 & +1.00 & 10 \\
subject change, context reset & coherence & +1.97 [+0.72, +3.15] & 0.023 & 0.035 & +0.70 & 10 \\
\bottomrule
\end{tabular}
\end{adjustbox}
\caption{Generated-only protocol. vs no interruption: paired by seed.}
\label{tab:gen-reset-2}
\end{table}

\begin{table}[H]
\centering
\scriptsize
\begin{adjustbox}{max width=\textwidth}
\begin{tabular}{@{}lcccc@{}}
\toprule
condition & n cells & surprise & connection & coherence \\
\midrule
no interruption & 10 & 1.96 [1.30, 2.61] & 1.26 [0.77, 1.76] & 3.99 [2.89, 5.07] \\
paragraph break (sham) & 10 & 1.92 [1.40, 2.44] & 1.52 [0.93, 2.19] & 4.05 [2.96, 5.11] \\
subject change, context preserved & 10 & 3.60 [3.10, 4.15] & 3.54 [3.07, 4.10] & 5.65 [5.15, 6.17] \\
subject change, context reset & 10 & 4.10 [3.78, 4.42] & 2.90 [2.62, 3.18] & 6.65 [6.23, 7.03] \\
\bottomrule
\end{tabular}
\end{adjustbox}
\caption{Generated-only protocol. Reset vs preserved vs sham at period 300 on OLMo-2-13B: cell means (mean over cells [95\% CI over cells]).}
\label{tab:gen-reset-3}
\end{table}

\begin{table}[H]
\centering
\scriptsize
\begin{adjustbox}{max width=\textwidth}
\begin{tabular}{@{}p{4.6cm}cccccc@{}}
\toprule
condition & dim & $\Delta$ [CI] & p (perm) & q (BH) & Cliff $\delta$ & n seeds \\
\midrule
paragraph break (sham) & surprise & -0.04 [-0.91, +0.82] & 0.938 & 0.938 & -0.08 & 10 \\
paragraph break (sham) & connection & +0.27 [-0.48, +1.01] & 0.541 & 0.696 & +0.09 & 10 \\
paragraph break (sham) & coherence & +0.06 [-1.33, +1.37] & 0.938 & 0.938 & -0.01 & 10 \\
subject change, context preserved & surprise & +1.64 [+0.74, +2.51] & 0.012 & 0.021 & +0.80 & 10 \\
subject change, context preserved & connection & +2.28 [+1.58, +2.95] & 0.004 & 0.009 & +0.97 & 10 \\
subject change, context preserved & coherence & +1.66 [+0.37, +2.88] & 0.043 & 0.064 & +0.53 & 10 \\
subject change, context reset & surprise & +2.14 [+1.32, +2.92] & 0.004 & 0.009 & +0.90 & 10 \\
subject change, context reset & connection & +1.64 [+1.10, +2.25] & 0.002 & 0.009 & +0.90 & 10 \\
subject change, context reset & coherence & +2.66 [+1.61, +3.85] & 0.002 & 0.009 & +0.83 & 10 \\
\bottomrule
\end{tabular}
\end{adjustbox}
\caption{Generated-only protocol. vs no interruption: paired by seed.}
\label{tab:gen-reset-4}
\end{table}

\begin{table}[H]
\centering
\scriptsize
\begin{adjustbox}{max width=\textwidth}
\begin{tabular}{@{}lcccc@{}}
\toprule
condition & n cells & surprise & connection & coherence \\
\midrule
no interruption & 10 & 1.35 [1.12, 1.63] & 0.95 [0.68, 1.23] & 4.48 [3.92, 5.03] \\
subject change, context preserved & 10 & 2.95 [2.63, 3.23] & 2.42 [2.07, 2.93] & 6.52 [6.10, 6.90] \\
subject change, context reset & 10 & 2.95 [2.58, 3.35] & 2.45 [2.02, 2.92] & 6.63 [6.32, 6.92] \\
\bottomrule
\end{tabular}
\end{adjustbox}
\caption{Generated-only protocol. Second genre: expository openings on the main generator (period 300): cell means (mean over cells [95\% CI over cells]).}
\label{tab:gen-second}
\end{table}

\begin{table}[H]
\centering
\scriptsize
\begin{adjustbox}{max width=\textwidth}
\begin{tabular}{@{}p{4.6cm}cccccc@{}}
\toprule
condition & dim & $\Delta$ [CI] & p (perm) & q (BH) & Cliff $\delta$ & n seeds \\
\midrule
subject change, context preserved & surprise & +1.60 [+1.18, +2.02] & 0.002 & 0.002 & +0.97 & 10 \\
subject change, context preserved & connection & +1.47 [+1.02, +2.02] & 0.002 & 0.002 & +1.00 & 10 \\
subject change, context preserved & coherence & +2.03 [+1.42, +2.65] & 0.002 & 0.002 & +0.96 & 10 \\
subject change, context reset & surprise & +1.60 [+1.17, +2.00] & 0.002 & 0.002 & +0.98 & 10 \\
subject change, context reset & connection & +1.50 [+0.90, +2.10] & 0.004 & 0.004 & +0.92 & 10 \\
subject change, context reset & coherence & +2.15 [+1.42, +2.87] & 0.002 & 0.002 & +0.98 & 10 \\
\bottomrule
\end{tabular}
\end{adjustbox}
\caption{Generated-only protocol. vs no interruption: paired by seed.}
\label{tab:gen-second-2}
\end{table}

\begin{table}[H]
\centering
\scriptsize
\begin{adjustbox}{max width=\textwidth}
\begin{tabular}{@{}lcccc@{}}
\toprule
condition & n cells & surprise & connection & coherence \\
\midrule
no interruption & 10 & 1.60 [1.18, 2.00] & 1.32 [0.87, 1.82] & 4.40 [3.45, 5.42] \\
clock 150 & 10 & 3.02 [2.53, 3.53] & 3.68 [2.92, 4.52] & 6.12 [5.60, 6.57] \\
clock 300 & 10 & 2.90 [2.45, 3.35] & 2.40 [2.03, 2.78] & 6.40 [6.08, 6.70] \\
judge-gated 150 & 10 & 2.67 [2.12, 3.20] & 3.03 [2.47, 3.58] & 5.93 [5.63, 6.27] \\
\bottomrule
\end{tabular}
\end{adjustbox}
\caption{Generated-only protocol. Judge-gated interruption (DREAM's Review with a gate that opens) vs the clock: cell means (mean over cells [95\% CI over cells]).}
\label{tab:gen-judge-gated}
\end{table}

\begin{table}[H]
\centering
\scriptsize
\begin{adjustbox}{max width=\textwidth}
\begin{tabular}{@{}p{4.6cm}cccccc@{}}
\toprule
condition & dim & $\Delta$ [CI] & p (perm) & q (BH) & Cliff $\delta$ & n seeds \\
\midrule
no interruption & surprise & -1.42 [-2.03, -0.92] & 0.002 & 0.009 & -0.85 & 10 \\
no interruption & connection & -2.37 [-3.28, -1.57] & 0.002 & 0.009 & -0.93 & 10 \\
no interruption & coherence & -1.72 [-2.90, -0.35] & 0.043 & 0.097 & -0.61 & 10 \\
clock 300 & surprise & -0.12 [-0.75, +0.47] & 0.781 & 0.781 & -0.02 & 10 \\
clock 300 & connection & -1.28 [-2.10, -0.53] & 0.008 & 0.023 & -0.64 & 10 \\
clock 300 & coherence & +0.28 [-0.28, +0.82] & 0.389 & 0.500 & +0.22 & 10 \\
judge-gated 150 & surprise & -0.35 [-0.95, +0.12] & 0.320 & 0.480 & -0.22 & 10 \\
judge-gated 150 & connection & -0.65 [-1.47, +0.05] & 0.180 & 0.323 & -0.22 & 10 \\
judge-gated 150 & coherence & -0.18 [-0.62, +0.28] & 0.508 & 0.571 & -0.26 & 10 \\
\bottomrule
\end{tabular}
\end{adjustbox}
\caption{Generated-only protocol. vs clock 150: paired by seed.}
\label{tab:gen-judge-gated-2}
\end{table}

\begin{table}[H]
\centering
\scriptsize
\begin{adjustbox}{max width=\textwidth}
\begin{tabular}{@{}lcccc@{}}
\toprule
condition & n cells & surprise & connection & coherence \\
\midrule
no interruption & 10 & 1.35 [0.73, 2.07] & 1.02 [0.52, 1.70] & 4.37 [3.42, 5.33] \\
subject change, context preserved & 10 & 2.83 [2.57, 3.10] & 2.68 [2.18, 3.23] & 6.27 [5.82, 6.68] \\
paragraph break (sham) & 10 & 1.98 [1.47, 2.50] & 1.33 [0.75, 2.03] & 6.04 [5.42, 6.60] \\
subject change, no habituation & 10 & 2.40 [1.92, 2.93] & 2.07 [1.62, 2.53] & 5.42 [4.70, 6.12] \\
subject change, context reset & 10 & 4.08 [3.65, 4.47] & 3.22 [2.83, 3.58] & 6.68 [6.45, 6.92] \\
\bottomrule
\end{tabular}
\end{adjustbox}
\caption{Generated-only protocol. Confirmatory: period 300 on new premises: cell means (mean over cells [95\% CI over cells]).}
\label{tab:gen-confirmatory}
\end{table}

\begin{table}[H]
\centering
\scriptsize
\begin{adjustbox}{max width=\textwidth}
\begin{tabular}{@{}p{4.6cm}cccccc@{}}
\toprule
condition & dim & $\Delta$ [CI] & p (perm) & q (BH) & Cliff $\delta$ & n seeds \\
\midrule
subject change, context preserved & surprise & +1.48 [+0.70, +2.07] & 0.006 & 0.018 & +0.71 & 10 \\
subject change, context preserved & connection & +1.67 [+0.72, +2.40] & 0.016 & 0.035 & +0.78 & 10 \\
subject change, context preserved & coherence & +1.90 [+0.73, +3.02] & 0.018 & 0.035 & +0.67 & 10 \\
paragraph break (sham) & surprise & +0.63 [-0.07, +1.43] & 0.164 & 0.179 & +0.42 & 10 \\
paragraph break (sham) & connection & +0.32 [-0.42, +1.22] & 0.602 & 0.602 & +0.27 & 10 \\
paragraph break (sham) & coherence & +1.68 [+0.54, +2.75] & 0.029 & 0.050 & +0.61 & 10 \\
subject change, no habituation & surprise & +1.05 [+0.03, +1.95] & 0.078 & 0.104 & +0.62 & 10 \\
subject change, no habituation & connection & +1.05 [+0.08, +1.87] & 0.068 & 0.103 & +0.68 & 10 \\
subject change, no habituation & coherence & +1.05 [-0.12, +2.17] & 0.121 & 0.145 & +0.39 & 10 \\
subject change, context reset & surprise & +2.73 [+2.03, +3.33] & 0.002 & 0.016 & +0.94 & 10 \\
subject change, context reset & connection & +2.20 [+1.53, +2.68] & 0.004 & 0.016 & +0.85 & 10 \\
subject change, context reset & coherence & +2.32 [+1.32, +3.28] & 0.004 & 0.016 & +0.86 & 10 \\
\bottomrule
\end{tabular}
\end{adjustbox}
\caption{Generated-only protocol. vs no interruption: paired by seed.}
\label{tab:gen-confirmatory-2}
\end{table}

\begin{table}[H]
\centering
\scriptsize
\begin{adjustbox}{max width=\textwidth}
\begin{tabular}{@{}p{4.6cm}ccccc@{}}
\toprule
hypothesis & contrast & dim & $\Delta$ (mean of paired differences) [CI] & p & n seeds \\
\midrule
H1 (primary) & clock300 vs bare\_habit & surprise & +1.48 [+0.70, +2.07] & 0.0029 & 10 \\
H2 & clock300 vs sham\_break300 & surprise & +0.85 [+0.20, +1.45] & 0.0205 & 10 \\
H3 & clock300 vs reset\_reseed300 & connection & -0.53 [-1.02, -0.02] & 0.9639 & 10 \\
H4 (two-sided) & clock300 vs nohabit300 & surprise & +0.43 [-0.08, +0.92] & 0.1641 & 10 \\
\bottomrule
\end{tabular}
\end{adjustbox}
\caption{Generated-only protocol. Pre-registered contrasts (exact one-sided sign-flip permutation unless stated).}
\label{tab:gen-pre-registered}
\end{table}

\begin{table}[H]
\centering
\scriptsize
\begin{adjustbox}{max width=\textwidth}
\begin{tabular}{@{}lc@{}}
\toprule
condition & copied \\
\midrule
bare + habituation & 47\% \\
paragraph break 300 (sham) & 12\% \\
reseed 300 & 58\% \\
reseed 300, no habituation & 67\% \\
reset + new subject 300 & 0\% \\
\bottomrule
\end{tabular}
\end{adjustbox}
\caption{Generated-only protocol. Confirmatory: self-copy rates and the pre-registered contrasts on fresh windows only.}
\label{tab:gen-confirmatory-3}
\end{table}

\begin{table}[H]
\centering
\scriptsize
\begin{adjustbox}{max width=\textwidth}
\begin{tabular}{@{}p{4.6cm}ccccc@{}}
\toprule
hypothesis (fresh only) & contrast & dim & $\Delta$ [CI] & p & n seeds \\
\midrule
H1 & clock300 vs bare\_habit & surprise & +0.82 [+0.03, +1.44] & 0.0352 & 10 \\
H2 & clock300 vs sham\_break300 & surprise & +0.52 [-0.60, +1.52] & 0.1982 & 10 \\
H3 & clock300 vs reset\_reseed300 & connection & -0.92 [-1.47, -0.30] & 0.9912 & 10 \\
H4 (two-sided) & clock300 vs nohabit300 & surprise & +0.00 [-0.80, +0.73] & 1.0000 & 10 \\
\bottomrule
\end{tabular}
\end{adjustbox}
\caption{Generated-only protocol. Confirmatory: self-copy rates and the pre-registered contrasts on fresh windows only.}
\label{tab:gen-confirmatory-4}
\end{table}

\begin{table}[H]
\centering
\scriptsize
\begin{adjustbox}{max width=\textwidth}
\begin{tabular}{@{}p{4.6cm}ccccc@{}}
\toprule
condition & n & integration & development & coherence & surprise \\
\midrule
bare + habituation & 10 & 1.40 & 0.90 & 1.50 & 1.10 \\
paragraph break 300 (sham) & 10 & 0.60 & 0.40 & 1.40 & 1.10 \\
reseed 300 & 10 & 1.20 & 0.40 & 1.30 & 0.60 \\
reseed 300, no habituation & 10 & 0.80 & 0.10 & 1.20 & 0.10 \\
reset + new subject 300 & 10 & 2.10 & 1.60 & 1.90 & 2.50 \\
\bottomrule
\end{tabular}
\end{adjustbox}
\caption{Generated-only protocol. Confirmatory premises: document level (50 documents).}
\label{tab:gen-confirmatory-5}
\end{table}

\begin{table}[H]
\centering
\scriptsize
\begin{adjustbox}{max width=\textwidth}
\begin{tabular}{@{}lcccc@{}}
\toprule
contrast & dim & $\Delta$ [CI] & p (perm) & n \\
\midrule
preserved vs habituation & integration & -0.20 [-0.90, +0.50] & 0.797 & 10 \\
preserved vs habituation & development & -0.50 [-1.00, +0.10] & 0.234 & 10 \\
preserved vs habituation & coherence & -0.20 [-0.60, +0.20] & 0.625 & 10 \\
preserved vs habituation & surprise & -0.50 [-1.20, +0.20] & 0.312 & 10 \\
reset vs habituation & integration & +0.70 [+0.10, +1.20] & 0.094 & 10 \\
reset vs habituation & development & +0.70 [+0.30, +1.10] & 0.031 & 10 \\
reset vs habituation & coherence & +0.40 [+0.00, +0.80] & 0.219 & 10 \\
reset vs habituation & surprise & +1.40 [+0.90, +1.90] & 0.004 & 10 \\
reset vs preserved & integration & +0.90 [+0.40, +1.40] & 0.031 & 10 \\
reset vs preserved & development & +1.20 [+0.80, +1.60] & 0.004 & 10 \\
reset vs preserved & coherence & +0.60 [+0.30, +0.90] & 0.031 & 10 \\
reset vs preserved & surprise & +1.90 [+1.50, +2.30] & 0.002 & 10 \\
\bottomrule
\end{tabular}
\end{adjustbox}
\caption{Generated-only protocol. Confirmatory premises: document level (50 documents).}
\label{tab:gen-confirmatory-6}
\end{table}

\subsection*{Event-window protocol of the first version (descriptive)}
\begin{table}[H]
\centering
\scriptsize
\begin{adjustbox}{max width=\textwidth}
\begin{tabular}{@{}lcccccc@{}}
\toprule
dimension & condition & n & mean $\pm$ sd & scaffold $-$ cond CI95 & Cliff's $\delta$ & Mann-Whitney p \\
\midrule
surprise & DREAM scaffold ($\lambda$=0) & 47 & 3.19 $\pm$ 1.92 & -- & -- & -- \\
surprise & salience only & 38 & 2.63 $\pm$ 1.75 & [-0.22, +1.33] & +0.16 & 0.195 \\
surprise & bare + clock reseed & 60 & 3.08 $\pm$ 1.48 & [-0.54, +0.78] & +0.00 & 0.967 \\
surprise & bare generation & 50 & 0.28 $\pm$ 0.72 & [+2.34, +3.50] & +0.88 & 4.58e-15 \\
connection & DREAM scaffold ($\lambda$=0) & 47 & 1.64 $\pm$ 1.56 & -- & -- & -- \\
connection & salience only & 38 & 1.39 $\pm$ 1.42 & [-0.40, +0.87] & +0.09 & 0.446 \\
connection & bare + clock reseed & 60 & 3.15 $\pm$ 1.53 & [-2.09, -0.91] & -0.58 & 1.52e-07 \\
connection & bare generation & 50 & 0.20 $\pm$ 0.57 & [+0.98, +1.93] & +0.65 & 7.08e-10 \\
coherence & DREAM scaffold ($\lambda$=0) & 47 & 3.68 $\pm$ 1.69 & -- & -- & -- \\
coherence & salience only & 38 & 3.71 $\pm$ 1.60 & [-0.72, +0.67] & -0.03 & 0.819 \\
coherence & bare + clock reseed & 60 & 4.78 $\pm$ 0.95 & [-1.65, -0.56] & -0.42 & 0.00014 \\
coherence & bare generation & 50 & 2.06 $\pm$ 2.01 & [+0.87, +2.35] & +0.57 & 5.34e-07 \\
\bottomrule
\end{tabular}
\end{adjustbox}
\caption{Ablation battery (Qwen3-30B-A3B, 10 seeds, Opus 5 k=5): scaffold vs each condition, unit = window.}
\label{tab:app-ablation}
\end{table}

\begin{table}[H]
\centering
\scriptsize
\begin{adjustbox}{max width=\textwidth}
\begin{tabular}{@{}lccc@{}}
\toprule
dimension & windows & mean spread & p90 spread \\
\midrule
surprise & 242 & 0.61 & 1.00 \\
connection & 242 & 0.60 & 1.00 \\
coherence & 242 & 0.68 & 1.00 \\
\bottomrule
\end{tabular}
\end{adjustbox}
\caption{Instrument calibration: intra-window spread of k=5 judgments per dimension (Opus 5).}
\label{tab:app-calib}
\end{table}

\begin{table}[H]
\centering
\scriptsize
\begin{adjustbox}{max width=\textwidth}
\begin{tabular}{@{}lcccccc@{}}
\toprule
arm & n & novel 4-grams \% & novel 6-grams \% & $\Delta$4 vs baseline CI95 & Cliff's $\delta$ & max copied span (words) \\
\midrule
min-p baseline & 15 & 21.5 $\pm$ 10.1 & 70.9 $\pm$ 11.9 & -- & -- & 10 \\
$\lambda$=1 & 15 & 36.0 $\pm$ 12.4 & 87.6 $\pm$ 7.4 & [+0.069, +0.230] & +0.68 & 10 \\
$\lambda$=2 & 15 & 45.5 $\pm$ 10.8 & 89.7 $\pm$ 5.9 & [+0.163, +0.314] & +0.88 & 13 \\
\bottomrule
\end{tabular}
\end{adjustbox}
\caption{Phase 1: objective novelty against the OLMo-2 training corpus (3 prompts $\times$ 5 seeds).}
\label{tab:app-phase1}
\end{table}

\begin{table}[H]
\centering
\scriptsize
\begin{adjustbox}{max width=\textwidth}
\begin{tabular}{@{}lcccc@{}}
\toprule
arm & runs & train-plateau escapes & mean champion test excess & best \\
\midrule
plain & 5 & 2/5 & 0.0602 & 0.0600 \\
antiprob & 5 & 3/5 & 0.0604 & 0.0601 \\
\bottomrule
\end{tabular}
\end{adjustbox}
\caption{Verified search on Qwen3-30B-A3B (bin packing): the anti-probable operator does not separate from plain sampling.}
\label{tab:app-verified}
\end{table}

\begin{table}[H]
\centering
\scriptsize
\begin{adjustbox}{max width=\textwidth}
\begin{tabular}{@{}lcccccc@{}}
\toprule
condition & windows & kind & surprise & connection & coherence & good (S$\geq$5\&H$\geq$5) \\
\midrule
bare (no habituation, no interruption) & 50 & jump & 0.28 $\pm$ 0.72 & 0.20 $\pm$ 0.57 & 2.06 $\pm$ 2.01 & 0/50 \\
bare + habituation (no interruption) & 60 & clock & 1.70 $\pm$ 1.71 & 1.08 $\pm$ 1.10 & 4.08 $\pm$ 2.19 & 4/60 \\
bare + habituation + clock reseed 150 & 60 & clock & 3.08 $\pm$ 1.48 & 3.15 $\pm$ 1.53 & 4.78 $\pm$ 0.95 & 12/60 \\
DREAM scaffold ($\lambda$=0) & 47 & crystallize/jump/recurrence & 3.19 $\pm$ 1.92 & 1.64 $\pm$ 1.56 & 3.68 $\pm$ 1.69 & 8/47 \\
\bottomrule
\end{tabular}
\end{adjustbox}
\caption{Battery 2, Q1: habituation and interruption (Qwen3-30B-A3B).}
\label{tab:app-q1}
\end{table}

\begin{table}[H]
\centering
\scriptsize
\begin{adjustbox}{max width=\textwidth}
\begin{tabular}{@{}p{4.2cm}lcccc@{}}
\toprule
condition & dim & $\Delta$ [CI] & $\delta$ & p & paired-by-seed $\Delta$ [CI] (n seeds) \\
\midrule
bare (no habituation, no interruption) & surprise & -1.42 [-1.89, -0.95] & -0.58 & 1.4e-08 & -1.42 [-1.72, -1.12] (10) \\
bare (no habituation, no interruption) & connection & -0.88 [-1.20, -0.56] & -0.52 & 1.1e-07 & -0.88 [-1.22, -0.54] (10) \\
bare (no habituation, no interruption) & coherence & -2.02 [-2.79, -1.23] & -0.55 & 3.4e-07 & -2.02 [-3.37, -0.68] (10) \\
bare + habituation + clock reseed 150 & surprise & +1.38 [+0.80, +1.93] & +0.50 & 1.3e-06 & +1.38 [+0.92, +1.92] (10) \\
bare + habituation + clock reseed 150 & connection & +2.07 [+1.60, +2.55] & +0.75 & 5.2e-13 & +2.07 [+1.63, +2.57] (10) \\
bare + habituation + clock reseed 150 & coherence & +0.70 [+0.10, +1.30] & +0.21 & 4.3e-02 & +0.70 [-0.27, +1.72] (10) \\
DREAM scaffold ($\lambda$=0) & surprise & +1.49 [+0.80, +2.19] & +0.46 & 3.4e-05 & +1.80 [+0.97, +2.75] (10) \\
DREAM scaffold ($\lambda$=0) & connection & +0.55 [+0.06, +1.10] & +0.20 & 6.2e-02 & +0.61 [+0.22, +1.01] (10) \\
DREAM scaffold ($\lambda$=0) & coherence & -0.40 [-1.14, +0.34] & -0.10 & 3.8e-01 & -0.26 [-1.03, +0.56] (10) \\
\bottomrule
\end{tabular}
\end{adjustbox}
\caption{Battery 2, Q1: habituation and interruption (Qwen3-30B-A3B); differences vs the reference arm ($\Delta$ mean with 95\% CI over windows; Cliff's $\delta$; Mann--Whitney p; paired-by-seed $\Delta$ [CI]).}
\label{tab:app-q1-diff}
\end{table}

\begin{table}[H]
\centering
\scriptsize
\begin{adjustbox}{max width=\textwidth}
\begin{tabular}{@{}lcccccc@{}}
\toprule
condition & windows & kind & surprise & connection & coherence & good (S$\geq$5\&H$\geq$5) \\
\midrule
neutral subject change (clock 150) & 60 & clock & 3.08 $\pm$ 1.48 & 3.15 $\pm$ 1.53 & 4.78 $\pm$ 0.95 & 12/60 \\
re-encounter stitch (clock 150) & 50 & cut & 2.68 $\pm$ 1.46 & 3.64 $\pm$ 1.57 & 4.72 $\pm$ 1.33 & 5/50 \\
the premise itself (clock 150) & 50 & cut & 1.14 $\pm$ 1.73 & 1.04 $\pm$ 1.57 & 3.18 $\pm$ 1.90 & 3/50 \\
the stream's own past (clock 150) & 50 & cut & 1.42 $\pm$ 1.43 & 1.52 $\pm$ 1.35 & 3.06 $\pm$ 1.42 & 2/50 \\
\bottomrule
\end{tabular}
\end{adjustbox}
\caption{Battery 2, Q2: content of the interruption at period 150.}
\label{tab:app-q2}
\end{table}

\begin{table}[H]
\centering
\scriptsize
\begin{adjustbox}{max width=\textwidth}
\begin{tabular}{@{}p{4.2cm}lcccc@{}}
\toprule
condition & dim & $\Delta$ [CI] & $\delta$ & p & paired-by-seed $\Delta$ [CI] (n seeds) \\
\midrule
re-encounter stitch (clock 150) & surprise & -0.40 [-0.95, +0.14] & -0.13 & 2.4e-01 & -0.40 [-1.29, +0.46] (10) \\
re-encounter stitch (clock 150) & connection & +0.49 [-0.09, +1.07] & +0.20 & 7.1e-02 & +0.49 [-0.56, +1.56] (10) \\
re-encounter stitch (clock 150) & coherence & -0.06 [-0.50, +0.38] & -0.00 & 9.8e-01 & -0.06 [-0.75, +0.57] (10) \\
the premise itself (clock 150) & surprise & -1.94 [-2.53, -1.31] & -0.68 & 6.1e-10 & -1.94 [-2.65, -1.30] (10) \\
the premise itself (clock 150) & connection & -2.11 [-2.68, -1.51] & -0.72 & 4.1e-11 & -2.11 [-2.83, -1.46] (10) \\
the premise itself (clock 150) & coherence & -1.60 [-2.15, -1.01] & -0.53 & 1.2e-06 & -1.60 [-2.36, -0.70] (10) \\
the stream's own past (clock 150) & surprise & -1.66 [-2.21, -1.11] & -0.61 & 2.1e-08 & -1.66 [-2.26, -1.07] (10) \\
the stream's own past (clock 150) & connection & -1.63 [-2.17, -1.08] & -0.62 & 9.0e-09 & -1.63 [-2.37, -0.91] (10) \\
the stream's own past (clock 150) & coherence & -1.72 [-2.18, -1.24] & -0.70 & 1.1e-10 & -1.72 [-2.30, -1.11] (10) \\
\bottomrule
\end{tabular}
\end{adjustbox}
\caption{Battery 2, Q2: content of the interruption at period 150; differences vs the reference arm ($\Delta$ mean with 95\% CI over windows; Cliff's $\delta$; Mann--Whitney p; paired-by-seed $\Delta$ [CI]).}
\label{tab:app-q2-diff}
\end{table}

\begin{table}[H]
\centering
\scriptsize
\begin{adjustbox}{max width=\textwidth}
\begin{tabular}{@{}lcccccc@{}}
\toprule
condition & windows & kind & surprise & connection & coherence & good (S$\geq$5\&H$\geq$5) \\
\midrule
re-encounter on the clock (150) & 50 & cut & 2.68 $\pm$ 1.46 & 3.64 $\pm$ 1.57 & 4.72 $\pm$ 1.33 & 5/50 \\
re-encounter on the clock (900; matched frequency) & 40 & cut & 2.73 $\pm$ 1.95 & 2.30 $\pm$ 1.82 & 4.67 $\pm$ 1.59 & 6/40 \\
neutral change on the clock (900; matched frequency) & 40 & cut & 2.48 $\pm$ 1.87 & 2.02 $\pm$ 1.68 & 4.95 $\pm$ 2.16 & 5/40 \\
re-encounter on salience events (event windows) & 80 & crystallize/cut/jump/recurrence & 3.92 $\pm$ 1.42 & 2.99 $\pm$ 1.73 & 3.75 $\pm$ 1.18 & 8/80 \\
re-encounter on salience events (uniform windows) & 60 & clock & 2.18 $\pm$ 2.04 & 1.68 $\pm$ 1.65 & 3.93 $\pm$ 1.88 & 4/60 \\
re-encounter on the clock (900), uniform windows & 60 & clock & 4.70 $\pm$ 1.69 & 3.48 $\pm$ 1.52 & 5.50 $\pm$ 1.41 & 31/60 \\
neutral change on the clock (900), uniform windows & 60 & clock & 5.48 $\pm$ 1.64 & 2.87 $\pm$ 1.56 & 5.95 $\pm$ 1.32 & 45/60 \\
salience only, no injection (event windows) & 38 & crystallize/jump/recurrence & 2.63 $\pm$ 1.75 & 1.39 $\pm$ 1.42 & 3.71 $\pm$ 1.60 & 2/38 \\
\bottomrule
\end{tabular}
\end{adjustbox}
\caption{Battery 2, Q3: timing of the re-encounter, clock vs salience.}
\label{tab:app-q3}
\end{table}

\begin{table}[H]
\centering
\scriptsize
\begin{adjustbox}{max width=\textwidth}
\begin{tabular}{@{}p{4.2cm}lcccc@{}}
\toprule
condition & dim & $\Delta$ [CI] & $\delta$ & p & paired-by-seed $\Delta$ [CI] (n seeds) \\
\midrule
re-encounter on the clock (900; matched frequency) & surprise & +0.04 [-0.68, +0.78] & -0.03 & 8.1e-01 & +0.04 [-0.53, +0.66] (10) \\
re-encounter on the clock (900; matched frequency) & connection & -1.34 [-2.03, -0.62] & -0.45 & 2.2e-04 & -1.34 [-2.17, -0.53] (10) \\
re-encounter on the clock (900; matched frequency) & coherence & -0.04 [-0.65, +0.58] & -0.02 & 8.9e-01 & -0.05 [-0.87, +0.73] (10) \\
neutral change on the clock (900; matched frequency) & surprise & -0.21 [-0.90, +0.51] & -0.10 & 4.4e-01 & -0.21 [-0.84, +0.36] (10) \\
neutral change on the clock (900; matched frequency) & connection & -1.62 [-2.28, -0.92] & -0.51 & 2.5e-05 & -1.61 [-2.17, -1.13] (10) \\
neutral change on the clock (900; matched frequency) & coherence & +0.23 [-0.56, +0.99] & +0.11 & 3.5e-01 & +0.23 [-0.45, +1.07] (10) \\
re-encounter on salience events (event windows) & surprise & +1.24 [+0.73, +1.75] & +0.44 & 2.0e-05 & +1.10 [+0.34, +1.85] (10) \\
\bottomrule
\end{tabular}
\end{adjustbox}
\caption{Battery 2, Q3: timing of the re-encounter, clock vs salience; differences vs the reference arm ($\Delta$ mean with 95\% CI over windows; Cliff's $\delta$; Mann--Whitney p; paired-by-seed $\Delta$ [CI]).}
\label{tab:app-q3-diff}
\end{table}

\begin{table}[H]
\centering
\scriptsize
\begin{adjustbox}{max width=\textwidth}
\begin{tabular}{@{}lcccccc@{}}
\toprule
condition & windows & kind & surprise & connection & coherence & good (S$\geq$5\&H$\geq$5) \\
\midrule
every 75 (before injection) & 50 & cut & 0.86 $\pm$ 1.04 & 1.28 $\pm$ 1.04 & 3.08 $\pm$ 0.63 & 1/50 \\
every 150 (before injection) & 60 & clock & 3.08 $\pm$ 1.48 & 3.15 $\pm$ 1.53 & 4.78 $\pm$ 0.95 & 12/60 \\
every 300 (before injection) & 50 & cut & 2.90 $\pm$ 1.50 & 2.70 $\pm$ 1.63 & 5.88 $\pm$ 0.82 & 7/50 \\
every 600 (before injection) & 50 & cut & 3.06 $\pm$ 1.58 & 2.56 $\pm$ 1.66 & 5.32 $\pm$ 1.62 & 6/50 \\
every 900 (before injection) & 40 & cut & 2.48 $\pm$ 1.87 & 2.02 $\pm$ 1.68 & 4.95 $\pm$ 2.16 & 5/40 \\
every 300 (mid-segment) & 60 & clock & 4.93 $\pm$ 1.53 & 3.47 $\pm$ 1.68 & 6.07 $\pm$ 0.98 & 34/60 \\
every 600 (mid-segment) & 60 & clock & 4.12 $\pm$ 1.83 & 2.70 $\pm$ 1.63 & 5.65 $\pm$ 1.54 & 23/60 \\
every 900 (mid-segment) & 60 & clock & 5.48 $\pm$ 1.64 & 2.87 $\pm$ 1.56 & 5.95 $\pm$ 1.32 & 45/60 \\
\bottomrule
\end{tabular}
\end{adjustbox}
\caption{Battery 2, Q4: frequency of interruption (neutral change).}
\label{tab:app-q4}
\end{table}

\begin{table}[H]
\centering
\scriptsize
\begin{adjustbox}{max width=\textwidth}
\begin{tabular}{@{}p{4.2cm}lcccc@{}}
\toprule
condition & dim & $\Delta$ [CI] & $\delta$ & p & paired-by-seed $\Delta$ [CI] (n seeds) \\
\midrule
every 75 (before injection) & surprise & -2.22 [-2.69, -1.75] & -0.83 & 1.7e-14 & -2.22 [-2.91, -1.73] (10) \\
every 75 (before injection) & connection & -1.87 [-2.35, -1.38] & -0.76 & 1.1e-12 & -1.87 [-2.49, -1.32] (10) \\
every 75 (before injection) & coherence & -1.70 [-2.00, -1.41] & -0.86 & 1.5e-15 & -1.70 [-2.00, -1.39] (10) \\
every 300 (before injection) & surprise & -0.18 [-0.74, +0.37] & -0.06 & 5.7e-01 & -0.18 [-0.81, +0.39] (10) \\
every 300 (before injection) & connection & -0.45 [-1.04, +0.12] & -0.18 & 1.1e-01 & -0.45 [-1.31, +0.45] (10) \\
every 300 (before injection) & coherence & +1.10 [+0.78, +1.42] & +0.59 & 2.5e-08 & +1.10 [+0.69, +1.57] (10) \\
every 600 (before injection) & surprise & -0.02 [-0.60, +0.54] & -0.00 & 9.8e-01 & -0.02 [-0.74, +0.60] (10) \\
\bottomrule
\end{tabular}
\end{adjustbox}
\caption{Battery 2, Q4: frequency of interruption (neutral change); differences vs the reference arm ($\Delta$ mean with 95\% CI over windows; Cliff's $\delta$; Mann--Whitney p; paired-by-seed $\Delta$ [CI]).}
\label{tab:app-q4-diff}
\end{table}

\begin{table}[H]
\centering
\scriptsize
\begin{adjustbox}{max width=\textwidth}
\begin{tabular}{@{}lccccccc@{}}
\toprule
every & windows & pooled surprise & conn & coh & early ($\leq$160 tokens since injection) S / C / H & late (>160) S / C / H & phase-weighted stream mean S / C / H \\
\midrule
75 & 60 & 0.88 & 1.35 & 3.15 & 0.88 / 1.35 / 3.15 (n=60) & ,  /: /: (n=0) & 0.88 / 1.35 / 3.15 \\
150 & 60 & 3.08 & 3.15 & 4.78 & 3.08 / 3.15 / 4.78 (n=60) & ,  /: /: (n=0) & 3.08 / 3.15 / 4.78 \\
300 & 110 & 4.01 & 3.12 & 5.98 & 5.28 / 3.35 / 5.95 (n=40) & 3.29 / 2.99 / 6.00 (n=70) & 4.35 / 3.18 / 5.97 \\
600 & 110 & 3.64 & 2.64 & 5.50 & 5.40 / 3.20 / 5.65 (n=20) & 3.24 / 2.51 / 5.47 (n=90) & 3.82 / 2.69 / 5.52 \\
900 & 100 & 4.28 & 2.53 & 5.55 & 5.90 / 2.80 / 6.10 (n=40) & 3.20 / 2.35 / 5.18 (n=60) & 3.68 / 2.43 / 5.35 \\
\bottomrule
\end{tabular}
\end{adjustbox}
\caption{Battery 2, Q4b: phase within the segment and phase-weighted stream means.}
\label{tab:app-q4b}
\end{table}

\begin{table}[H]
\centering
\scriptsize
\begin{adjustbox}{max width=\textwidth}
\begin{tabular}{@{}lcccccc@{}}
\toprule
condition & windows & kind & surprise & connection & coherence & good (S$\geq$5\&H$\geq$5) \\
\midrule
8B: bare & 110 & clock/jump & 0.68 $\pm$ 1.26 & 0.38 $\pm$ 0.73 & 3.45 $\pm$ 2.64 & 1/110 \\
8B: bare + habituation & 60 & clock & 1.47 $\pm$ 1.69 & 0.87 $\pm$ 1.23 & 3.93 $\pm$ 2.26 & 4/60 \\
8B: bare + habituation + clock reseed 150 & 60 & clock/cut & 2.73 $\pm$ 1.41 & 3.23 $\pm$ 1.60 & 4.42 $\pm$ 1.28 & 1/60 \\
8B: DREAM scaffold ($\lambda$=0) & 146 & clock/crystallize/cut/jump & 2.73 $\pm$ 2.13 & 1.53 $\pm$ 1.47 & 4.30 $\pm$ 1.88 & 15/146 \\
\bottomrule
\end{tabular}
\end{adjustbox}
\caption{Second generator family (Qwen3-8B-Base).}
\label{tab:app-q5}
\end{table}

\begin{table}[H]
\centering
\scriptsize
\begin{adjustbox}{max width=\textwidth}
\begin{tabular}{@{}p{4.2cm}lcccc@{}}
\toprule
condition & dim & $\Delta$ [CI] & $\delta$ & p & paired-by-seed $\Delta$ [CI] (n seeds) \\
\midrule
8B: bare & surprise & -0.78 [-1.28, -0.30] & -0.30 & 3.0e-04 & -0.78 [-1.13, -0.45] (10) \\
8B: bare & connection & -0.48 [-0.84, -0.16] & -0.27 & 7.1e-04 & -0.48 [-0.84, -0.14] (10) \\
8B: bare & coherence & -0.49 [-1.23, +0.28] & -0.13 & 1.5e-01 & -0.49 [-2.43, +1.58] (10) \\
8B: bare + habituation + clock reseed 150 & surprise & +1.26 [+0.69, +1.80] & +0.47 & 6.3e-06 & +1.26 [+0.52, +2.14] (10) \\
8B: bare + habituation + clock reseed 150 & connection & +2.37 [+1.87, +2.87] & +0.80 & 9.4e-15 & +2.37 [+1.68, +3.12] (10) \\
8B: bare + habituation + clock reseed 150 & coherence & +0.48 [-0.18, +1.13] & +0.13 & 2.1e-01 & +0.48 [-0.53, +1.43] (10) \\
8B: DREAM scaffold ($\lambda$=0) & surprise & +1.27 [+0.70, +1.81] & +0.38 & 1.8e-05 & +1.17 [+0.62, +1.68] (10) \\
8B: DREAM scaffold ($\lambda$=0) & connection & +0.66 [+0.25, +1.04] & +0.31 & 2.7e-04 & +0.67 [+0.27, +1.07] (10) \\
8B: DREAM scaffold ($\lambda$=0) & coherence & +0.36 [-0.30, +1.01] & +0.10 & 2.4e-01 & +0.52 [-0.44, +1.27] (10) \\
\bottomrule
\end{tabular}
\end{adjustbox}
\caption{Second generator family (Qwen3-8B-Base); differences vs the reference arm ($\Delta$ mean with 95\% CI over windows; Cliff's $\delta$; Mann--Whitney p; paired-by-seed $\Delta$ [CI]).}
\label{tab:app-q5-diff}
\end{table}

\begin{table}[H]
\centering
\scriptsize
\begin{adjustbox}{max width=\textwidth}
\begin{tabular}{@{}lcccccc@{}}
\toprule
condition & windows & kind & surprise & connection & coherence & good (S$\geq$5\&H$\geq$5) \\
\midrule
OLMo: bare & 109 & clock/jump & 1.35 $\pm$ 1.80 & 0.91 $\pm$ 1.53 & 2.97 $\pm$ 2.65 & 6/109 \\
OLMo: bare + habituation & 59 & clock & 2.37 $\pm$ 1.91 & 1.51 $\pm$ 1.50 & 4.05 $\pm$ 2.68 & 7/59 \\
OLMo: bare + habituation + clock reseed 150 & 60 & clock/cut & 3.11 $\pm$ 1.63 & 3.17 $\pm$ 1.54 & 3.28 $\pm$ 1.63 & 6/60 \\
OLMo: DREAM scaffold ($\lambda$=0) & 144 & clock/crystallize/cut/jump & 3.08 $\pm$ 1.94 & 1.89 $\pm$ 1.54 & 4.48 $\pm$ 2.32 & 22/144 \\
\bottomrule
\end{tabular}
\end{adjustbox}
\caption{Third generator family (OLMo-2-13B).}
\label{tab:app-q6}
\end{table}

\begin{table}[H]
\centering
\scriptsize
\begin{adjustbox}{max width=\textwidth}
%
\end{adjustbox}
\caption{Third generator family (OLMo-2-13B); differences vs the reference arm ($\Delta$ mean with 95\% CI over windows; Cliff's $\delta$; Mann--Whitney p; paired-by-seed $\Delta$ [CI]).}
\label{tab:app-q6-diff}
\end{table}

\begin{table}[H]
\centering
\scriptsize
\begin{adjustbox}{max width=\textwidth}
\begin{tabular}{@{}lcccccccc@{}}
\toprule
condition & n & radius L0 & radius L12 & radius L24 & radius L36 & radius L47 & commit layer (idx) & final entropy \\
\midrule
DREAM scaffold & 10 & 0.526 & 0.406 & 0.388 & 0.350 & 0.333 & 10.88 [10.82, 10.96] & 1.08 \\
salience only & 10 & 0.433 & 0.381 & 0.375 & 0.355 & 0.335 & 10.94 [10.83, 11.05] & 0.88 \\
bare + clock reseed & 10 & 0.418 & 0.389 & 0.383 & 0.381 & 0.339 & 10.96 [10.92, 11.00] & 0.42 \\
bare & 10 & 0.131 & 0.214 & 0.237 & 0.252 & 0.242 & 11.12 [11.04, 11.25] & 0.34 \\
\bottomrule
\end{tabular}
\end{adjustbox}
\caption{Residual-stream geometry (Qwen3-30B-A3B, ablation battery): explored radius per captured layer (subset), stable commitment layer index and final entropy, means over cells.}
\label{tab:app-h1}
\end{table}

\end{document}